\documentclass[10pt,twocolumn,letterpaper]{article}

\usepackage[pagenumbers]{cvpr} % To force page numbers, e.g. for an arXiv version

\definecolor{cvprblue}{rgb}{0.21,0.49,0.74}
\usepackage[pagebackref,breaklinks,colorlinks,allcolors=cvprblue]{hyperref}

\usepackage{amsmath,amsfonts,bm}

\def\eqref#1{equation~\ref{#1}}
\def\1{\bm{1}}

\DeclareMathAlphabet{\mathsfit}{\encodingdefault}{\sfdefault}{m}{sl}
\SetMathAlphabet{\mathsfit}{bold}{\encodingdefault}{\sfdefault}{bx}{n}

\usepackage{url}

\usepackage{algorithm}
\usepackage{algorithmic}
\usepackage{booktabs}
\usepackage{multirow}
\usepackage{float}

\usepackage{colortbl}

\definecolor{cb_orange}{RGB}{213,94,0}
\definecolor{cb_green}{RGB}{34,136,51}
\definecolor{cbgreen}{RGB}{34,136,51}
\definecolor{sky_blue}{RGB}{204, 238, 255}
\definecolor{cb_purple}{RGB}{170, 51, 119}
\definecolor{cb_red}{RGB}{204, 51, 17}
\definecolor{cb_blue}{RGB}{0, 119, 187}
\definecolor{mydarkblue}{rgb}{0,0.08,0.45}
\definecolor{forestgreen}{RGB}{34,139,34}
\definecolor{periwinkle}{rgb}{0.8, 0.8, 1.0}
\definecolor{royalazure}{rgb}{0.0, 0.22, 0.66}
\definecolor{royalblue}{rgb}{0.0, 0.14, 0.4}
\definecolor{richlilac}{rgb}{0.71, 0.4, 0.82}

\newcommand{\purplecell}[1]{\cellcolor{richlilac!25}#1}

\usepackage{bbding}
\usepackage{tcolorbox}

\def\paperID{***} % *** Enter the Paper ID here
\def\confName{CVPR}
\def\confYear{2026}

\title{MultiBanana: A Challenging Benchmark for Multi-Reference\\Text-to-Image Generation}
\author{
    Yuta Oshima$^{1,*}$ \quad
    Daiki Miyake$^{1,*}$ \quad
    Kohsei Matsutani$^{1}$ \quad
    Yusuke Iwasawa$^{1}$ \\
    Masahiro Suzuki$^{1}$ \quad
    Yutaka Matsuo$^{1}$ \quad
    Hiroki Furuta$^{2,\dagger}$ \\[1.0ex]
    $^{1}$The University of Tokyo \qquad
    $^{2}$Google DeepMind \\[1.0ex]
    {\tt\small \{yuta.oshima, daiki.miyake\}@weblab.t.u-tokyo.ac.jp}
}

\begin{document}

\twocolumn[{%
\renewcommand\twocolumn[1][]{#1}%
\maketitle
\vspace{-1.75em}
\includegraphics[width=\linewidth]{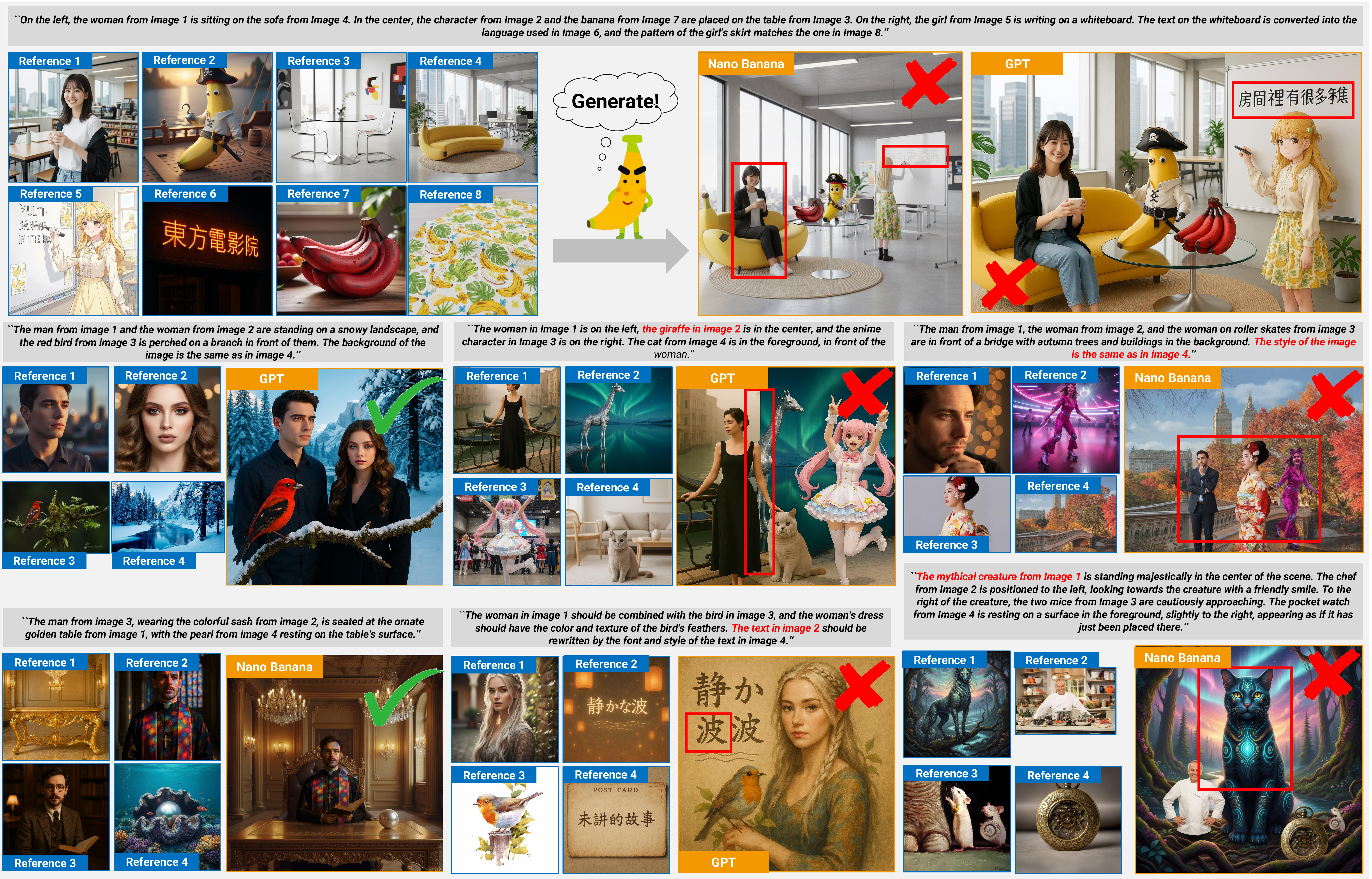}
% \vspace{-2em}
\captionof{figure}{The overview of \textbf{MultiBanana}. MultiBanana broadly covers problems specific to multi-reference settings, including varying the number of references (top row), domain and scale mismatches among references (two on the left in the middle row), multilingual text rendering (center in the bottom row), and the presence of rare concepts (right in the bottom row). \vspace{1.5em}}
\label{fig:teaser}
}]

\begingroup
\renewcommand\thefootnote{*}
\footnotetext{Equal contribution}
\renewcommand\thefootnote{$\dagger$}
\footnotetext{Work done as an advisory role only.}
\endgroup

\begin{abstract}
% Recent large-scale image generation models have acquired the ability of multi-reference generation, the ability to inherit the appearance of subjects from multiple reference images and re-render them under new contexts. 
% However, there is still no benchmark that systematically evaluates how model performance varies under different multi-reference conditions. 
% To address this gap, we introduce \textbf{MultiBanana}, a new benchmark for multi-reference image generation. 
% MultiBanana assesses model robustness across multiple reference-specific dimensions, including: (1) varying the number of references, (2) domain mismatch among references (e.g., photo vs. anime), (3) scale mismatch between reference and target scenes, (4) references containing rare concepts (e.g., a pink panda), and (5) multilingual textual references for rendering. 
% We analyze cross-model performance under these conditions to reveal robustness, typical failure modes, and areas for improvement. 
% MultiBanana will be released as an open benchmark to drive progress and establish a standardized basis for fair comparison in multi-reference image generation.
%
% However, the existing benchmark datasets often focus on the generation with single or a few reference images, which prevents us from measuring the progress on how model performance advances or pointing out their weaknesses, under different multi-reference conditions.
%
Recent text-to-image generation models have acquired the ability of multi-reference generation and editing; that is, to inherit the appearance of subjects from multiple reference images and re-render them in new contexts.
However, existing benchmark datasets often focus on generation using a single or a few reference images, which prevents us from measuring progress in model performance or identifying weaknesses when following instructions with a larger number of references.
In addition, their task definitions are still vague, limited to axes such as ``what to edit'' or ``how many references are given'', and therefore fail to capture the challenges inherent in combining heterogeneous references. 
To address this gap, we introduce \textbf{MultiBanana}, which is designed to assess the edge of model capabilities by widely covering problems specific to multi-reference settings: (1) varying the number of references (up to 8), (2) domain mismatch among references (e.g., photo vs. anime), (3) scale mismatch between reference and target scenes, (4) references containing rare concepts (e.g., a red banana), and (5) multilingual textual references for rendering. 
Our analysis among a variety of text-to-image models reveals their respective performances, typical failure modes, and areas for improvement. 
MultiBanana is released as an open benchmark to push the boundaries and establish a standardized basis for fair comparison in multi-reference image generation. Our data and code are available at \url{https://github.com/matsuolab/multibanana}.
\end{abstract}

\section{Introduction}
\label{sec:intro}

Recent advances in image generation, driven by large-scale multimodal LLM backbones, have enabled unprecedented levels of instruction-following~\citep{google2025nanobanana, openai2025gpt4oimage, wu2025qwenimagetechnicalreport, cao2025hunyuanimage3, wu2025less, xiao2024omnigen, wu2025omnigen2, xia2025dreamomni, xia2025dreamomni2, cui2025emu35nativemultimodalmodels}. 
Among these developments, multi-reference image generation has emerged as a notable capability. 
When a user provides multiple reference images, the model can inherit the subject’s appearance and re-render it in a new context while faithfully preserving identity. 
This shift moves image generation beyond single-reference conditioning toward greater controllability, driving increasing demand in industrial domains such as content production~\citep{ruiz2022dreambooth, ye2023ip-adapter, wang2024ms}, advertising~\citep{inoue2023layout, horita2024retrievalaugmented, morita2025tkg}, and fashion design~\citep{zhu2023tryondiffusion, choi2024improving, fang2024vivid}, where reference-guided generation is essential.

Despite this progress, we lack a benchmark to evaluate how well models follow instructions that involve multiple references under varying conditions.
As shown in \autoref{tab:main_benchmark_comparison}, existing benchmarks~\citep{wang2023editbench, basu2023editval, sheynin2024emuedit, zhang2023magicbrush, huang2024smartedit, yu2025anyedit, ma2024i2ebench, ye2025imgedit, ruiz2022dreambooth, wu2025omnigen2, xia2025dreamomni2} suffer from two fundamental limitations. 
% First, they have an extremely narrow coverage due to the small number of unique reference images, making it impossible to assess the impact of varying reference conditions in both single- and multi-reference tasks. Second, their task definitions still lack details, typically limited to axes such as``what to edit'' or ``how many references are given'', and therefore fail to capture the intrinsic difficulty of multi-reference settings. 
First, existing benchmarks limit the number of references to a narrow range (typically 1--4) and do not examine how well models perform when following instructions with a larger number of references.
% existing benchmark datasets often focus on generation using a single or a few reference images, which prevents us from measuring progress in model performance or identifying weaknesses when following instructions with multiple references.
Second, their task definitions are still vague, limited to axes such as ``what to edit'' or ``how many references are given'', and therefore fail to capture the challenges inherent in combining heterogeneous references. 

% To address these limitations, we introduce a new evaluation framework grounded not only in task categories but also in the intrinsic properties of the reference set itself. 
To address these limitations, we introduce a new evaluation framework that is grounded in the core challenges of multi-reference generation.
Our benchmark incorporates multiple orthogonal axes specific to the multi-reference regime, including broader coverage of the number of references (up to 8), domain gaps between references (e.g., photo--anime mixtures), mismatches in object scale between reference and target scenes, rare concepts in the references (e.g., a red banana), and multilingual textual rendering references (English, Chinese, and Japanese).

Using this benchmark, we analyze model capabilities, identify characteristic failure modes, and highlight remaining gaps. 
% For example, while performance degradation due to increasing reference counts can be mitigated by agent-style multi-stage generation pipelines, failures arising from unsupported styles or languages remain difficult to remedy with inference-time strategies.
For example, our analysis reveals that as instruction difficulty increases (through more references, cross-domain mixtures, or rare concepts), closed-source models (e.g., Nano Banana) tend to incorporate all required subjects but suffer from compositional distortion due to overfitting to reference details, while open-source models (e.g., Qwen-Image) generate visually cleaner outputs but often omit some reference subjects altogether.
We release MultiBanana as an open-source benchmark to establish a standard foundation for fair comparison and to accelerate progress in multi-reference image generation.

\begin{table*}[t]
\caption{
Comparison among major benchmarks for reference-based image generation.
Existing benchmarks do not provide systematic evaluation across diverse multi-reference conditions, support only a limited number of references, and fail to adequately account for important factors such as differences and compatibility among reference images.
Our benchmark expands the upper limit on the number of references and introduces \textit{difficult reference combinations}, including domain mismatches, scale mismatches, rare concepts, and multilingual prompts, thereby explicitly addressing challenges unique to multi-reference image generation.
% This design enables direct evaluation of the compatibility and interactions among multiple reference images, which represent core difficulties inherent to this task.
% Comparison of major benchmarks for reference-based image generation.
% Existing benchmarks lack systematic evaluation across diverse multi-reference conditions, covering only a limited number of references and vague task definitions.
% Our benchmark expands this scope by introducing multiple orthogonal evaluation axes, reference count, domain and scale mismatches, rare concepts, and multilingual prompts, enabling a more comprehensive assessment of model robustness.
}
\vspace{-2mm}
\label{tab:main_benchmark_comparison}
\centering
\scalebox{0.95}{
\begin{tabular}{lccccc}
\toprule
\textbf{Benchmark} & \textbf{\#Size} & \textbf{\#Tasks} & \textbf{\#References} & \textbf{Difficult Reference Combination} & \textbf{Metrics} \\
\midrule
EditBench~\citep{wang2023editbench}    & 240   & 1   & 1 & \textcolor{cb_red}{\XSolidBrush} & CLIP~\citep{radford2021clip} \\
EditVal~\citep{basu2023editval}        & 648   & 13  & 1 & \textcolor{cb_red}{\XSolidBrush} & CLIP, VLM, manual \\
EmuEdit~\citep{sheynin2024emuedit}        & 3055  & 7   & 1 & \textcolor{cb_red}{\XSolidBrush} & L1, CLIP, DINO~\citep{caron2021dino} \\
MagicBrush~\citep{zhang2023magicbrush}  & 1053  & 9   & 1 & \textcolor{cb_red}{\XSolidBrush} & L1, L2, CLIP, DINO \\
AnyEdit~\citep{yu2025anyedit}        & 1250  & 25  & 1 & \textcolor{cb_red}{\XSolidBrush} & L1, CLIP, DINO \\
I2EBench~\citep{ma2024i2ebench}      & 2240  & 16  & 1 & \textcolor{cb_red}{\XSolidBrush} & GPT~\citep{openai2023gpt4} \\
ImgEdit-Bench~\citep{ye2025imgedit}  & 811   & 14  & 1 & \textcolor{cb_red}{\XSolidBrush} & GPT (3 dim.), Fake Det.~\citep{xu2024fakeshield} \\
DreamBooth~\citep{ruiz2022dreambooth} & 75    & 1   & 1 & \textcolor{cb_red}{\XSolidBrush} & CLIP. DINO \\
OmniContext~\citep{wu2025omnigen2}    & 400   & 8   & 3 & \textcolor{cb_red}{\XSolidBrush} & GPT (3 dim.) \\
DreamOmni2~\citep{xia2025dreamomni2}  & 319   & 29  & 4 & \textcolor{cb_red}{\XSolidBrush} & Gemini~\citep{geminiteam2023gemini}, Doubao~\citep{bytedance2025doubao} \\
\purplecell{\textbf{MultiBanana (Ours)}}            & \purplecell{\textbf{3769}}  & \purplecell{\textbf{36}}  & \purplecell{\textbf{8}} & \purplecell{\textcolor{cbgreen}{\textbf{\Checkmark}}} & \purplecell{GPT, Gemini (\textbf{5 dim.})} \\
\bottomrule
\end{tabular}
}
\vspace{-2mm}
\end{table*}

\section{Related Works}
\label{sec:related}

\subsection{Controllable Text-to-Image Generation}
Diffusion models have achieved state-of-the-art performance in high-fidelity image synthesis~\citep{sohl-dickstein15, ho2020ddpm}.
%Representative systems, such as Stable Diffusion~\citep{rombach2022ldm, podell2024sdxl, esser2024sd3}, FLUX~\citep{flux2024, labs2025flux1kontext}, DALL-E~\citep{ramesh2022dalle}, and Imagen~\citep{saharia2022imagen, imagenteamgoogle2024imagen3}, have demonstrated strong text-to-image generation capabilities, establishing a controllable and scalable foundation for image generation.
%To enhance controllability, models such as ControlNet~\citep{zhang2023controlnet} and T2I-Adapter~\citep{mou2023t2i-adapter} introduced external conditioning modules, enabling image-conditioned generation.
Large pretrained diffusion models with textual conditioning, such as Stable Diffusion~\citep{rombach2022ldm, podell2024sdxl, esser2024sd3} and FLUX~\citep{flux2024, labs2025flux1kontext}, can generate novel images while preserving the identity depicted in the reference images.
Building on these developments, approaches based on fine-tuning (e.g., DreamBooth~\citep{ruiz2022dreambooth}) and adapter-based methods (e.g., IP-Adapter~\citep{ye2023ip-adapter}) have been explored to improve the controllability~\citep{jiang2025infiniteyou,gal2023an,brooks2023instructpix2pix,mokady2023null,tan2025ominicontrol,NEURIPS2023_602e1a5d}.
Recent research on multimodal LLM-based image generation has gained momentum, aiming to handle multiple image generation tasks within a single model.
For example, GPT-Image-1~\citep{openai2025gpt4oimage} and Nano Banana~\citep{google2025nanobanana} can jointly process text and image inputs, enabling integrated image generation and editing within a unified framework.
Similarly, open-source models such as Qwen-Image~\citep{wu2025qwenimagetechnicalreport} and FLUX.1-Kontext-dev~\citep{labs2025flux1kontext} demonstrate high-quality, flexible image generation and editing.
Beyond these, numerous open-source unified generative models continue to emerge~\citep{xiao2024omnigen, xia2025dreamomni, wu2025omnigen2, xia2025dreamomni2}.
These advancements demonstrate that diffusion models are becoming flexible and adaptable across diverse conditional generation settings.

%With the advent of multimodal foundation models capable of jointly handling text and images~\citep{google2025nanobanana, openai2025gpt4oimage}, a new task has recently become feasible: multi-reference-based image generation.
%In this task, the model receives multiple reference images along with textual instructions and generates a new scene while preserving the identity of the reference images.
%However, despite its novelty and potential, a reliable benchmark for evaluating this innovative task has yet to be established, leaving it as an essential open challenge for future research.

\subsection{Benchmarks for Reference-Based Generation}
Among reference-driven image generation tasks, image editing is the most widely recognized: it involves generating an edited image from a single reference image and a textual instruction describing how the edit should be performed.
Representative benchmarks such as MagicBrush~\citep{zhang2023magicbrush} and EMU-Edit~\citep{sheynin2024emuedit} have extended task-specific evaluations but still rely heavily on similarity-based metrics.
SmartEdit~\citep{huang2024smartedit} supports editing of complex scenes, but does not sufficiently cover more general settings.
Similarly, I2E-Bench~\citep{ma2024i2ebench} employs GPT-4o to provide human-aligned evaluations across a variety of editing tasks, yet it uses different metrics for each task and therefore fails to capture the shared characteristics of editing as a whole.
ImgEdit~\citep{ye2025imgedit} has enabled evaluation of multi-turn editing tasks, in which image generation and editing are alternated interactively.
One of the earliest benchmarks for reference-based image generation beyond instruction-based editing is the DreamBooth~\citep{ruiz2022dreambooth} dataset.
However, such benchmarks are still limited to generation tasks conditioned on a single reference image and thus cannot handle more complex or compositional settings.
Recently, advances demonstrated by GPT-Image-1~\citep{openai2025gpt4oimage} and Nano Banana~\citep{google2025nanobanana} have drawn increasing attention to multi-reference image generation, where multiple reference images are jointly used for generation.
Benchmarks such as OmniContext~\citep{wu2025omnigen2} and DreamOmni2~\citep{xia2025dreamomni2} have been proposed to evaluate this task.
Nevertheless, these benchmarks encompass only a small number of images and task types, thereby limiting the scope of the evaluation.
Moreover, although designed for multi-reference generation, they do not account for the heterogeneous nature of reference images.
This challenge is analogous to instruction following in LLMs~\citep{zhou2023ifeval, jiang-etal-2024-followbench, he2024multiif, harada2025multi-instructions, laban2026llms}, where performance degrades as the number of constraints grows; multi-reference image generation similarly requires models to satisfy all given references without neglecting any.
% , and their setups remain similar to those of conventional single-image editing benchmarks.

\begin{figure*}
    \centering
    \includegraphics[width=0.98\linewidth]{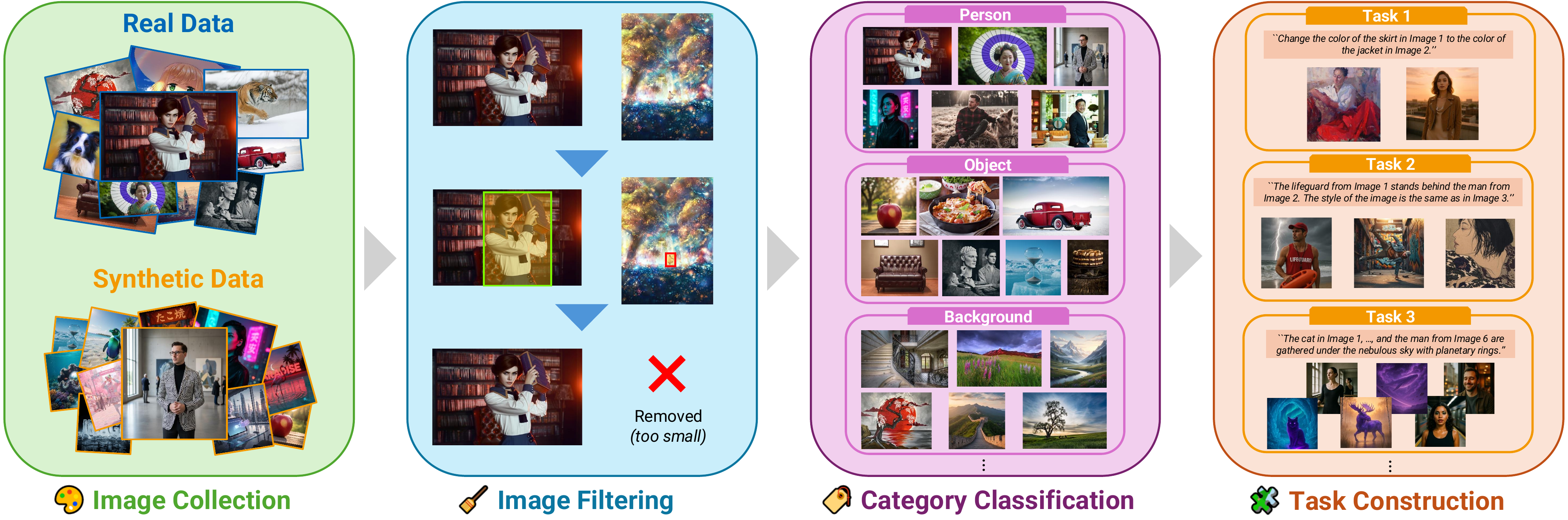}
    \vspace{-1mm}
    \caption{Construction pipeline for our benchmark, consisting of four stages: (1) collecting high-quality real and synthetic images, (2) filtering out inappropriate or low-quality samples, (3) performing hierarchical category classification, and (4) generating and validating editing instructions by Gemini and humans.}
    \label{fig:const_pipeline}
    \vspace{-2.0mm}
\end{figure*}
%どうにかChatGPTに修正させる

\section{MultiBanana}
\label{sec:multibanana}

\subsection{Task Types}
\label{sec:task_types}

In reference-driven image generation, tasks can be categorized based on the number and relational structure of reference images. 
In this work, we follow the overall design philosophy of prior benchmarks such as DreamOmni2~\citep{xia2025dreamomni2}, but extend the number of references up to 8 to enable a more fine-grained analysis of scalability. 
We also go beyond task-category-based evaluation by incorporating axes that capture the intrinsic properties of the reference set itself, such as domain gaps between references, scale mismatches, rare concepts, and multilingual text rendering.
% In this work, we follow the overall design philosophy of prior benchmarks such as DreamOmni2~\citep{xia2025dreamomni2}, but place a stronger emphasis on an aspect that has been largely underexplored so far: the diversity and compositional relationships among multiple references.  
% While existing benchmarks mainly evaluate performance in terms of the number of references or single-image editing accuracy, our benchmark explicitly measures a model’s ability to understand and reason about inter-reference variations.

\noindent\textbf{Single Reference.}~~
This is the simplest setting, corresponding to the standard image editing task.  
Given a single reference image and a textual instruction, the model generates an edited image that maintains semantic consistency and preserves the subject's visual identity.  
This configuration follows traditional instruction-driven editing benchmarks such as MagicBrush~\citep{zhang2023magicbrush} and EMU-Edit~\citep{sheynin2024emuedit}, which represent the foundation of reference-driven generation.

\noindent\textbf{Two References -- 11 Task Types.}~~
The two-reference setting provides the model with two reference images (typically a main subject and an auxiliary one) and a textual instruction.  
We define 11 editing task types, following prior work, while balancing feasibility and challenge: subject addition, subject replacement, background change, color modification, material modification, pose modification, hairstyle modification, makeup modification, tone transformation (lighting, season, weather, etc.), style transfer (oil painting, sketch, anime, etc.), and text correction (in-image text editing or replacement).
% \begin{itemize}
%     \item Subject addition
%     \item Subject replacement
%     \item Background change
%     \item Color modification
%     \item Material modification
%     \item Action modification (pose or motion)
%     \item Hairstyle modification
%     \item Makeup modification
%     \item Tone transformation (lighting, season, weather, etc.)
%     \item Style transfer (oil painting, sketch, anime, etc.)
%     \item Text correction (in-image text editing or replacement)
% \end{itemize}
These tasks span both local attribute manipulations and global transformations, evaluating a model’s ability to maintain visual coherence, subject identity, and contextual adaptability.

\noindent\textbf{Multi References -- Compositional Generation and Inter-Reference Relationships.}~~
The multi-reference setting uses 3--8 reference images, requiring the model to perform more compositional generation.  
% While prior works, such as DreamOmni2~\citep{xia2025dreamomni2}, mainly focus on compositional generation using multiple references, our benchmark takes a step further by making inter-reference diversity, consistency, and structural relations central to the evaluation.
We define four base task structures, each evaluated with $\{3, 4, 5, 6, 7, 8\}$ reference images, resulting in 24 tasks:
\begin{itemize}
    \item \textbf{X Objects}: Compose $X$ objects from the reference set.
    \item \textbf{X--1 Objects + Background}: Place $X-1$ objects within the background of another reference.
    \item \textbf{X--1 Objects + Local}: Apply a local change (specified by a reference) to one of the $X-1$ objects.
    \item \textbf{X--1 Objects + Global}: Arrange $X-1$ objects and modify the overall tone according to another reference.
\end{itemize}
This design enables us to assess whether a model can integrate multiple reference sources into a consistent scene.

\noindent\textbf{Diversity Factors in Multi-Reference Generation.}~~
A central contribution of our benchmark is the explicit incorporation of inter-reference diversity factors, which have been largely overlooked in previous benchmarks such as DreamOmni2~\citep{xia2025dreamomni2}. 
We highlight the following dimensions:
\begin{itemize}
    % \item \textbf{Number of references}: across 3--8 inputs.
    \item \textbf{Cross-domain diversity}: mixing references from different domains such as real photos, anime, CG, or sketches.
    \item \textbf{Scale and viewpoint differences}: variation in object size or camera perspective between references and outputs.
    \item \textbf{Rare concept references}: inclusion of uncommon concepts (e.g., ``a red banana'').
    \item \textbf{Multilingual references}: assessing multilingual instruction understanding during rendering.
\end{itemize}
%
% These factors go beyond conventional image-editing evaluation and are critical for assessing a model’s compositional reasoning, identity preservation, and cross-domain generalization.  
% In summary, our benchmark diverges from the conventional notion of multi-reference generation, such as in DreamOmni2, which primarily focuses on simple reference composition. Instead, it introduces a new framework to evaluate a model's capabilities to interpret relationships among heterogeneous references and integrate them into coherent generations.
% These factors go beyond prior benchmarks that focus on simple reference composition, ours evaluates whether models can interpret relationships among heterogeneous references and integrate them into coherent outputs.
These factors go beyond prior benchmarks that focus on simple reference composition; our benchmark evaluates whether models can interpret relationships among heterogeneous references and integrate them into coherent outputs.

% \begin{figure}
%     \centering
%     \includegraphics[width=\linewidth]{figures/category_distribution_donut_chart.pdf}
%     \caption{Number of images per category.}
%     \label{fig:data_distribution}
% \end{figure}

\begin{figure*}[t]
  \centering
  % \begin{minipage}[t]{0.36\linewidth}
  %   \centering
  %   \includegraphics[width=\linewidth]{figures/category_distribution_donut_chart.pdf}
  % \end{minipage}
  \begin{minipage}[t]{0.60\linewidth}
    \centering
    \includegraphics[width=\linewidth]{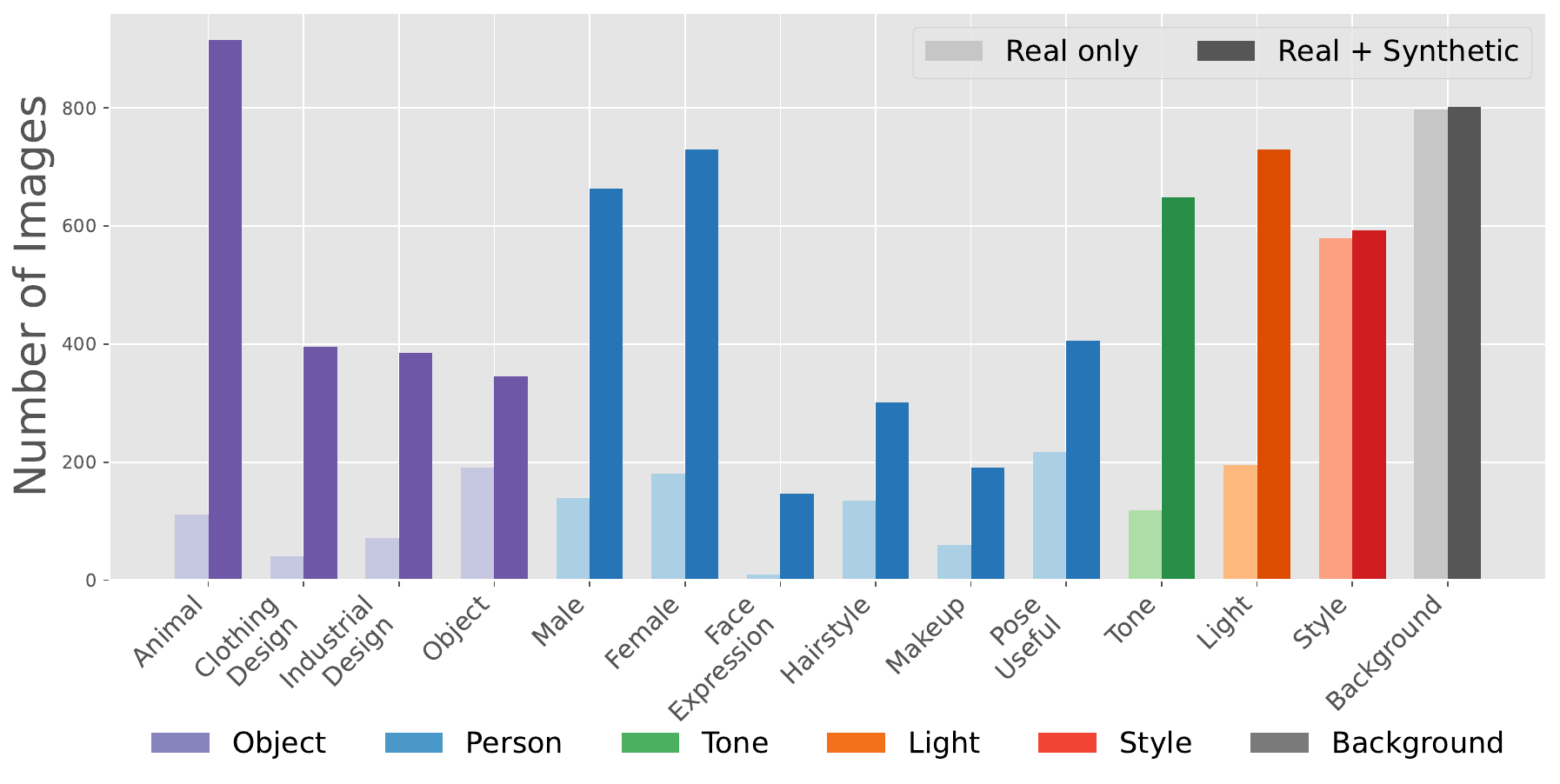}
  \end{minipage}
  \begin{minipage}[t]{0.37\linewidth}
    \centering
    \includegraphics[width=\linewidth]{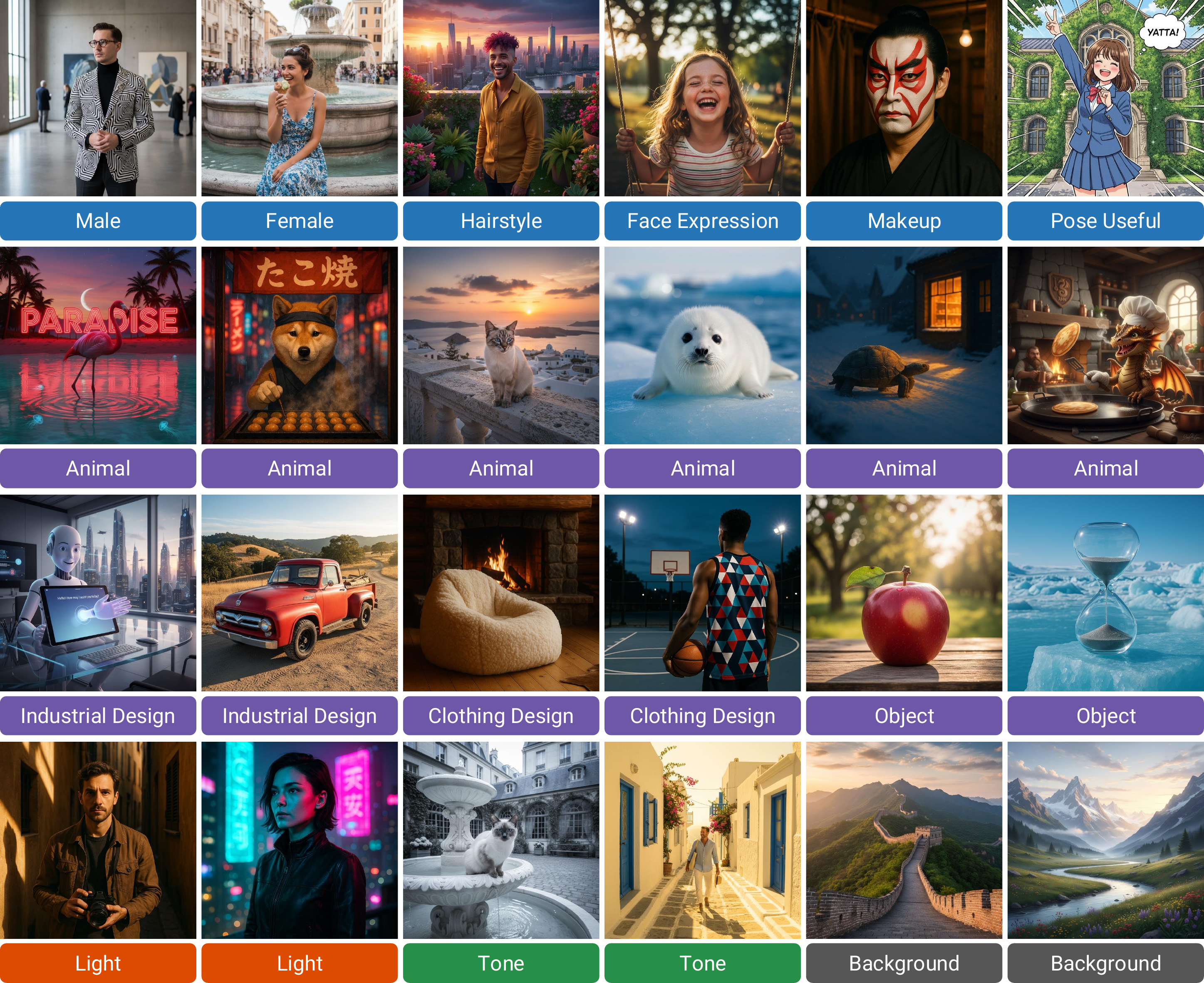}
  \end{minipage}
  \vspace{-1.5mm}
  \caption{(\textbf{Left}) Comparison between the statistics of real data only and those after adding synthetic data.
  The original dataset was biased toward background images, with few person- and object-related samples.
  To correct this imbalance, we generated additional synthetic images using Nano Banana and GPT-Image-1, focusing on clear subjects such as people, animals, and objects.
  This results in a more balanced and comprehensive distribution of data.
(\textbf{Right}) Examples of synthesized images in each category.}
  \label{fig:source_distribution}
  \vspace{-2.0mm}
\end{figure*}

\subsection{How to Construct MultiBanana}

The construction pipeline is illustrated in \autoref{fig:const_pipeline}.

\noindent\textbf{Image Collection.}~~
The reference images used in our benchmark consist of both real and synthetic data.
All real images were extracted from the LAION-5B dataset~\citep{schuhmann2022laion}.
To ensure high quality, we selected only images with an Aesthetic score~\citep{laion2022aesthentic} higher than 6.25 and a resolution greater than 512 pixels.
However, the collected real images exhibited a distributional bias (\autoref{fig:source_distribution}; left).
Specifically, landscape scenes appeared frequently, whereas humans and objects were underrepresented.
While landscape images can serve as useful background references, the scarcity of human and object references limits the diversity of editing tasks.
To address this imbalance, we generated additional synthetic images under diverse conditions using Nano Banana~\citep{google2025nanobanana} and GPT-Image-1~\citep{openai2025gpt4oimage}, and used them as supplementary reference images (\autoref{fig:source_distribution}; right).
This approach not only mitigated the categorical bias in the reference set but also expanded its coverage and diversity by incorporating both real and synthetic examples.

\noindent\textbf{Image Filtering.} \quad
Among the collected images, some contained small subjects that were difficult to use as references, or were inappropriate for reference use, such as images containing harmful content, tables, or corrupted data.
To filter out such images and retain only those suitable for a reference-based image editing benchmark, we applied the following procedure.
Following \cite{ye2025imgedit}, we extracted images that contained large and clearly visible objects suitable for editing tasks.
Object detection was performed using YOLOv12~\cite{tian2025yolov12} and SAM2~\cite{ravi2025sam}, while CLIP~\cite{radford2021clip} was used to verify semantic consistency.
Additionally, harmful or corrupted images were removed through a combination of automatic filtering with Gemini and manual inspection, resulting in the final curated set of real images.

\noindent\textbf{Category Classification.}\quad
To construct multi-reference image generation tasks, we categorized each reference image.
For example, in a portrait transformation task that modifies a person’s hairstyle or makeup, both reference images must contain a person.
Similarly, in replacement tasks involving humans or objects, both reference images must include either a person or an object, and for background replacement tasks, images categorized as background must be used.
To satisfy such task-dependent conditions, we performed hierarchical image classification on a combined set of real and synthetic images.
We first obtained automated multi-level labels using Gemini, then conducted human verification to ensure correct categorization.
At the top level, we defined six major categories: \textit{person}, \textit{object}, \textit{background}, \textit{light}, \textit{tone}, and \textit{style}.
Each of these top-level categories was further divided into more specific subcategories.
For instance, under the \textit{person category}, we included \textit{male}, \textit{female}, \textit{hairstyle}, \textit{makeup}, \textit{pose useful}, and \textit{face expression};
and under the \textit{object} category, we defined \textit{animal}, \textit{clothing design}, \textit{industrial design}, and \textit{object}.
This hierarchical labeling enabled systematic task construction.

\noindent\textbf{Task Construction.}\quad
The editing instructions were generated using Gemini.
For each task, we manually selected suitable reference categories, randomly sampled images, and presented them to Gemini to generate corresponding editing instructions.
Gemini then evaluated whether following each instruction would lead to visual breakdowns. 
For instance, in a replacement task, whether the replaced object would appear unnaturally suspended in mid-air.
Among the generated candidates, we retained only the reference–instruction pairs that did not exhibit visual breakdowns and then asked Gemini and human annotators to further filter out tasks that were nonsensical or too trivial.
% Among the generated candidates, we retained only the reference–instruction pairs that did not exhibit visual breakdowns and then asked Gemini to further assess whether each instruction constituted a challenging editing scenario, such as a cross-domain combination. 
% Finally, we manually reviewed the remaining data to remove any instructions deemed too trivial.

\noindent\textbf{Difficult Reference Categorization.}\quad
For the cross-domain category, we selected samples whose reference images spanned at least two domains (e.g., color photos, anime, paintings).
The different-scale category included samples with both close-up and non–close-up views.
The rare-concept category contained samples featuring animals with unusual colors or patterns, humans or animals with atypical makeup or clothing, or fictional creatures.
The multilingual category included samples with text in two different languages, including English, Chinese, and Japanese.
% Samples not matching any category were grouped as others.

% \input{figures_and_tables/category_example}

\begin{figure*}[t]
  \centering
  \begin{minipage}[t]{0.26\linewidth}
    \centering
    \includegraphics[width=\linewidth]{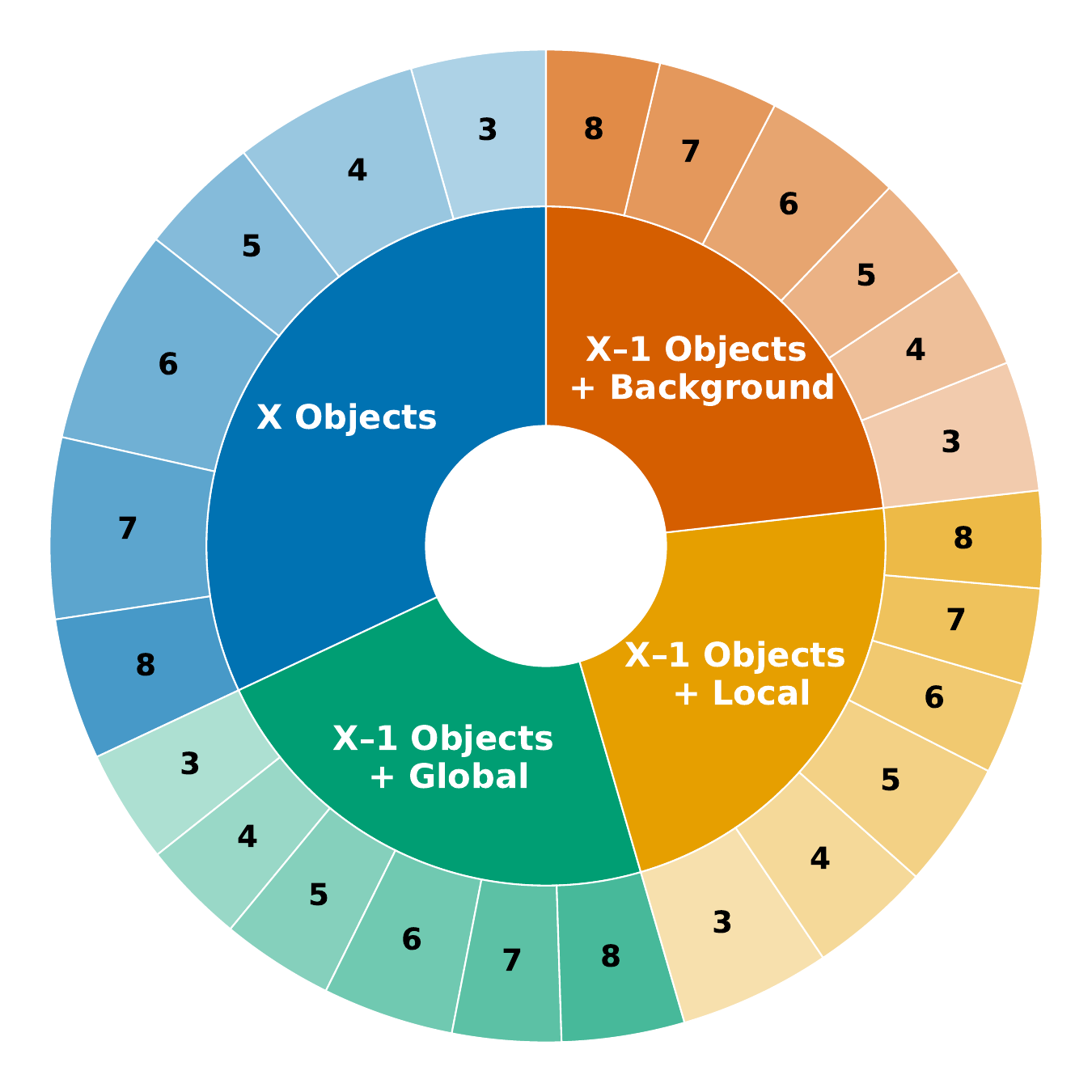}
  \end{minipage}
  \begin{minipage}[t]{0.41\linewidth}
    \centering
    \includegraphics[width=\linewidth]{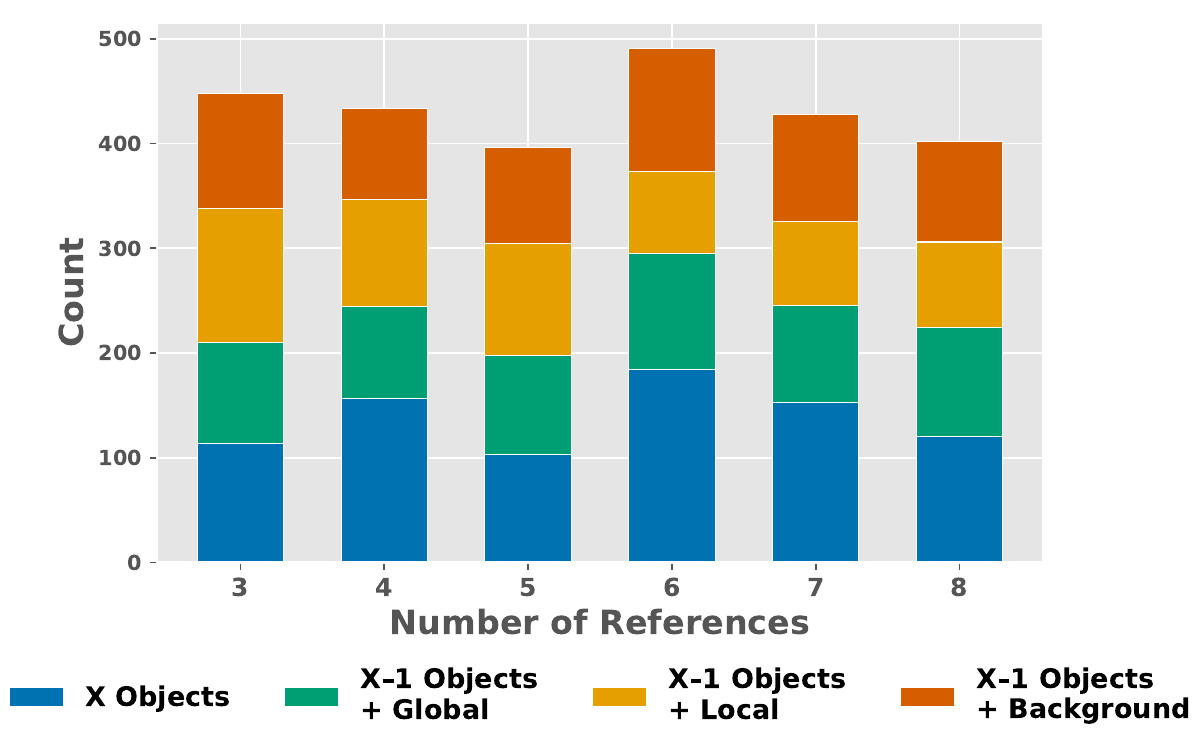}
  \end{minipage}
  \begin{minipage}[t]{0.29\linewidth}
    \centering
    \raisebox{0.05\height}{
      \includegraphics[width=\linewidth]{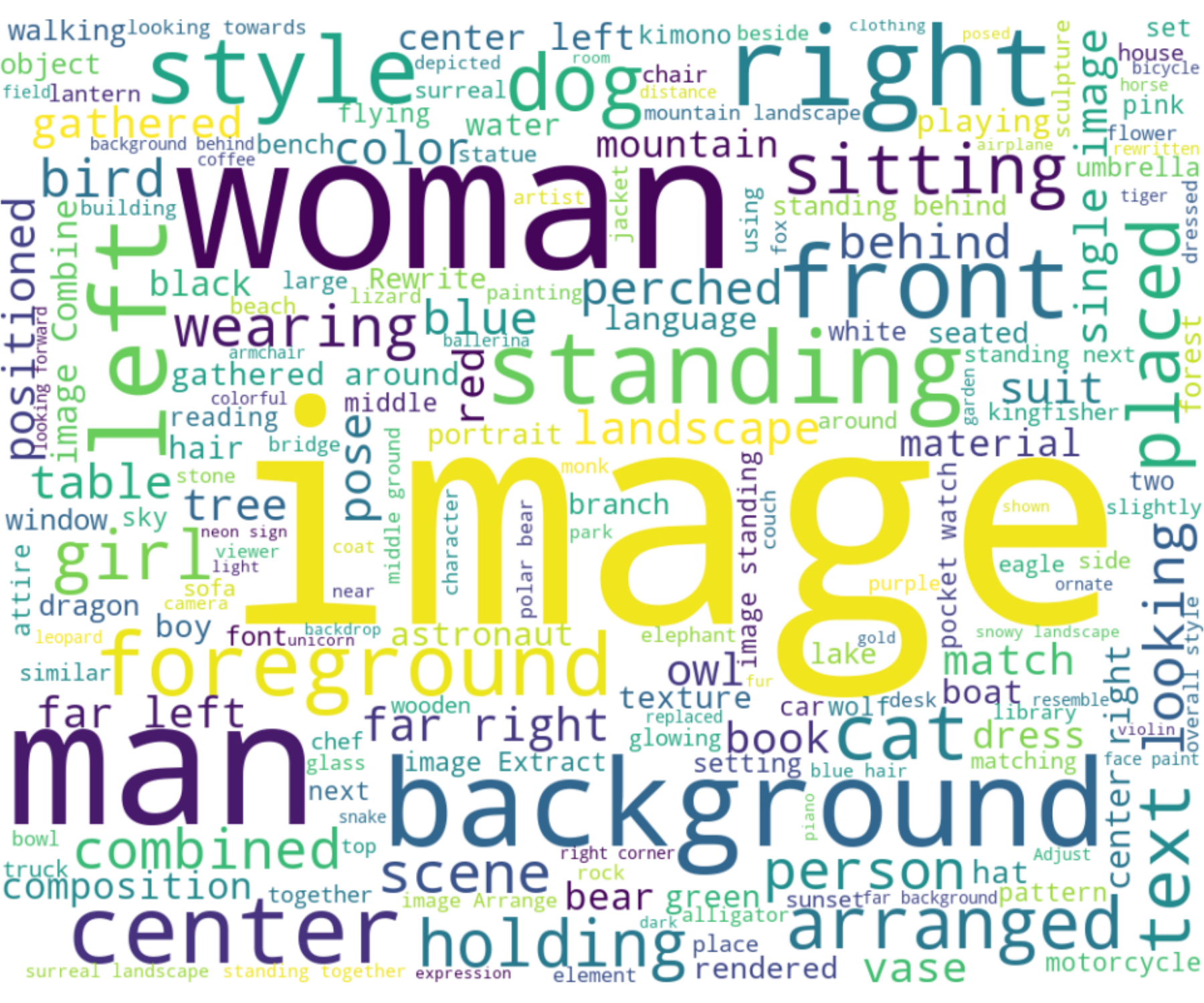}
    }
  \end{minipage}
  \vspace{-2.2mm}
  \caption{
  (\textbf{Left}) Breakdown of the multi-reference tasks.
  The editing sets were selected to ensure that the number of sets per task is balanced across reference counts.
  (\textbf{Middle}) For every X--references task, the dataset contains at least 390 editing sets. Furthermore, each colored task category includes at least 70 sets, which is larger than in prior work~\cite{xia2025dreamomni2}.
  (\textbf{Right}) Word cloud generated from all prompts. It primarily consists of terms that describe a wide range of object categories and words indicating spatial directions.
  }
  \label{fig:data_distribution}
  \vspace{-3mm}
\end{figure*}

% \begin{figure*}[t]
%   \centering
%   \begin{minipage}[t]{0.24\linewidth}
%     \centering
%     \includegraphics[width=\linewidth]{figures/reference_distribution_twolevel_customlabels_ver2.pdf}
%   \end{minipage}
%   \begin{minipage}[t]{0.4\linewidth}
%     \centering
%     \includegraphics[width=\linewidth]{figures/reference_histogram_by_number_customlegend_ver2.pdf}
%   \end{minipage}
%   \begin{minipage}[t]{0.28\linewidth}
%     \centering
%     \raisebox{0.2\height}{
%       \includegraphics[width=\linewidth]{figures/wordcloud.pdf}
%     }
%   \end{minipage}
%   \caption{
%   (\textbf{Left}) Breakdown of the multi-reference tasks.
%   The editing sets were selected to ensure that the number of sets within each task is balanced across different reference counts.
%   (\textbf{Middle}) For every reference count, the dataset contains at least 100 editing sets. Each colored task category also includes at least 20 sets, which is larger than the prior work~\cite{xia2025dreamomni2}.
%   (\textbf{Right}) Word cloud generated from all prompts. It contains object names spanning a wide range of categories.
%   }
%   \label{fig:data_distribution}
% \end{figure*}

\subsection{Statistics in MultiBanana}
\noindent\textbf{Image Category Statistics.}~~
This section presents a statistical overview of both the real and synthetic source data used to construct our benchmark.
The dataset is organized into six major categories and thirteen subcategories based on visual and semantic attributes.

In the real dataset, the largest major category is background, comprising 798 images, one-fourth of the total. These mainly consist of images without distinct subjects. The next-largest categories are person (741 images) and style (579 images), and together these three categories account for 70\% of the dataset. 
In contrast, light (195 images), object (413 images), and tone (118 images) are small in number. Although these categories are useful as references for changing global conditions such as illumination and overall atmosphere, their sample sizes remain limited.
Within the object category, there are 111 images of animals, 40 of clothing and textile design, 71 of industrial design, and 191 of miscellaneous objects, indicating that data focusing on specific materials or design patterns is scarce.
Most images in the background, style, and tone categories lack clear subjects. Furthermore, images containing textual elements, such as logos or signs, are nearly absent.
The person category is subdivided into finer attributes: face expression (10), female (181), male (139), hairstyle (135), makeup (59), and pose usefulness (217). While pose-related data account for the largest portion (around 30\%), reference data capturing subtle characteristics such as facial expression or makeup remain limited, resulting in a skewed distribution.

To address this imbalance, synthetic data were generated, focusing primarily on images containing explicit subjects (e.g., persons, animals, and objects) and on those including text. The generation process was designed to intentionally expand the coverage of underrepresented subcategories and improve the overall balance among attributes.

As a result, the distribution of major categories changed: background (802), light (729), object (2041), person (2437), style (592), and tone (649).
At the subcategory level, significant increases were observed: animal (111$\rightarrow$915), clothing design (40$\rightarrow$396), industrial design (71$\rightarrow$385), and face expression (10$\rightarrow$146), indicating a notable enrichment in person- and object-related data.

Consequently, the dataset became more balanced between background-centric and subject-centric images, with broader coverage of diverse visual attributes.
This improvement enhances the benchmark’s comprehensiveness and enables more versatile evaluation and instruction-based generation tasks under a wide range of conditions.

\noindent\textbf{Task Statistics.}~~
After filtering the editing instructions using both Gemini and manual checks, we obtained 3,769 high-quality reference images and instruction sets.
Among them, 264 sets correspond to single-reference tasks, 907 to two-reference tasks, and 2,598 to multi-reference tasks.
\autoref{fig:data_distribution} provides a further breakdown of the multi-reference tasks, showing that they are evenly distributed with respect to the number of reference images.
Additionally, for each task type, X Object, X–1 Object + Background, X–1 Object + Local, and X–1 Object + Global, we ensured that at least 70 sets were included.
This is larger than the number of sets in multi-reference tasks of OmniContext~\citep{wu2025omnigen2} and DreamOmni2~\citep{xia2025dreamomni2}, enabling more reliable benchmarking.
Furthermore, \autoref{tab:dataset_category} shows that our newly introduced evaluation axes -- cross-domain, scale and viewpoint differences, rare concept, and multilingual -- provide sufficient samples.
See Appendix~\ref{sec:further_statistics} for details.

\begin{table}[t]
\centering
\caption{
Statistics and ratios of difficult reference combinations relative to the total 3{,}769 tasks. Our benchmark provides sufficient samples for assessing a model's capacity to interpret relationships among heterogeneous references and integrate them into outputs.
% compositional reasoning, identity preservation, and cross-domain generalization.
}
\label{tab:dataset_category}
\vspace{-2mm}
\scalebox{0.88}{
\begin{tabular}{lcccc}
\toprule
\textbf{} & \textbf{Cross} & \textbf{Different} & \textbf{Rare} & \textbf{Multi-} \\
& \textbf{Domain} & \textbf{Scale \&  View} & \textbf{Concept} & \textbf{lingual} \\
\midrule
Count     & 1063 & 1357 & 743 & 99 \\
Ratio (\%) & 28.2\% & 36.0\% & 19.7\% & 2.6\% \\
\bottomrule
\end{tabular}
}
\vspace{-3mm}
\end{table}

\begin{table*}[t]
\centering
\caption{Average performance per model across different task categories (10-point scale; the higher the better).
The average scores from Gemini 2.5 and GPT-5 are reported.
Both open-source and closed-source models exhibit notably lower performance on background replacement tasks, regardless of the number of reference images.}
\vspace{-2mm}
\scalebox{1.0}{%
\begin{tabular}{lcccccc}
\midrule
\textbf{Model} &
\textbf{Single} &
\textbf{Two} &
\textbf{X Objects} &
%\textbf{X-1 Objects + Local} &
%\textbf{X-1 Objects + Global} &
%\textbf{X-1 Objects + Background} \\
\textbf{X-1 + Local} &
\textbf{X-1 + Global} &
\textbf{X-1 + Background} \\
\midrule
Nano Banana       & 7.817 & 4.891 & 4.453 & 4.118 & 4.698 & 3.575 \\
+ Agent (IPR) & 7.606 & 5.030 & 4.433 & 4.030 & 4.496 & 3.767 \\
\midrule
GPT-Image-1       & 7.804 & 6.585 & 5.086 & 5.147 & 5.757 & 5.019 \\
+ Agent (IPR) & 7.820 & 6.808 & 5.258 & 5.419 & 5.775 & 5.284 \\
\midrule
Qwen-Image-Edit-2509 & 7.499 & 3.699 & 2.256 & 2.031 & 2.351 & 2.033 \\
DreamOmni2        & 6.520 & 4.069 & 2.804 & 3.037 & 2.867 & 2.594 \\
OmniGen2          & 5.919 & 3.442 & 3.256 & 3.598 & 3.369 & 3.022 \\
\bottomrule
\end{tabular}%
}
% \vspace{-0.5mm}
\label{fig:result_main}
\end{table*}

\begin{figure*}
    \centering
    \includegraphics[width=1.0\linewidth]{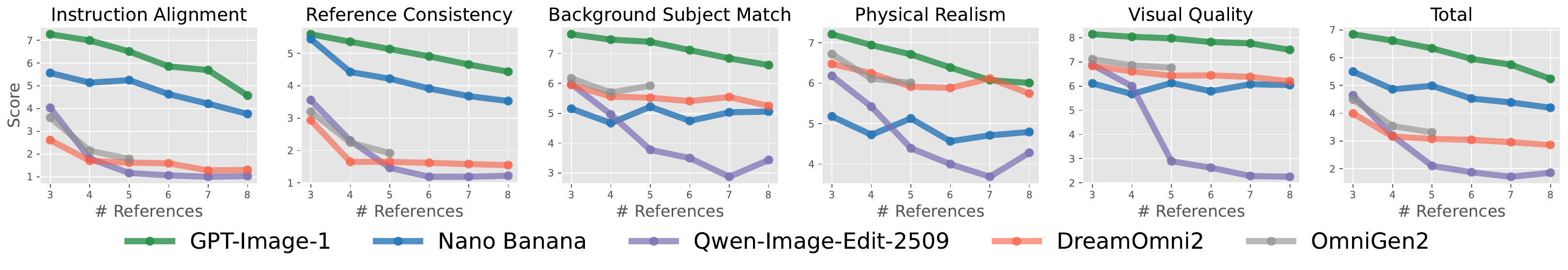}
    \vspace{-7.5mm}
    \caption{Changes in scores for each evaluation criterion when varying the number of reference images.
    Both open-source and closed-source models exhibit a general trend of decreasing scores across all metrics as the number of references increases.}
    \label{fig:results_num_references}
    \vspace{-3mm}
\end{figure*}

\subsection{Evaluation Setting}
\label{sec:evaluation_setting}
For evaluating the quality of edited images, human evaluation is among the most valuable methods, yet gathering human feedback at scale is prohibitively costly.
A practical approach to reducing time and cost is to leverage AI feedback from VLMs (e.g., Gemini 2.5~\citep{geminiteam2023gemini} and GPT-5~\citep{openai2025gpt5}), which has been shown to align reasonably with human judgments of image quality~\citep{ku2023viescore, na2024boost, wu2024boosting, furuta2024improving, oshima2025inference}. Therefore, we used Gemini 2.5 and GPT-5 to rate all generated images.

In our evaluation protocol, all generated images are assessed against five criteria to evaluate the performance of multi-reference image generation: Text-Instruction Alignment, Reference Consistency, Background-Subject Match, Physical Realism, and Visual Quality.
Text-Instruction Alignment and Reference Consistency are fundamental criteria for multi-reference image generation and core components of many prior benchmarks~\citep{ye2025imgedit, wu2025omnigen2}.
In contrast, Background-Subject Match, Physical Realism, and Visual Quality refine what previous work typically treated as a single ``overall quality'' dimension~\citep{ye2025imgedit, wu2025omnigen2}, offering a more fine-grained assessment of holistic visual fidelity.
Each of the five criteria is scored on a 10-point scale (10 is the best).
We then compute a total weighted score, assigning weights of $\lbrace$3, 3, 1, 1, 1$\rbrace$ to
$\lbrace$Instruction Alignment, Reference Consistency, Background-Subject Match, Physical Realism, Visual Quality$\rbrace$, respectively.
% See Appendix~\ref{sec:ai_human_eval} for details.

\section{Experiments}
\label{sec:experiments}

% To verify that our benchmark includes challenging tasks for advanced image gemeratopm models, we evaluate
We adopt several closed models (i.e., Nano Banana~\citep{google2025nanobanana}, GPT-Image-1~\citep{openai2025gpt4oimage}) and open models (i.e., Qwen-Image-Edit-2509~\citep{wu2025qwenimagetechnicalreport}, DreamOmni2~\citep{xia2025dreamomni2}, OmniGen2~\citep{wu2025omnigen2}) for MultiBanana evaluation.
For all models except OmniGen2, we generate outputs for every combination of reference images and text prompts included in the MultiBanana benchmark. 
Since OmniGen2 does not support more than 6 reference images, we evaluate it only under the 5-reference setting.
As described in Section~\ref{sec:evaluation_setting}, Gemini 2.5~\citep{geminiteam2023gemini} and GPT-5~\citep{openai2025gpt5} are used for evaluation. 
In the main paper, we report the average scores from Gemini 2.5 and GPT-5.
See Appendix~\ref{sec:qwen3_vl_judge} for the results of Qwen3-VL as judge, and Appendix~\ref{sec:further_results} for the detailed results and discussion. 

% \begin{figure}
%     \centering
%     % \includegraphics[width=0.7\linewidth]{figures/pipeline.pdf}
%     \includegraphics[width=\linewidth]{figures/qualitative.pdf}
%     \caption{Qualitative Results of our Benchmark (two references) with Nano Banana.}
%     \label{fig:qualitative}
% \end{figure}

\begin{figure*}[ht]
    \centering
    \includegraphics[width=\linewidth]{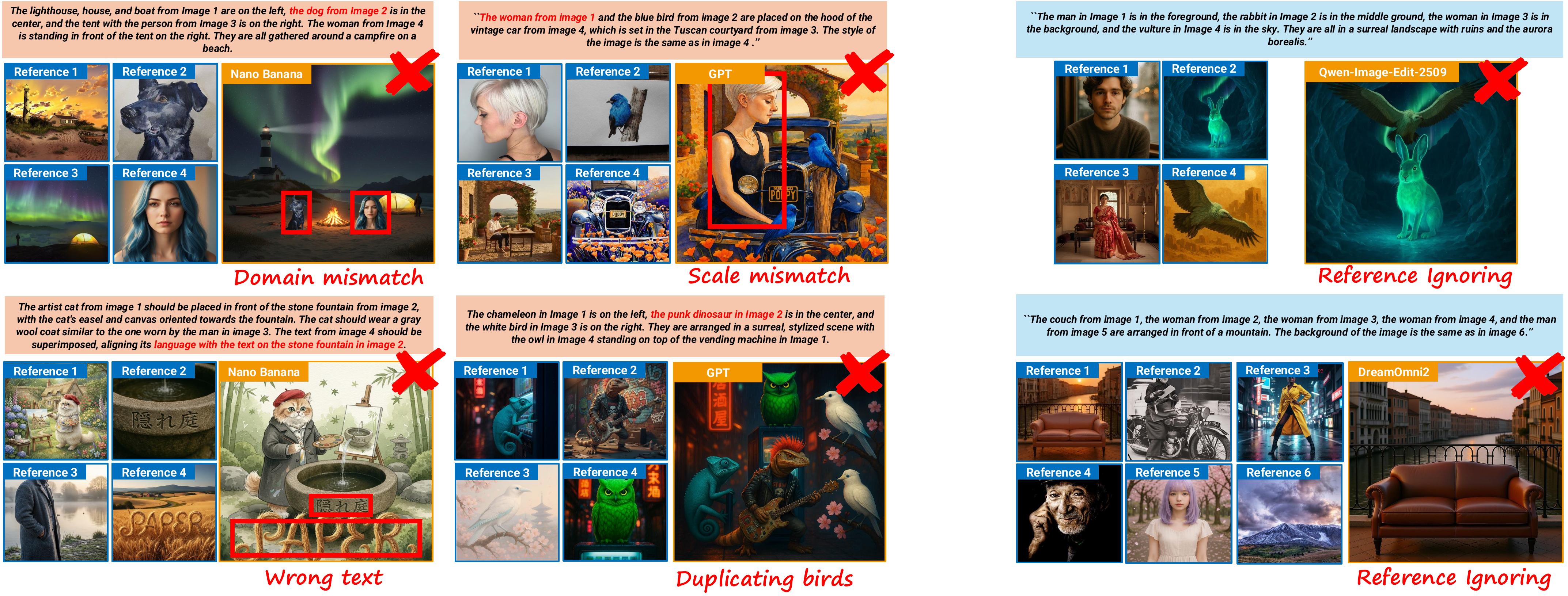}
    \vspace{-6mm}
    \caption{(\textbf{Left}) Failure cases of closed-source models under challenging conditions such as cross-domain references, scale mismatches, rare concepts, and multilingual text. (\textbf{Right}) Failure cases of open-source models under many-reference conditions. They tend to ignore some reference subjects entirely.}
    \label{fig:qualitative}
    \vspace{-3.5mm}
\end{figure*}

\subsection{Per-Task Evaluation}
We compute the MultiBanana scores for each task type defined in Section~\ref{sec:task_types}, and present the results in \autoref{fig:result_main}.
For the two-reference tasks, we report the average score across all 11 task variants.
For the multi-reference tasks, we compute the average score for each task type (i.e., X Objects, X--1 Objects + Local, X--1 Objects + Global, X--1 Objects + Background) across $X=\{3,4,5,6,7,8\}$.

Both open-source and closed-source models exhibit lower performance on background-replacement tasks, regardless of the number of reference images.
The results also reveal clear performance gaps between closed-source and open-source models, even in relatively simple editing scenarios, such as single- and two-reference tasks.

% Furthermore, differences among closed-source models are evident.
% For example, NanoBanana struggles with style transformation, whereas ChatGPT-5 demonstrates strong performance in such tasks, indicating distinct strengths and weaknesses across models.

\subsection{Effect of the Number of References}
We investigate how the MultiBanana score changes when varying the number of reference images $X = \{3,4,5,6,7,8\}$ for each multi-reference generation task (\autoref{fig:results_num_references}). 
Both open-source and closed-source models exhibit a general trend of decreasing Total Score with increasing number of references. However, the nature of this degradation differs between the two model families.

Closed-source models demonstrate strong adherence to reference attributes and prompt instructions. As a result, even when the number of references grows, the decline in Instruction Alignment and Reference Consistency remains relatively mild. 
Nevertheless, because these models attempt to strictly satisfy all reference constraints, increasing the number of subjects often degrades overall visual quality, including overcrowded scenes and compositional inconsistencies (see \autoref{fig:teaser} and \autoref{fig:qualitative}; left).

In contrast, open-source models exhibit weaker adherence to prompts and references; even under the 3-reference setting, both Instruction Alignment and Reference Consistency remain low, and these scores deteriorate further as the number of references increases. Qualitative inspection also reveals that open models frequently ignore multiple subjects under high-reference conditions. However, their overall quality (background consistency, physical realism, visual quality) remains relatively stable, and they tend not to suffer from the severe compositional collapse observed in closed models (see \autoref{fig:qualitative}; right).

In summary, closed-source models succeed in incorporating all required subjects but often generate globally inconsistent or distorted scenes due to overfitting to fine-grained reference details. Open-source models, by contrast, tend to omit multiple reference subjects in complex scenarios but generate visually clean images with fewer structural failures. These observations highlight a clear trade-off between strict reference fidelity and holistic image coherence.

\begin{figure}
    \centering
    \includegraphics[width=\linewidth]{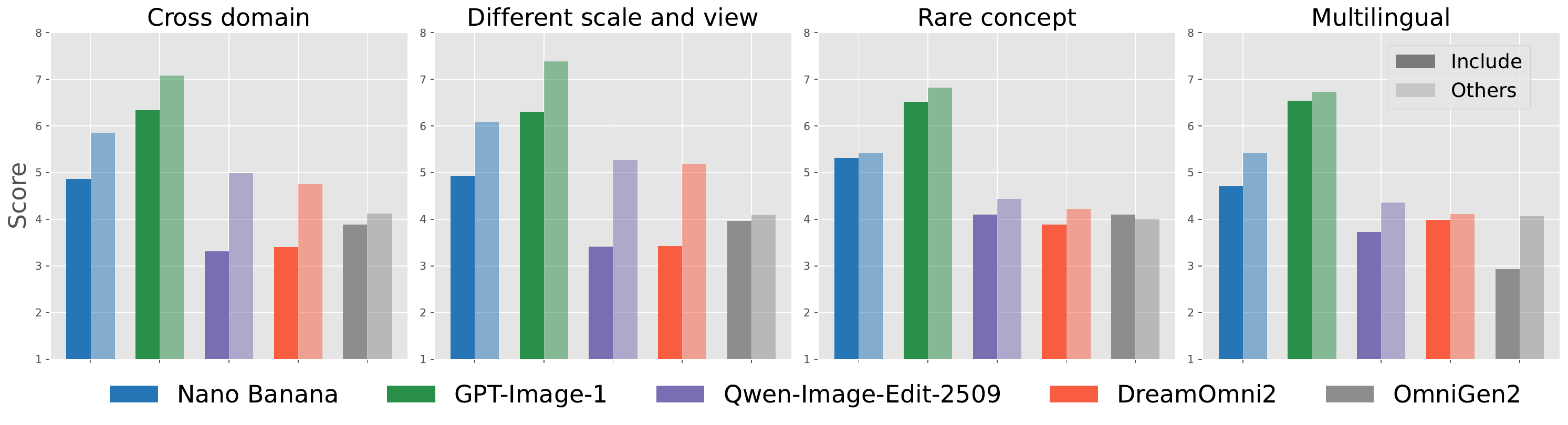}
    \vspace{-6.5mm}
    \caption{Results for difficult reference combinations.
    For cross-domain and different scale and view tasks, every model shows lower scores than tasks without such conditions.}
    \label{fig:results_difficult_combination}
    \vspace{-3.5mm}
\end{figure}

\subsection{Analysis of Difficult Reference Combinations}
To analyze the challenging reference combinations described in Section~\ref{sec:task_types}, we compare MultiBanana scores between tasks containing these difficult combinations and those that do not, as shown in \autoref{fig:results_difficult_combination}. 
For cross-domain, different scale and view, rare-concept, and multilingual tasks, most models achieve lower scores than on tasks without such conditions. 
These results indicate that the difficult reference concepts focused on by our benchmark remain challenging, even for state-of-the-art closed models. 
This demonstrates that our benchmark effectively uncovers the current limitations of multi-reference image generation.
% In contrast, for multilingual tasks, we observe that some models can achieve performance comparable to, or even exceeding, their results on non-multilingual tasks.

% Daiki
% \noindent\textbf{Multi References.}
% First, we verify that multi-reference editing remains challenging in our benchmark, even for state-of-the-art models.
% We evaluated multiple image editing models, including Nano Banana, on tasks that contain three or more reference images.

% \noindent\textbf{Two References.}
% Next, we further demonstrate, on two-reference tasks, that editing remains challenging for recent models when the reference images originate from different domains, vary in object scale, or include rare concept attributes.
% \autoref{tab:results_two_ref} shows the evaluation results of each model on each task with two reference tasks.

% \begin{table*}[t]
% \caption{
% Comparison of the two references' editing results.
% The scores were assigned by GPT-5, which rated each sample on a 5-point scale.
% }
% \centering
% \scalebox{0.9}{%
% \begin{tabular}{lccccccccccc}
% \toprule
% Models & Add & Replace & Background & Color & Material & Action & Hair & Makeup & Tone & Style & Text \\
% \midrule
% GPT-Image-1 & & & & & & & & & & & \\
% % \midrule
% Nano Banana & 3.64 & - & 3.24 & 2.98 & - & 2.96 & 3.27 & 2.87 & - & - & - \\
% \midrule
% Qwen-Image & \\
% OmniGen2 & \\
% DreamOmni2 & \\
% \bottomrule
% \end{tabular}%
% }
% \label{tab:results_two_ref}
% \end{table*}

\subsection{Reliability of VLM Judge}
\label{sec:vlm-human correlation}
To assess the reliability of VLM-based evaluation, we measured the correlation between VLM and human ratings using three human evaluators on 32 samples from our benchmark (\autoref{tab:vlm_human_correlation}).
Both GPT-5 and Gemini 2.5 show strong correlation with human judgments.
This validates the use of VLM judges as a time- and cost-efficient proxy for human evaluation, as described in Section~\ref{sec:evaluation_setting}.
We also confirmed that Qwen3-VL-8B-Instruct~\citep{bai2025qwen3vl} is a reliable open-source alternative, thereby broadening accessibility for researchers without API access.
See Appendix~\ref{sec:ai_human_eval} for further discussion.

% As described in Section~\ref{sec:evaluation_setting}, we employ VLM judges as a time- and cost-efficient proxy for human evaluation. 
% We measured the correlation between VLM and human ratings using 3 evaluators on 32 samples from our benchmark (\autoref{tab:vlm_human_correlation}), finding strong alignment that validates our VLM-based evaluation. 
% We also confirmed that Qwen3-VL-8B-Instruct~\citep{bai2025qwen3vl} serves as a reliable open-source alternative, broadening accessibility for researchers without API access to closed models. 
% See Appendix~\ref{sec:ai_human_eval} for further discussion.

\subsection{Agentic Framework} %kohsei
To address the performance limitations on our challenging benchmark, we further construct three agentic framework baselines: Iterative Prompt Refinement (IPR), Context-Aware Feedback Generation (CAFG), and Selective Reference Adaptation (SRA). We then conduct experiments on our MultiBanana benchmark. We show the results of IPR in \autoref{fig:result_main}, where the generator creates images, and the planner rewrites the prompt over multiple steps based on these images. While GPT's performance improved with the agent framework, Nano Banana showed no improvement and, in some cases, performance degraded due to information loss from the initial instruction. For details on the problem setting, ablations, and results, see Appendix~\ref{sec:agent}.

\begin{table}[t]
\centering
\caption{Correlation between human and VLM judges.
% on a subset of the benchmark, which consists of 32 data points.
% Human scores are obtained by averaging the ratings from three human evaluators.
% All correlations are statistically significant than $0$ with $p<0.01$.
}
\label{tab:vlm_human_correlation}
% \small
\vskip -0.1in
\scalebox{0.95}{
\begin{tabular}{lccc}
\toprule
\textbf{Judge} & \textbf{Pearson $r$} & \textbf{Spearman $r$} & \textbf{Cohen's $\kappa$} \\
\midrule
GPT-5      & 0.6936 & 0.7134 & 0.6135 \\
Gemini 2.5 & 0.5726 & 0.6113 & 0.3908 \\
Qwen3-VL   & 0.5457 & 0.5642 & 0.5235 \\
\bottomrule
\end{tabular}
}
\vspace{-3mm}
\end{table}

\section{Conclusion}
\label{sec:conclusion}
We constructed a comprehensive benchmark for multi-reference image generation, focusing on challenges specific to this setting, such as difficult reference combinations. Using it, we evaluated state-of-the-art open and closed models and found two notable failure modes as the number of references increased: models either ignored editing instructions or collapsed when trying to follow them.
Across different types of reference difficulty, tasks involving cross-domain inputs, large-scale or viewpoint gaps, rare concepts, and multilingual inputs are consistently challenging. 
% In contrast, performance on multilingual tasks depends primarily on whether the underlying model has multilingual capabilities.
% First, we constructed a comprehensive and multidimensional benchmark for evaluating multi-reference image generation, offering broader coverage across diverse conditions.
% Using this benchmark, we evaluated and analyzed the performance of state-of-the-art multi-reference image generation models.
% Our findings reveal that these models perform well with a small number of references but degrade as the number of references increases.
% (1) While agent-based iterative strategies can partially mitigate this degradation, they are ineffective for abilities that the base model inherently lacks.
% (2) Moreover, there exists a trade-off between reference fidelity and instruction fidelity—models that closely follow reference appearances tend to be less responsive to editing instructions, and vice versa.

\section*{Acknowledgements}
We thank Minsu Kim and Heiga Zen for their support with this work and for their review of the initial version of the paper. 
We also appreciate the funding support from Google Japan. MS was supported by JSPS KAKENHI Grant Number JP23H04974.

% \clearpage
{
    \small
    \bibliographystyle{ieeenat_fullname}
    \bibliography{main}
}

\clearpage
\setcounter{page}{1}
\maketitlesupplementary

\renewcommand{\thesection}{\Alph{section}}
\renewcommand{\thesubsection}{\Alph{section}.\arabic{subsection}}
\setcounter{section}{0}

% \section*{Appendix}

\section{Results of Qwen3-VL Judge}
\label{sec:qwen3_vl_judge}

\begin{table}[b]
\makebox[\textwidth]{%
\begin{minipage}{\textwidth}
\captionof{table}{
Per-task total scores for each single- and two-reference task, evaluated by Qwen3-VL.
``Back.'' denotes the Background task.
Both open-source and closed-source models exhibit lower performance on background replacement tasks.
Note that for Nano Banana and GPT-Image-1, we use the versions available as of January 2026.}
\label{tab:single_qwen}
\end{minipage}
}
\centering
\scalebox{0.9}{
\begin{tabular}{lcccccccccccc}
\toprule
\textbf{Model} & \textbf{Single} & \textbf{Add} & \textbf{Back.} & \textbf{Color}
& \textbf{Hair} & \textbf{Makeup} & \textbf{Material} & \textbf{Pose}
& \textbf{Replace} & \textbf{Style} & \textbf{Text} & \textbf{Tone} \\
\midrule
DreamOmni2
& 8.47 & 6.51 & 4.59 & 8.60
& 7.40 & 7.89 & 6.76 & 6.25
& 6.86 & 5.10 & 5.48 & 5.14 \\
OmniGen2
& 9.04 & 5.34 & 6.04 & 5.07
& 6.70 & 6.68 & 6.12 & 5.91
& 6.34 & 3.24 & 4.29 & 4.00 \\
Qwen-Image-Edit-2509
& 9.16 & 7.01 & 4.34 & 6.46
& 8.00 & 6.68 & 5.86 & 5.87
& 7.57 & 3.41 & 4.95 & 6.08 \\
Nano Banana
& 9.45 & 8.13 & 7.97 & 9.67
& 9.16 & 9.25 & 8.95 & 8.76
& 8.30 & 9.03 & 8.81 & 8.52 \\
GPT-image-1
& 9.51 & 8.09 & 8.08 & 9.70
& 9.04 & 9.24 & 9.17 & 8.80
& 8.30 & 9.10 & 8.52 & 8.78 \\
\bottomrule
\end{tabular}
}
\end{table}
\begin{table}[b]
\makebox[\textwidth]{%
\begin{minipage}{\textwidth}
\captionof{table}{
Per-task total scores for each multi-reference task, evaluated by Qwen3-VL.
The local, global, back, and object columns under X correspond to X--1 Objects + Local, X--1 Objects + Global, X--1 Objects + Background, and X Object, respectively.
Both open-source and closed-source models exhibit a general trend of decreasing scores across all tasks as the number of references increases.
They also tend to perform worse on background replacement tasks, especially as the number of reference images increases.
Note that for Nano Banana and GPT-Image-1, we use the versions available as of January 2026.
}
\label{tab:multi_qwen}
\end{minipage}
}
\centering
\scalebox{0.96}{
\begin{tabular}{lcccccccccccc}
\toprule
\textbf{Model} &
\multicolumn{4}{c}{\textbf{3-references}} &
\multicolumn{4}{c}{\textbf{4-references}} &
\multicolumn{4}{c}{\textbf{5-references}} \\
\cmidrule(lr){2-5}
\cmidrule(lr){6-9}
\cmidrule(lr){10-13}
& local & global & back & object
& local & global & back & object
& local & global & back & object \\
\midrule
DreamOmni2
& 4.39 & 4.27 & 4.33 & 3.68
& 3.80 & 2.95 & 2.77 & 2.90
& 3.23 & 2.12 & 2.86 & 2.26 \\
OmniGen2
& 4.71 & 4.86 & 4.39 & 4.13
& 4.37 & 3.86 & 3.43 & 2.95
& 3.99 & 2.75 & 2.14 & 2.26 \\
Qwen-Image-Edit-2509
& 5.22 & 5.81 & 5.81 & 5.63
& 3.24 & 4.51 & 4.42 & 3.94
& 1.05 & 1.48 & 1.28 & 1.10 \\
Nano Banana
& 7.63 & 7.21 & 6.96 & 6.91
& 7.96 & 6.77 & 6.47 & 6.78
& 7.96 & 6.54 & 6.03 & 5.87 \\
GPT-image-1
& 7.83 & 7.26 & 6.99 & 6.72
& 7.73 & 6.74 & 5.94 & 6.83
& 7.92 & 6.54 & 6.16 & 5.83 \\
\bottomrule
\toprule
\textbf{Model} &
\multicolumn{4}{c}{\textbf{6-references}} &
\multicolumn{4}{c}{\textbf{7-references}} &
\multicolumn{4}{c}{\textbf{8-references}} \\
\cmidrule(lr){2-5}
\cmidrule(lr){6-9}
\cmidrule(lr){10-13}
& local & global & back & object
& local & global & back & object
& local & global & back & object \\
\midrule
DreamOmni2
& 3.34 & 2.26 & 1.72 & 1.93
& 2.61 & 2.45 & 2.10 & 1.81
& 2.44 & 1.65 & 1.83 & 1.59 \\
OmniGen2
& - & - & - & -
& - & - & - & -
& - & - & - & - \\
Qwen-Image-Edit-2509
& 1.00 & 1.49 & 1.00 & 1.00
& 1.00 & 1.46 & 1.00 & 1.13
& 1.00 & 1.00 & 1.00 & 1.00 \\
Nano Banana
& 6.84 & 6.08 & 5.34 & 5.64
& 6.37 & 5.78 & 5.13 & 5.44
& 6.46 & 6.24 & 5.31 & 5.28 \\
GPT-image-1
& 7.11 & 5.86 & 5.51 & 5.92
& 6.81 & 5.78 & 5.37 & 5.74
& 7.07 & 5.86 & 5.14 & 5.55 \\
\bottomrule
\end{tabular}
}
\end{table}

In Section~\ref{sec:experiments}, we report the average scores from Gemini 2.5 and GPT-5 as judges.
As demonstrated in Section~\ref{sec:vlm-human correlation}, Qwen3-VL-8B-Instruct~\citep{bai2025qwen3vl} achieves strong correlation with human judgments comparable to GPT-5~\citep{openai2025gpt5} and Gemini 2.5~\citep{geminiteam2023gemini}, confirming its reliability as a judge model (see \autoref{tab:vlm_human_correlation}).
Although we used stable API versions for the closed-source judges, to ensure full reproducibility, we additionally employ Qwen3-VL-8B-Instruct, an open-weight, fixed-version model, as the judge.
This also broadens access to our benchmark for researchers who do not have access to closed-source VLMs via APIs.

\autoref{tab:single_qwen} and \autoref{tab:multi_qwen} present per-task total scores evaluated using Qwen3-VL as the sole judge, calculated in the same way as described in Section~\ref{sec:evaluation_setting}.
The results are broadly consistent with those obtained using closed-model judges. The overall ranking of image generation models is preserved.
We also observe the same key trends: performance degrades as the number of reference images increases, and background replacement tasks tend to yield lower scores across both open-source and closed-source models.
These observations reinforce the validity of Qwen3-VL as a reliable, reproducible alternative.
We expect this fixed, open-weight judge to serve as a dependable baseline for future evaluations on our benchmark.

\clearpage
\section{Details on Benchmark Construction}
\label{sec:benchmark_construction_details}

\subsection{Image Collection}
We constructed the reference image collection for our benchmark using both real images sourced from LAION-5B~\citep{schuhmann2022laion} and synthetic images generated by Nano Banana~\citep{google2025nanobanana} and GPT-Image-1~\citep{openai2025gpt4oimage}.
By combining real and synthetic data, we mitigated the categorical biases inherent in the real-image distribution while expanding the coverage and diversity of the reference set.
To generate synthetic images, we designed distinct axes of variation for four major categories: humans, animals, objects, and text-containing images.

\noindent\textbf{Human Category.}~~We systematically combined attributes such as makeup, hairstyle, hair color, pose, strong facial expressions, lighting, and overall tonal style to create a wide range of appearance conditions. 
Examples include ethnic or fashion-model-style makeup, vivid or unconventional hair colors, model-like or manga-like poses, strong emotional expressions (e.g., anger, joy, surprise), dramatic lighting (e.g., window light through blinds or neon illumination), and global color palettes (e.g., sepia or bluish). 
These conditions were intentionally selected to supplement the variations underrepresented in real-world data.

\noindent\textbf{Animal Category.}~~We first specified major species groups (dogs, cats, birds, reptiles, wild animals, and fantasy animals). 
We then added attributes to enhance the diversity of animal imagery. 
These included human-like behaviors (e.g., a bear reading a book, a fox playing in a rock band), rare colors (e.g., a blue horse or a pink panda), distinctive lighting conditions, and varied global tones such as sepia. This incorporated rare or extreme examples that seldom appear in real datasets.

\noindent\textbf{Object Category.}~~We expanded diversity along two axes: intrinsic object attributes and photographic conditions. The former included distinctive textures (e.g., denim, leather, knit), characteristic prints (e.g., floral, striped, checkered), and variations in material or design (e.g., matte, glossy, transparent). The latter included variations in lighting and global tonal adjustments (e.g., sepia, cool color tones), thereby ensuring a broad range of appearance conditions for objects.

\noindent\textbf{Images Containing Text.}~~Images containing text, which were largely absent from real data, were generated exclusively using GPT-Image-1, as Nano Banana exhibited limited ability to render non-English text. We defined three levels of text length (a single word, a short phrase of up to five words, and a longer sentence). We combined each with two scene types: text rendered directly within a landscape (e.g., ocean, forest) and text appearing naturally in the environment (e.g., on signs in real scenes). These six prompt patterns were generated in English and subsequently translated into Japanese and Chinese to enrich the multilingual text in images.

\noindent\textbf{Scene Diversity.}~~Finally, all synthetic prompts were designed to incorporate scene diversity, including European cityscapes, university campuses, tropical beaches, event halls, and other varied environments. The generated subjects were well-framed, resulting in images suitable for use as high-quality reference material.

\begin{figure}[ht]
  \centering
  \includegraphics[width=0.73\linewidth]{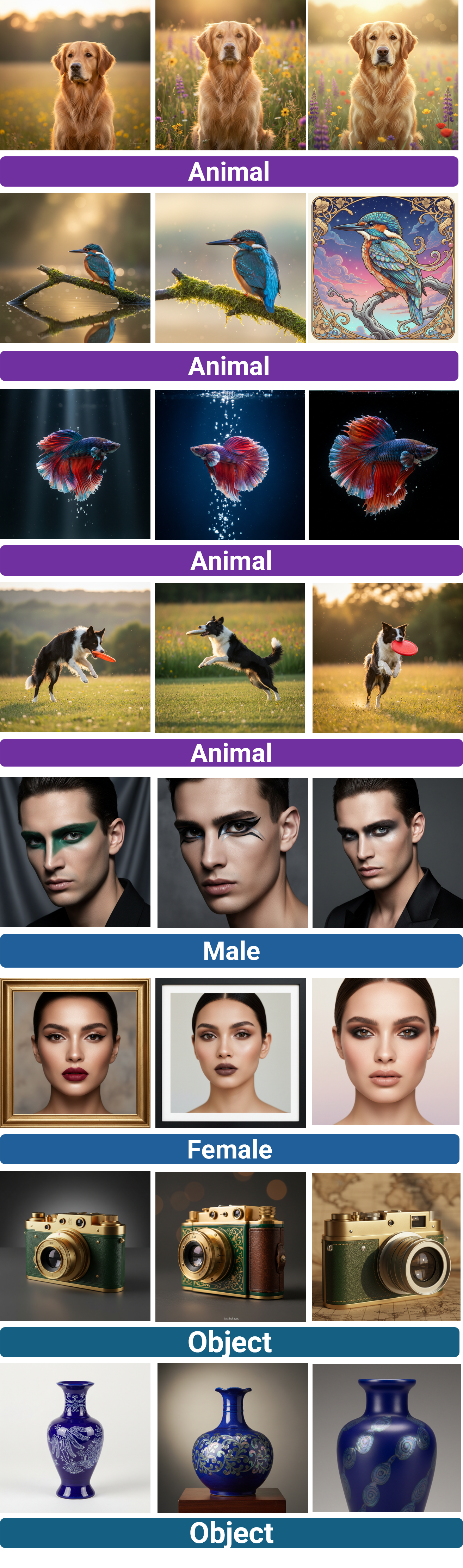}
  \caption{Examples of a duplicated synthetic image.}
  \label{fig:example_duplicated}
\end{figure}

\subsection{Image Filtering}
First, to remove images unsuitable for editing, we filtered them based on the size of the foreground objects, as in \citep{ye2025imgedit}.
YOLOv12~\citep{tian2025yolov12} was applied to all images to detect objects and then performed semantic segmentation using SAM~\citep{ravi2025sam}, conditioned on the detected bounding boxes.
Images were discarded if they met any of the following criteria: the bounding box area covered less than 2\% of the entire image, the segmented region within the bounding box covered less than 30\%,
or the CLIP similarity~\citep{ma2022xclip} between the YOLO-predicted class name and the bounding box region was below 20.
These cases were judged to contain objects that were not clearly visible.
However, images where no objects were detected were retained, assuming they are useful for background or style-level editing tasks.
After this filtering, 48\% of the images were retained.

We also filtered out images that were inappropriate or unsuitable as references, such as unsafe content, charts, and screenshots of system messages, using both automated screening based on Gemini and human review.
For inappropriate content, we specifically targeted categories including hate, harassment, violence, self-harm, sexual content, nudity, shocking content, illegal activity, and other distressing material, ultimately excluding approximately 3\% of the collected images.
We also manually removed nearly identical synthetic data. An example is shown in \autoref{fig:example_duplicated}.

\begin{figure*}[t]
  \centering
  \begin{minipage}[t]{0.60\linewidth}
    \centering
    \includegraphics[width=\linewidth]{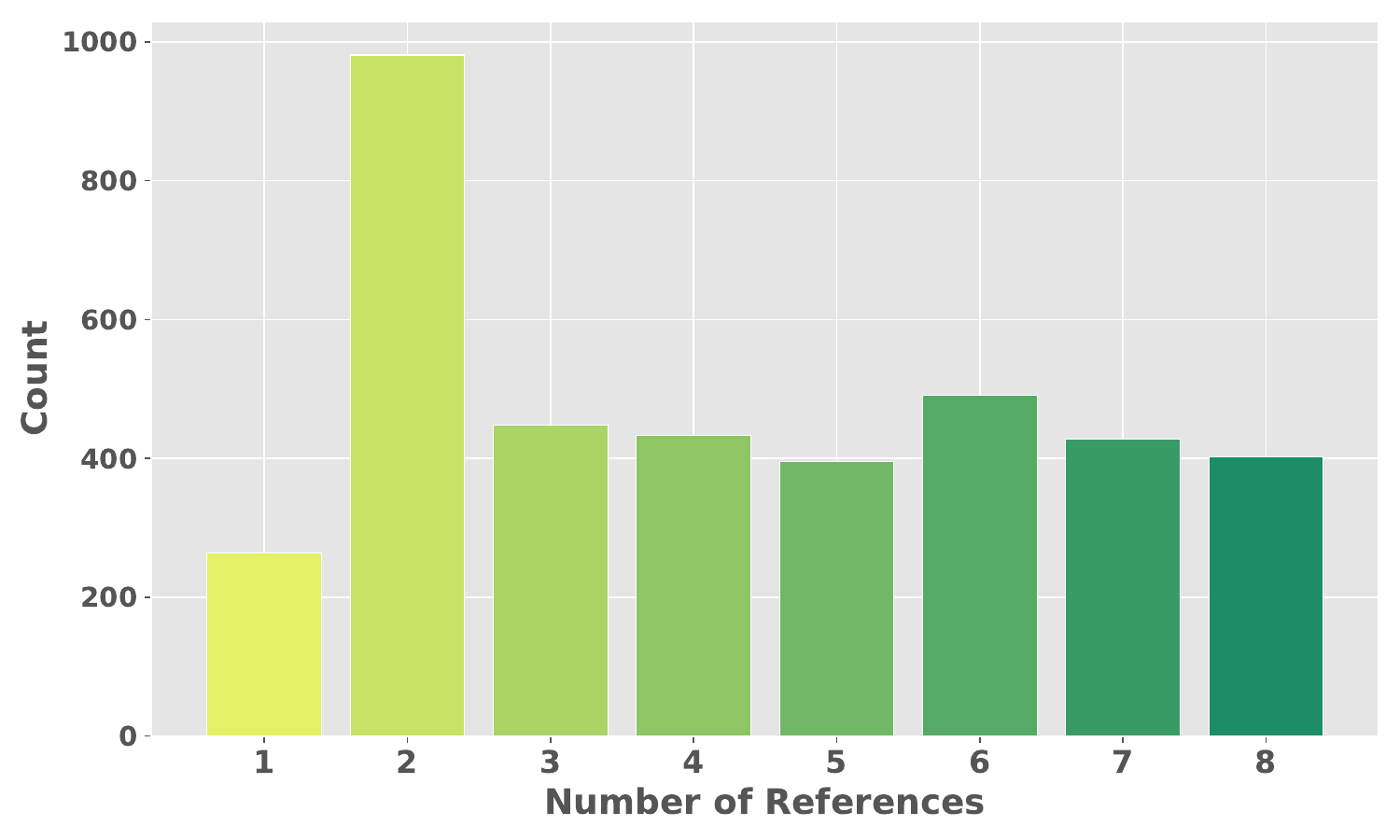}
  \end{minipage}
  \begin{minipage}[t]{0.34\linewidth}
    \centering
    \includegraphics[width=\linewidth]{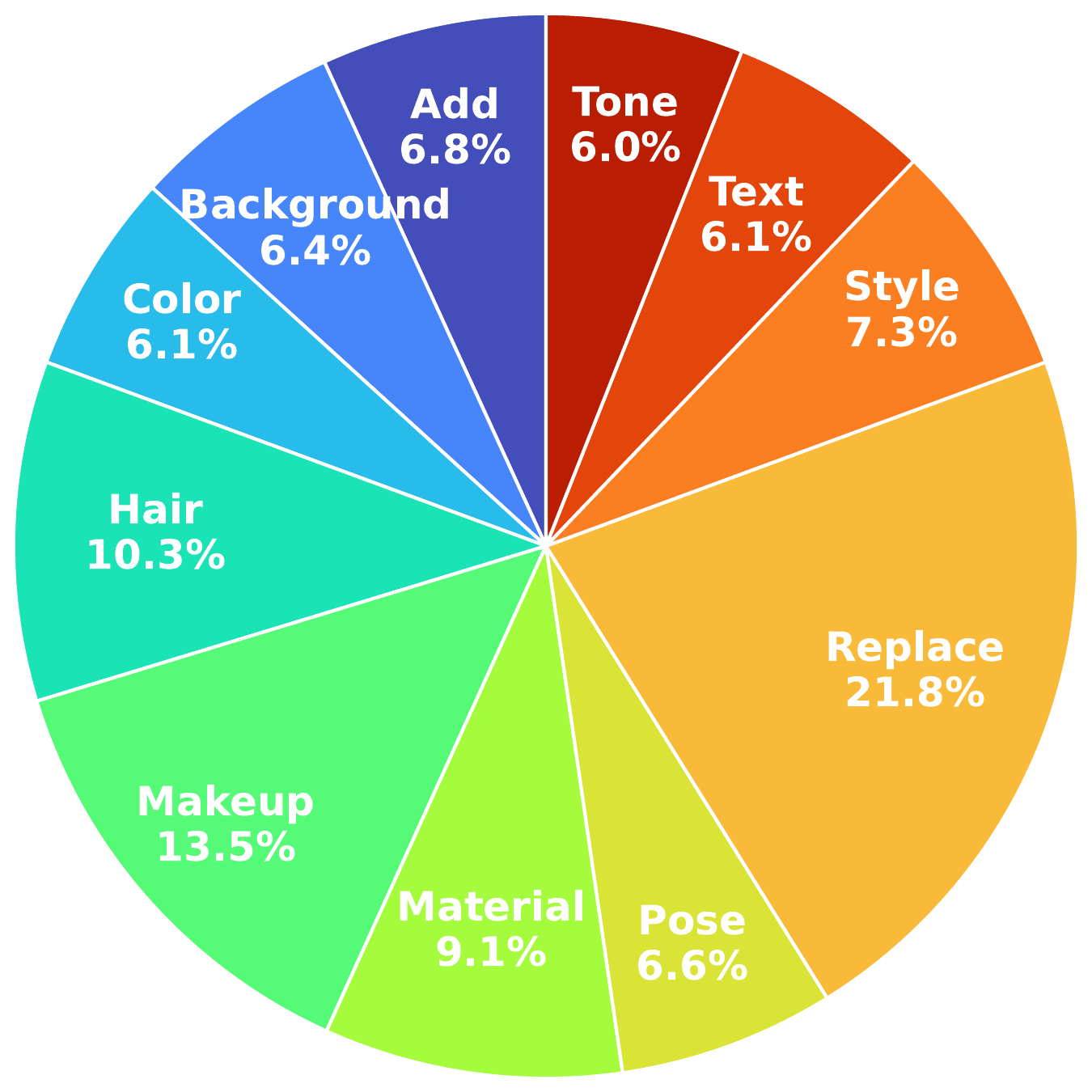}
  \end{minipage}
  \caption{
  (\textbf{Left}) Number of tasks by reference count. The two-reference tasks contain more samples than the other reference tasks because they include multiple task types.
  (\textbf{Right}) Breakdown of the two-reference tasks. It contains eleven tasks.
  }
  \label{fig:data_distribution_2ref}

  \vspace{1.5em}
  
  \includegraphics[width=\textwidth]{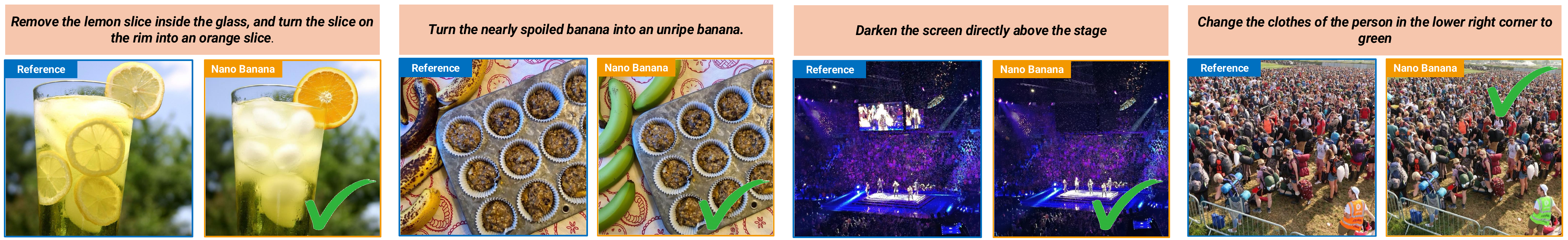}
  \caption{Example of ``Hard'' category of ImgEdit. These tasks are almost fully solvable by advanced models such as Nano Banana.}
  \label{fig:example_imgedit}
  
  \vspace{1.0em}
  
  \includegraphics[width=\textwidth]{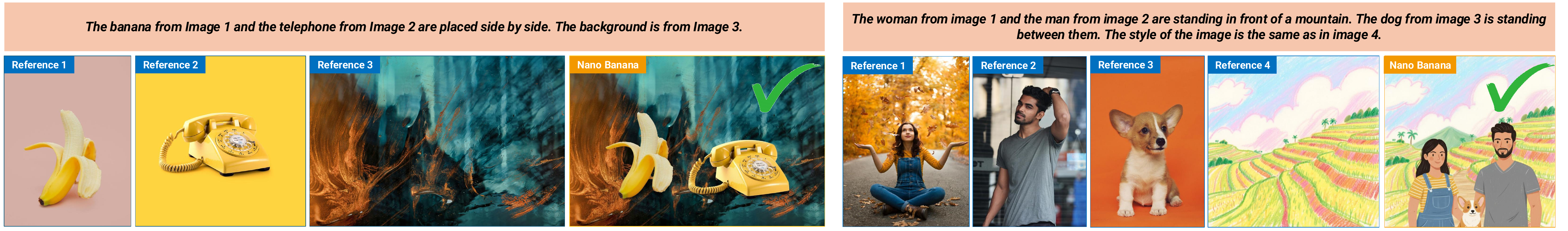}
  \caption{Example of DreamOmni2 benchmark. These tasks are almost fully solvable by advanced models such as Nano Banana.}
  \label{fig:example_dreamomni2}

\end{figure*}

\subsection{Category Classification}

To construct multi-reference image editing tasks, each reference image had to be accurately categorized. For example, in portrait transformation tasks that modify a person's hairstyle or makeup, both reference images must contain a person. Similarly, in replacement tasks involving humans or objects, the corresponding entities must be present in the reference images, and in background replacement tasks, the background scenes must be identified in the reference images. To satisfy these task-dependent requirements, we performed hierarchical image classification on a large combined set of real and synthetic images.

For synthetic data, attributes can be specified directly in the prompt during generation, eliminating the need for additional classification. In contrast, real images exhibit substantial variability, requiring precise categorization to determine whether they can serve as valid reference images. To address this, we designed a structured annotation protocol using Gemini that consists of three classification processes for the human, object, and background components of an image.

\noindent\textbf{Human Image Classification}~~Human classification begins with identifying how many people appear in the image. 
If there is precisely one person, the model additionally classifies attributes such as gender, makeup, hairstyle distinctiveness, pose clarity, and facial expression characteristics. The output follows a fixed template, enabling consistent extraction of appearance features relevant to multi-reference generation.

\noindent\textbf{Object Image Classification}~~Object image classification follows a similar structure.
The model first determines whether the image contains no object, a single object, or multiple objects. 
When exactly one object is detected, it is further categorized according to its visual and material properties, distinguishing among animal-like forms, patterned or decorative textiles, engineered or manufactured materials, and objects lacking distinctive design features.

\noindent\textbf{Background Image Classification}~~Background classification is conducted from three viewpoints: realism, tonal uniformity, and lighting strength. Realism distinguishes photographic scenes from stylized or synthetic ones; tone indicates whether the overall color distribution is uniform or varied; and lighting strength identifies whether a clear directional key light is present.

\subsection{Task Construction}
The editing instructions were generated using Gemini. 
First, we manually selected suitable reference categories for each task, for instance, the \textit{person} category for the hairstyle modification task.
Next, we randomly sampled images from each category and presented them to Gemini to produce corresponding editing instructions.
For example, in editing tasks that require object transformations, we first prompted the model to determine which reference image’s object should be modified and which reference image's attributes (e.g., color) should be used as the target.
Afterward, several instruction examples were provided to Gemini, and we asked Gemini to generate new instructions following the same syntax.
Afterward, editing instructions that could cause visual breakdowns were removed after being scored with Gemini.
Additionally, Gemini assessed the difficulty of editing instructions, and those judged easy, such as cases without domain differences, were removed.
Finally, the editing instructions were manually verified not to cause image breakdowns or result in overly simple edits,
and assigned the remaining samples to the appropriate difficulty categories (cross-domain, scale and viewpoint differences, rare concept, and multilingual text rendering).

\section{Further Statistics for MultiBanana}
\label{sec:further_statistics}
The left of \autoref{fig:data_distribution_2ref} shows the number of tasks by each reference count. 
The two-reference tasks contain more samples than the other tasks because they include multiple task types.
The right of \autoref{fig:data_distribution_2ref} shows the breakdown of the two-reference tasks.
It contains eleven tasks considered in the prior work~\citep{xia2025dreamomni2}: subject addition, subject replacement, background change, color modification, material modification, pose modification, hairstyle modification, makeup modification, tone transformation, style transfer, and text correction.
Each task is guaranteed to include at least 6\% of the editing samples, corresponding to roughly 60 samples, which exceeds the number of samples in \citet{xia2025dreamomni2}.

\begin{table*}[t]
\centering
\caption{
Comparison results of different models on ImgEdit-Bench~\citep{ye2025imgedit}. 
``Overall'' is computed by averaging the scores across all task types. 
GPT-4.1~\citep{openai2023gpt4} is used for evaluation.
The results show that recent closed-source, state-of-the-art models achieve substantially higher scores, suggesting that the benchmark may be approaching its ceiling in distinguishing high-end models.
Evaluation results excluding Nano Banana are taken from the official GitHub repository of \citet{ye2025imgedit}.
}
\label{tab:result_imgedit_bench}
\scalebox{1.0}{
\begin{tabular}{lcccccccc|c}
\toprule
\textbf{Model} &
\textbf{Add} &
\textbf{Adjust} &
\textbf{Extract} &
\textbf{Replace} &
\textbf{Remove} &
\textbf{Background} &
\textbf{Style} &
\textbf{Hybrid} &
\textbf{Overall} \\
\midrule

MagicBrush~\citep{zhang2023magicbrush} 
& 2.84 
& 1.58 
& 1.51 
& 1.97 
& 1.58 
& 1.75 
& 2.38 
& 1.62 
& 1.83 \\

AnyEdit~\citep{yu2025anyedit} 
& 3.18 
& 2.95 
& 1.88 
& 2.47 
& 2.23 
& 2.24 
& 2.85 
& 1.56 
& 2.45 \\

OmniGen2~\citep{wu2025omnigen2} 
& 3.57 
& 3.06 
& 1.77 
& 3.74 
& 3.20 
& 3.57 
& \underline{4.81} 
& 2.52 
& 3.44 \\

Kontext-dev~\citep{labs2025flux1kontext} 
& \underline{3.83} 
& \underline{3.65} 
& \underline{2.27} 
& \underline{4.45} 
& 3.17 
& \underline{3.98} 
& 4.55 
& \underline{3.35} 
& \underline{3.71} \\

\midrule
Nano Banana~\citep{google2025nanobanana}
& 3.63 
& \textbf{4.66} 
& \textbf{3.73} 
& \textbf{4.69} 
& \textbf{4.71} 
& \textbf{4.60} 
& 4.47 
& \textbf{4.10} 
& \textbf{4.37} \\

GPT-Image-1~\citep{openai2025gpt4oimage} 
& \textbf{4.61} 
& \underline{4.33} 
& \underline{2.90} 
& 4.35 
& \underline{3.66} 
& \underline{4.57} 
& \textbf{4.93} 
& \underline{3.96} 
& \underline{4.20} \\

\bottomrule
\end{tabular}
}

\vspace{1.5em}

\centering
\caption{Quantitative comparison of multimodal instruction-based editing and generation in DreamOmni2 Benchmark~\citep{xia2025dreamomni2}. 
Both tasks are evaluated using Gemini~\citep{geminiteam2023gemini} and Doubao~\citep{bytedance2025doubao}.
This result shows that recent closed-source, state-of-the-art models achieve substantially higher scores, suggesting that the benchmark may be approaching its ceiling in distinguishing among high-end models.
We refer to evaluation results from \citet{xia2025dreamomni2}.}
\label{tab:result_dreamomni2}
% \renewcommand{\arraystretch}{1.05}
% \scalebox{1.0}{
\begin{tabular}{lcccccccc}
\toprule
\multirow{3}{*}{\textbf{Method}} 
& \multicolumn{4}{c}{\textbf{Editing Task}} 
& \multicolumn{4}{c}{\textbf{Generation Task}} \\
\cmidrule(lr){2-5} \cmidrule(lr){6-9}
& \multicolumn{2}{c}{\textbf{Concrete $\uparrow$}} 
& \multicolumn{2}{c}{\textbf{Abstract $\downarrow$}}
& \multicolumn{2}{c}{\textbf{Concrete $\uparrow$}}
& \multicolumn{2}{c}{\textbf{Abstract $\downarrow$}} \\
\cmidrule(lr){2-3} \cmidrule(lr){4-5}
\cmidrule(lr){6-7} \cmidrule(lr){8-9}
& \textbf{Gemini} & \textbf{Doubao} & \textbf{Gemini} & \textbf{Doubao} 
& \textbf{Gemini} & \textbf{Doubao} & \textbf{Gemini} & \textbf{Doubao} \\
\midrule
Omnigen2~\citep{wu2025omnigen2}        
& 0.2195 & 0.2927 & 0.0427 & 0.0793 
& 0.2083 & 0.2500 & 0.1000 & 0.0778 \\
% Qwen-Image-Edit~\citep{wu2025qwenimagetechnicalreport} 
% & 0.0976 & 0.1463 & 0.0244 & 0.0183 
% & 0.0417 & 0.1250 & 0.0889 & 0.1000 \\
Kontext~\citep{labs2025flux1kontext}     
& 0.0488 & 0.1220 & 0.0183 & 0.0122 
& 0.2500 & 0.3750 & 0.0556 & 0.1222 \\
Qwen-Image-Edit-2509~\citep{wu2025qwenimagetechnicalreport}
& 0.2683 & 0.2927 & 0.0488 & 0.1159 
& 0.1250 & 0.2917 & 0.1111 & 0.1556 \\
DreamOmni2~\citep{xia2025dreamomni2}         
& \underline{0.5854} & \underline{0.6585} & 0.5854 & \underline{0.6280}
& \underline{0.5833} & \textbf{0.6667} & \underline{0.5778} & \textbf{0.6333} \\
\midrule
Nano Banana~\citep{google2025nanobanana}        
& \textbf{0.6829} & \underline{0.7073} & \underline{0.6463} & 0.5488 
& 0.5000 & 0.5417 & 0.5556 & \underline{0.5488} \\
GPT-Image-1~\citep{openai2025gpt4oimage}              
& \textbf{0.6829} & \textbf{0.7805} & \textbf{0.7195} & \textbf{0.7439} 
& \textbf{0.6250} & \underline{0.6250} & \textbf{0.6889} & \textbf{0.6333} \\
\bottomrule
\end{tabular}
% }
\end{table*}

\section{Comparison with Prior Benchmarks}
\autoref{tab:main_benchmark_comparison} shows that existing benchmarks do not provide systematic evaluation across diverse multi-reference conditions, support only a limited number of references, and fail to adequately account for heterogeneity among reference images. 
To illustrate that existing benchmarks are almost solved by state-of-the-art models such as Nano Banana~\citep{google2025nanobanana} and GPT-Image-1~\citep{openai2025gpt4oimage}, we evaluate the image generation capabilities of these models using ImgEdit~\citep{ye2025imgedit}, the latest benchmark for image editing, and DreamOmni2~\citep{xia2025dreamomni2}, the state-of-the-art benchmark for multi-reference image generation.
The quantitative results show that recent closed-source, state-of-the-art models achieve consistently high scores, indicating that these benchmarks are nearing saturation and may soon be unable to meaningfully distinguish among high-performing models (\autoref{tab:result_imgedit_bench} and \autoref{tab:result_dreamomni2}).
The evaluation results are taken from the official GitHub repository of ImgEdit\footnote{\url{https://github.com/PKU-YuanGroup/ImgEdit?tab=readme-ov-file}} and the results reported in the DreamOmni2 paper.
Qualitative results suggest that, even in the additional ``Hard'' category of ImgEdit, most tasks are already solvable by advanced models such as Nano Banana (\autoref{fig:example_imgedit}).
A similar tendency is observed for DreamOmni2 benchmark as well (\autoref{fig:example_dreamomni2}).

Taken together, these findings indicate that current benchmarks for image editing and multi-reference image generation are approaching their limits in evaluating cutting-edge models.
There is a clear need for more challenging benchmarks that better capture the capabilities and failure modes of the latest generation of image models.

\section{Reliability and Cost of VLM Judges} \label{sec:ai_human_eval}
As described in Section~\ref{sec:evaluation_setting}, we employ VLM judges as a time- and cost-efficient proxy for human evaluation.
% The correlation between AI and human evaluation has been demonstrated in prior work~\citep{ku2023viescore, na2024boost, wu2024boosting, furuta2024improving, oshima2025inference}.
% Consequently, using VLMs is becoming a standard method for the automatic evaluation of generated images~\citep{ye2025imgedit, wu2025omnigen2, xia2025dreamomni2}.
In Section~\ref{sec:vlm-human correlation}, we demonstrate that the VLM judge correlates strongly with human ratings, supporting our usage of VLM.
Here, we provide a more detailed analysis of their reliability and the associated cost considerations.
% We used VLMs to evaluate the images along five metrics: Instruction Alignment, Reference Consistency, Background Subject Match, Physical Realism, and Visual Quality.
% The overall score is calculated as a weighted average of these metrics.
% The first two metrics represent the fundamental components of reference-based generation, while the last three metrics break down the overall quality.
% Therefore, the overall score is calculated as $\mathrm{avg}(\mathrm{Instruction\,Alignment} + \mathrm{Reference\,Consistency} + \mathrm{avg}(\mathrm{Background\,Subject\,Match} + \mathrm{Physical\,Realism} + \mathrm{Visual\,Quality}))$.

\subsection{Cross-VLM Correlation}
To demonstrate the consistency of our evaluation metrics, we examined correlations among different VLMs.
We use the evaluated results for Nano Banana and GPT-Image-1 and examine the correlation between GPT's and Gemini's evaluation scores.
To assess the reliability of the correlation coefficient, we partitioned the evaluation results into 10 subsets and computed the mean correlation across them.
\autoref{tab:correlation} shows that the evaluated scores exhibit positive correlations, which demonstrates that our evaluation criteria are well-defined and consistent.

%\begin{figure}[ht]
%    \centering
%    \includegraphics[width=0.4\linewidth]{figures/correlation.pdf}
%    \caption{Correlation between GPT's and Gemini's evaluated scores. The top $R$ value represents the correlation coefficient and its 95\% confidence interval.}
%    \label{fig:correlation}
%\end{figure}
\begin{table}[t]
\centering
\caption{Correlation between GPT's and Gemini's evaluated scores. $\pm$ shows 95\% confidence interval.}
\label{tab:correlation}
\begin{tabular}{c c}
\toprule
\textbf{Evaluation criteria} & \textbf{Correlation coefficients} \\
\midrule
Instruction Alignment & 0.645 $\pm\,0.021$ \\
Reference Consistency & 0.549 $\pm\,0.026$ \\
Background Subject Match & 0.601 $\pm\,0.020$ \\
Physical Realism & 0.620 $\pm\,0.022$ \\
Visual Quality & 0.577 $\pm\,0.018$ \\
\midrule
Overall & 0.650 $\pm\,0.014$ \\
\bottomrule
\end{tabular}
\end{table}

\subsection{Self-Consistency Analysis}
We measure the standard error of the VLM judge scores across three different random seeds in \autoref{tab:score_standard_error}.
On the 1–10 rating scale, the observed standard errors are below 0.1, and we do not observe an increasing trend in standard error as the number of references increases, indicating that the VLM judge provides self-consistent evaluations for multi-reference image generation.

\begin{table}[t]
% \vskip -0.05in
\centering
\caption{Standard error of total scores evaluated by the average of GPT and Gemini, for GPT-Image-1 and Nano Banana.}
\label{tab:score_standard_error}
\scalebox{0.85}{
\begin{tabular}{lcccccc}
\toprule
\textbf{Model} & \textbf{3-ref} & \textbf{4-ref} & \textbf{5-ref} & \textbf{6-ref} & \textbf{7-ref} & \textbf{8-ref} \\
\midrule
GPT-Image-1 & $.0234$ & $.0138$ & $.0139$ & $.0153$ & $.0343$ & $.0244$  \\
Nano Banana & $.0260$ & $.0423$ & $.0033$ & $.0089$ & $.0062$ & $.0254$ \\
\bottomrule
\end{tabular}
}
\end{table}

\subsection{Sensitivity Analysis}
To further ensure the reliability of our evaluation metrics, we examined whether they can detect performance gains from fine-tuning on external image-editing datasets, thereby demonstrating that they align well with those used in other image-editing benchmarks.
We compare Qwen-Image-Edit with Qwen-Image-Edit-2509, a fine-tuned variant designed to improve reference consistency and support multi-reference inputs.
As shown in \autoref{tab:qwen_edit}, our evaluation metrics successfully capture the performance gains of Qwen-Image-Edit-2509 over Qwen-Image-Edit.
We conducted this comparison using single-reference tasks because Qwen-Image-Edit supports only a single input image.

\subsection{Cost of VLM Judge}
Regarding evaluation costs, a single model evaluation costs approximately \$44 with GPT-5 and \$29 with Gemini 2.5 in January 2026. 
As an API-free alternative, we confirmed that Qwen3-VL~\citep{bai2025qwen3vl} can be reliable enough (see \autoref{tab:vlm_human_correlation}).

\section{Detailed Results}
\label{sec:further_results}

\subsection{Per-Task Evaluation}
\label{sec:further_pertask}
\begin{figure*}[ht]
    \centering
    \includegraphics[width=0.98\textwidth]{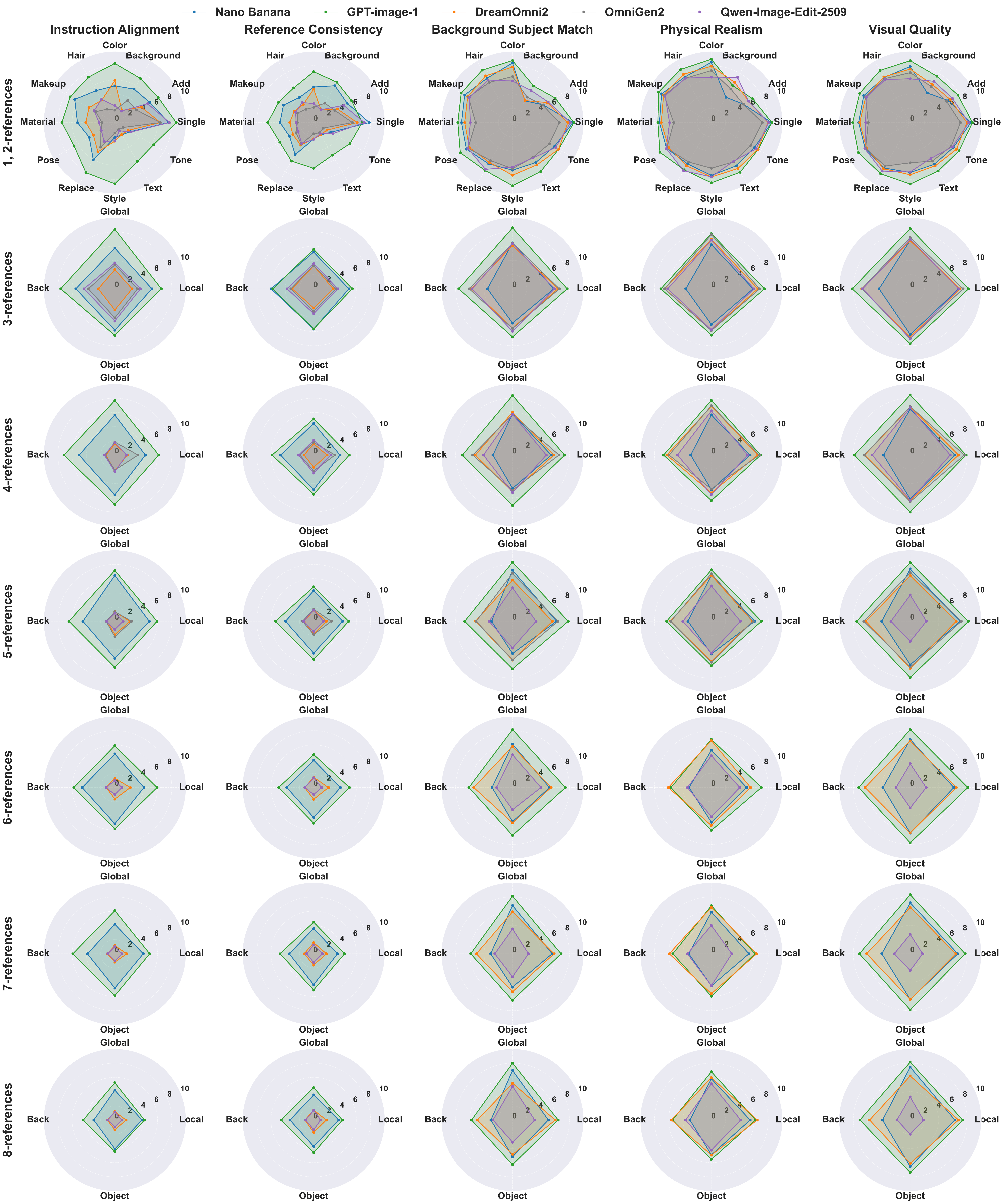}
    \caption{Scores of each model across evaluation metrics for each task by GPT. The horizontal axis denotes the scores for the five evaluation criteria, and the vertical axis denotes the number of reference images.}
    \label{fig:scores_gpt_pertask}
\end{figure*}

\begin{figure*}[ht]
    \centering
    \includegraphics[width=0.98\textwidth]{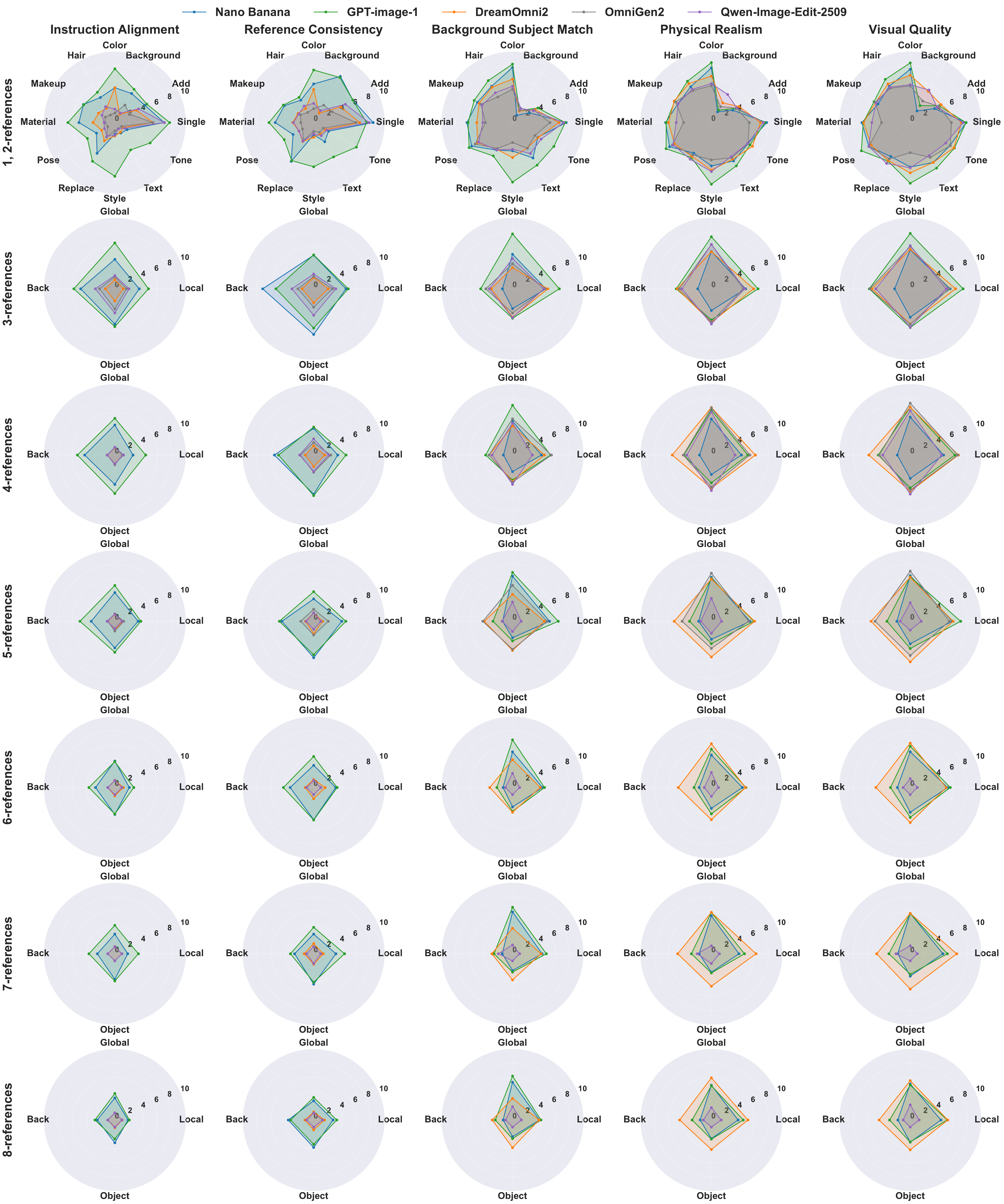}
    \caption{Scores of each model across evaluation metrics for each task by Gemini. The horizontal axis denotes the scores for the five evaluation criteria, and the vertical axis denotes the number of reference images.}
    \label{fig:scores_gemini_pertask}
\end{figure*}

\begin{figure*}[ht]
    \centering
    \includegraphics[width=0.95\linewidth]{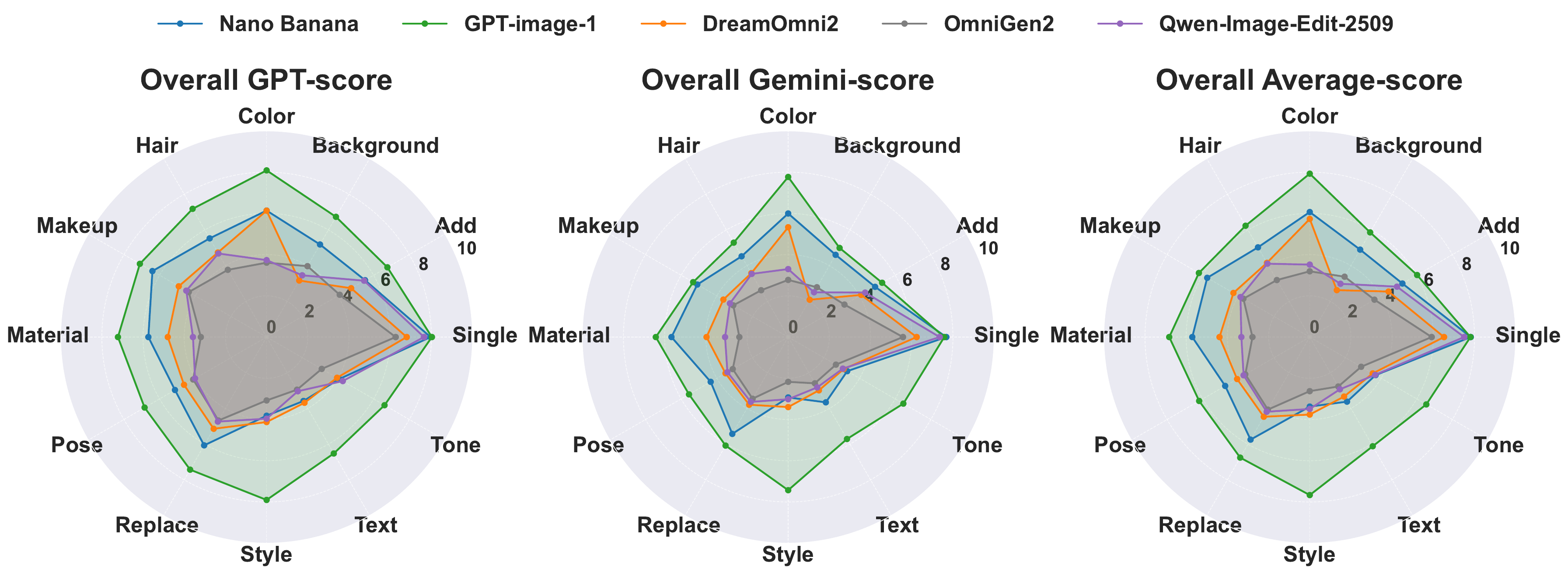}
    \caption{Total scores of each model for single and two-reference tasks by GPT, Gemini, and their average.}
    \label{fig:allscore_pertask}
\end{figure*}

\autoref{fig:scores_gpt_pertask}, \autoref{fig:scores_gemini_pertask}, and \autoref{fig:scores_average_pertask} show the scores of each model across five evaluation metrics for each task, evaluated by GPT, Gemini, and their average, respectively.
Taking the weighted average across five metrics,
\autoref{fig:allscore_pertask} shows the total score of each model for single and two-reference tasks, evaluated by GPT, Gemini, and their average, respectively.
For multi-reference tasks,
\autoref{tab:multi_res_details_gpt}, \autoref{tab:multi_res_detail_gemini}, and \autoref{tab:multi_res_detail_all} show the total score, evaluated by GPT, Gemini, and their average, respectively.

Overall, we observed a substantial gap in Instruction Alignment and Reference Consistency between the closed-source models and the other open-source models.
In particular, GPT-Image-1 achieves significantly superior performance in Instruction Alignment, especially in style and text modification tasks in the two-reference task.
Meanwhile, in Reference Consistency, Nano Banana achieves the highest scores in the single-reference task.
Nano Banana also performs on par with GPT-Image-1 in Reference Consistency across background modification, subject replacement, subject addition, and makeup tasks, indicating its strong ability to leverage reference images, particularly when the number of references is small.
On the other hand, Nano Banana performs worse than the other models in background modification tasks in Background Subject Match, Physical Realism, and Visual Quality.
This result suggests that Nano Banana struggles with background modification.

\begin{table}[t]
\centering
\caption{Comparison of Qwen-Image-Edit and Qwen-Image-Edit-2509 on single-reference task.}
\label{tab:qwen_edit}
\begin{tabular}{c c c}
\toprule
 & \textbf{Qwen-Image-Edit} & \textbf{Qwen-Image-Edit-2509} \\
\midrule
single & 6.332 & 7.499 \\
\bottomrule
\end{tabular}
\end{table}

\begin{table*}[ht]
\caption{
Detailed per-task total scores for the multi-reference tasks, evaluated by GPT.
The local, global, back, and object columns under X correspond to X–1 Objects + Local, X–1 Objects + Global, X–1 Objects + Background, and X Object, respectively.
}
\label{tab:multi_res_details_gpt}
\centering
\scalebox{0.96}{
\begin{tabular}{lcccccccccccc}
\toprule
\textbf{Model} &
\multicolumn{4}{c}{\textbf{3-references}} &
\multicolumn{4}{c}{\textbf{4-references}} &
\multicolumn{4}{c}{\textbf{5-references}} \\
\cmidrule(lr){2-5}
\cmidrule(lr){6-9}
\cmidrule(lr){10-13}
& local & global & back & object
& local & global & back & object
& local & global & back & object \\
\midrule
DreamOmni2
& 3.94 & 4.19 & 3.92 & 3.91
& 3.34 & 3.24 & 2.87 & 3.20
& 3.30 & 3.06 & 2.73 & 3.22 \\
OmniGen2
& 4.42 & 4.59 & 4.47 & 4.47
& 4.41 & 3.37 & 3.10 & 3.27
& 3.78 & 3.28 & 2.87 & 3.31 \\
%Qwen-Image-Edit
%& 2.65 & 2.65 & 2.88 & 3.56
%& 2.22 & 2.32 & 2.90 & 3.02
%& 2.10 & 2.59 & 3.06 & 2.79 \\
Qwen-Image-Edit-2509
& 4.42 & 4.65 & 4.75 & 4.77
& 3.08 & 3.35 & 2.78 & 3.54
& 1.93 & 2.40 & 1.98 & 2.10 \\
Nano Banana
& 5.44 & 5.81 & 5.10 & 5.64
& 4.57 & 5.36 & 4.27 & 5.25
& 5.12 & 5.93 & 3.96 & 4.93 \\
GPT-image-1
& 6.58 & 7.40 & 6.98 & 6.40
& 6.18 & 6.98 & 6.73 & 6.56
& 6.18 & 6.66 & 6.22 & 6.28 \\
%Nano Banana Pro
%& 6.00 & 7.27 & 5.38 & 6.38
%& 5.56 & 6.83 & 5.05 & 5.78
%& 5.96 & 6.84 & 4.40 & 5.38 \\
\midrule
\midrule
% ------------------- 下段（6〜8） -------------------
\textbf{Model} &
\multicolumn{4}{c}{\textbf{6-references}} &
\multicolumn{4}{c}{\textbf{7-references}} &
\multicolumn{4}{c}{\textbf{8-references}} \\
\cmidrule(lr){2-5}
\cmidrule(lr){6-9}
\cmidrule(lr){10-13}
& local & global & back & object
& local & global & back & object
& local & global & back & object \\
\midrule
DreamOmni2
& 3.37 & 3.03 & 2.82 & 2.96
& 3.24 & 3.02 & 2.71 & 2.86
& 3.26 & 2.79 & 2.57 & 2.82 \\
OmniGen2
& - & - & - & -
& - & - & - & -
& - & - & - & - \\
%Qwen-Image-Edit
%& 2.14 & 2.84 & 2.99 & 3.47
%& 1.68 & 2.57 & 2.68 & 2.66
%& 2.54 & 2.98 & 2.62 & 3.40 \\
Qwen-Image-Edit-2509
& 1.85 & 2.18 & 1.68 & 1.79
& 1.51 & 1.89 & 1.57 & 1.88
& 1.68 & 2.29 & 1.62 & 1.83 \\
Nano Banana
& 4.43 & 4.88 & 3.89 & 4.90
& 4.40 & 4.78 & 3.57 & 4.81
& 4.31 & 4.82 & 3.21 & 4.45 \\
GPT-image-1
& 6.16 & 6.08 & 5.68 & 5.90
& 5.38 & 6.09 & 5.59 & 5.94
& 4.96 & 5.82 & 5.03 & 5.15 \\
%Nano Banana Pro
%& 5.45 & 6.36 & 4.90 & 5.22
%& 5.53 & 6.79 & 4.44 & 5.34
%& 5.21 & 6.54 & 4.40 & 4.99 \\
\bottomrule
\end{tabular}
}
\end{table*}

\begin{table*}[ht]
\caption{
Detailed per-task total scores for the multi-reference tasks, evaluated by Gemini.
The local, global, back, and object columns under X correspond to X-1 Objects + Local, X–1 Objects + Global, X–1 Objects +  Background, and X Object, respectively.
}
\label{tab:multi_res_detail_gemini}
\centering
\scalebox{0.96}{
\begin{tabular}{lcccccccccccc}
\toprule
\textbf{Model} &
\multicolumn{4}{c}{\textbf{3-references}} &
\multicolumn{4}{c}{\textbf{4-references}} &
\multicolumn{4}{c}{\textbf{5-references}} \\
\cmidrule(lr){2-5}
\cmidrule(lr){6-9}
\cmidrule(lr){10-13}
& local & global & back & object
& local & global & back & object
& local & global & back & object \\
\midrule
DreamOmni2
& 3.00 & 2.49 & 2.45 & 2.74
& 2.84 & 2.73 & 2.38 & 2.44
& 2.63 & 2.50 & 2.36 & 2.66 \\
OmniGen2
& 3.03 & 2.89 & 3.00 & 3.25
& 3.10 & 3.06 & 2.27 & 2.73
& 2.84 & 3.02 & 2.41 & 2.50 \\
%Qwen-Image-Edit
%& 1.98 & 1.63 & 2.04 & 3.15
%& 1.73 & 1.82 & 2.05 & 2.86
%& 1.79 & 1.62 & 2.20 & 2.56 \\
Qwen-Image-Edit-2509
& 3.27 & 3.16 & 3.29 & 3.97
& 2.29 & 3.00 & 2.15 & 2.79
& 1.14 & 1.69 & 1.21 & 1.20 \\
Nano Banana
& 4.34 & 4.69 & 4.61 & 4.89
& 3.42 & 4.34 & 3.79 & 4.14
& 4.25 & 4.50 & 3.16 & 3.85 \\
GPT-image-1
& 5.48 & 6.28 & 5.37 & 5.16
& 4.83 & 5.26 & 4.75 & 5.04
& 4.89 & 5.20 & 4.25 & 4.10 \\
%Nano Banana Pro
%& 5.00 & 6.37 & 4.09 & 5.30
%& 4.19 & 5.18 & 3.38 & 4.18
%& 4.44 & 4.63 & 2.72 & 3.50 \\
\midrule
\midrule
% ------------------- 下段（6〜8） -------------------
\textbf{Model} &
\multicolumn{4}{c}{\textbf{6-references}} &
\multicolumn{4}{c}{\textbf{7-references}} &
\multicolumn{4}{c}{\textbf{8-references}} \\
\cmidrule(lr){2-5}
\cmidrule(lr){6-9}
\cmidrule(lr){10-13}
& local & global & back & object
& local & global & back & object
& local & global & back & object \\
\midrule
DreamOmni2
& 2.47 & 2.54 & 2.14 & 2.36
& 2.65 & 2.50 & 2.15 & 2.30
& 2.42 & 2.33 & 2.04 & 2.19 \\
OmniGen2
& - & - & - & -
& - & - & - & -
& - & - & - & - \\
%Qwen-Image-Edit
%& 1.57 & 1.88 & 2.05 & 3.23
%& 1.30 & 2.09 & 2.05 & 2.69
%& 1.90 & 2.09 & 2.05 & 3.20 \\
Qwen-Image-Edit-2509
& 1.00 & 1.26 & 1.00 & 1.03
& 1.01 & 1.06 & 1.33 & 1.17
& 1.18 & 1.31 & 1.01 & 1.00 \\
Nano Banana
& 3.25 & 3.91 & 2.55 & 3.79
& 3.05 & 3.73 & 2.26 & 3.52
& 2.85 & 3.64 & 2.53 & 3.25 \\
GPT-image-1
& 3.66 & 4.66 & 3.46 & 4.02
& 4.18 & 4.58 & 3.24 & 3.52
& 3.28 & 4.07 & 2.91 & 2.96 \\
%Nano Banana Pro
%& 3.57 & 4.74 & 3.11 & 3.80
%& 3.79 & 5.21 & 2.91 & 3.67
%& 3.21 & 4.61 & 2.73 & 3.57 \\
\bottomrule
\end{tabular}
}
\end{table*}

\begin{table*}[ht]
\caption{
Detailed per-task total scores for each multi-reference task, evaluated by the average of GPT and Gemini.
The local, global, back, and object columns under X correspond to X–1 Objects + Local, X–1 Objects + Global, X–1 Objects + Background, and X Object, respectively.
}
\label{tab:multi_res_detail_all}
\centering
\scalebox{0.96}{
\begin{tabular}{lcccccccccccc}
\toprule
\textbf{Model} &
\multicolumn{4}{c}{\textbf{3-references}} &
\multicolumn{4}{c}{\textbf{4-references}} &
\multicolumn{4}{c}{\textbf{5-references}} \\
\cmidrule(lr){2-5}
\cmidrule(lr){6-9}
\cmidrule(lr){10-13}
& local & global & back & object
& local & global & back & object
& local & global & back & object \\
\midrule
DreamOmni2
& 3.47 & 3.34 & 3.18 & 3.33
& 3.09 & 2.98 & 2.62 & 2.82
& 2.97 & 2.78 & 2.54 & 2.94 \\
OmniGen2
& 3.73 & 3.74 & 3.74 & 3.86
& 3.76 & 3.22 & 2.69 & 3.00
& 3.31 & 3.15 & 2.64 & 2.91 \\
%Qwen-Image-Edit
%& 2.31 & 2.14 & 2.46 & 3.35
%& 1.97 & 2.07 & 2.47 & 2.94
%& 1.95 & 2.11 & 2.63 & 2.68 \\
Qwen-Image-Edit-2509
& 3.85 & 3.90 & 4.02 & 4.37
& 2.69 & 3.17 & 2.47 & 3.17
& 1.54 & 2.04 & 1.60 & 1.65 \\
Nano Banana
& 4.89 & 5.25 & 4.85 & 5.27
& 4.00 & 4.85 & 4.03 & 4.70
& 4.69 & 5.22 & 3.56 & 4.39 \\
GPT-image-1
& 6.03 & 6.84 & 6.18 & 5.78
& 5.50 & 6.12 & 5.74 & 5.80
& 5.54 & 5.93 & 5.24 & 5.19 \\
%Nano Banana Pro
%& 5.50 & 6.82 & 4.74 & 5.84
%& 4.87 & 6.00 & 4.22 & 4.98
%& 5.20 & 5.74 & 3.56 & 4.44 \\
\midrule
\midrule
% ------------------- 下段（6〜8） -------------------
\textbf{Model} &
\multicolumn{4}{c}{\textbf{6-references}} &
\multicolumn{4}{c}{\textbf{7-references}} &
\multicolumn{4}{c}{\textbf{8-references}} \\
\cmidrule(lr){2-5}
\cmidrule(lr){6-9}
\cmidrule(lr){10-13}
& local & global & back & object
& local & global & back & object
& local & global & back & object \\
\midrule
DreamOmni2
& 2.92 & 2.78 & 2.48 & 2.66
& 2.94 & 2.76 & 2.43 & 2.58
& 2.84 & 2.56 & 2.31 & 2.50 \\
OmniGen2
& - & - & - & -
& - & - & - & -
& - & - & - & - \\
%Qwen-Image-Edit
%& 1.85 & 2.36 & 2.52 & 3.35
%& 1.49 & 2.33 & 2.37 & 2.67
%& 2.22 & 2.53 & 2.33 & 3.30 \\
Qwen-Image-Edit-2509
& 1.42 & 1.72 & 1.34 & 1.41
& 1.26 & 1.47 & 1.45 & 1.52
& 1.43 & 1.80 & 1.32 & 1.42 \\
Nano Banana
& 3.84 & 4.39 & 3.22 & 4.35
& 3.72 & 4.25 & 2.92 & 4.16
& 3.58 & 4.23 & 2.87 & 3.85 \\
GPT-image-1
& 4.91 & 5.37 & 4.57 & 4.96
& 4.78 & 5.33 & 4.42 & 4.73
& 4.12 & 4.94 & 3.97 & 4.05 \\
%Nano Banana Pro
%& 4.51 & 5.55 & 4.01 & 4.51
%& 4.66 & 6.00 & 3.68 & 4.50
%& 4.21 & 5.57 & 3.57 & 4.28 \\
\bottomrule
\end{tabular}
}
\end{table*}

\subsection{The Effect of the Number of References}
\begin{figure*}[ht]
    \centering
    \includegraphics[width=0.98\textwidth]{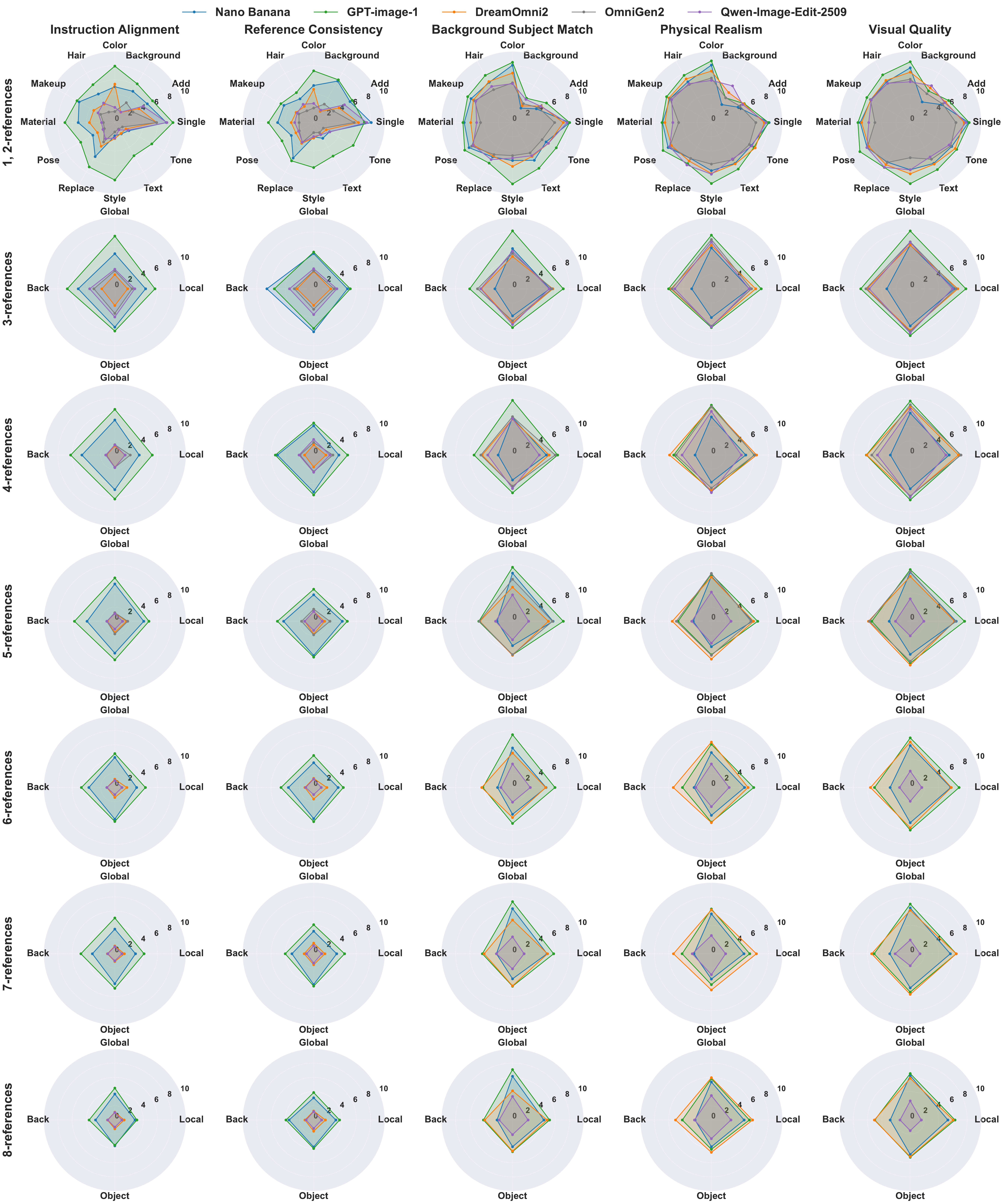}
    \caption{Average scores of each model across evaluation metrics for each task by the average of GPT and Gemini. The horizontal axis denotes the scores for the five evaluation criteria, and the vertical axis denotes the number of reference images.}
    \label{fig:scores_average_pertask}
\end{figure*}

According to the per-metric scores in \autoref{fig:scores_average_pertask} and the total scores in \autoref{tab:multi_res_detail_all}, all models exhibit decreasing performance as the number of reference images increases.
For Instruction Alignment and Reference Consistency, GPT-Image-1 and Nano Banana achieve relatively high scores, whereas the open-source model, DreamOmni2, tends to approach the minimum score of 1 as the number of references increases.
In contrast, for quality-related metrics, Background Subject Match, Physical Realism, and Visual Quality, the DreamOmni2 maintains high scores even when more reference images are provided.
As discussed in Section~\ref{sec:experiments}, this is because closed-source models prioritize adherence to the references and editing instructions over visual quality, while open-source models show a tendency to preserve visual quality but often ignore the references.

In the finer-grained task-level evaluation, we found that, even as the number of reference images increased, the X–1 Object + Global task did not show severe decreases in Background Subject Match, Physical Realism, or Visual Quality scores.
The task requires modifying the image to a specified style; therefore, even if the reference images come from different domains, which may degrade the visual quality, the final output often converges to a unified style.

\begin{figure*}[ht]
    \centering
    \includegraphics[width=0.75\linewidth]{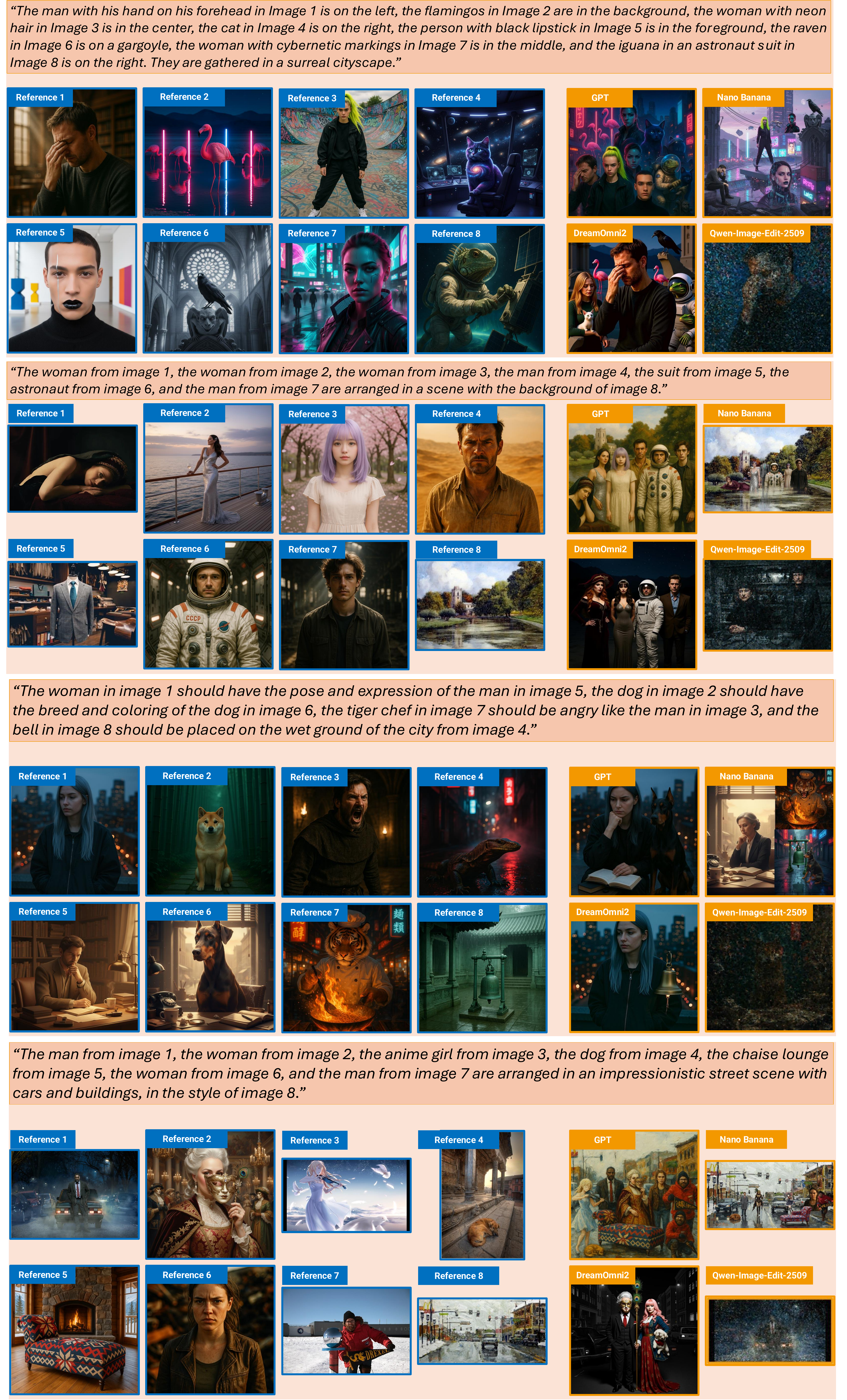}
    \caption{Qualitative example for Tasks with 8 references.}
    \label{fig:qualitative_8}
\end{figure*}

In the results of Qwen-Image-Edit-2509, the model may be trained to handle up to three reference images; therefore, when provided with four or more references, the input becomes out-of-distribution. 
This causes the model to produce perceptually invalid outputs, and then the scores approach the minimum value of 1 (see \autoref{fig:qualitative_8}).

\subsection{Difficult Reference Combination}
\begin{figure*}[ht]
    \centering
    \includegraphics[width=1.0\linewidth]{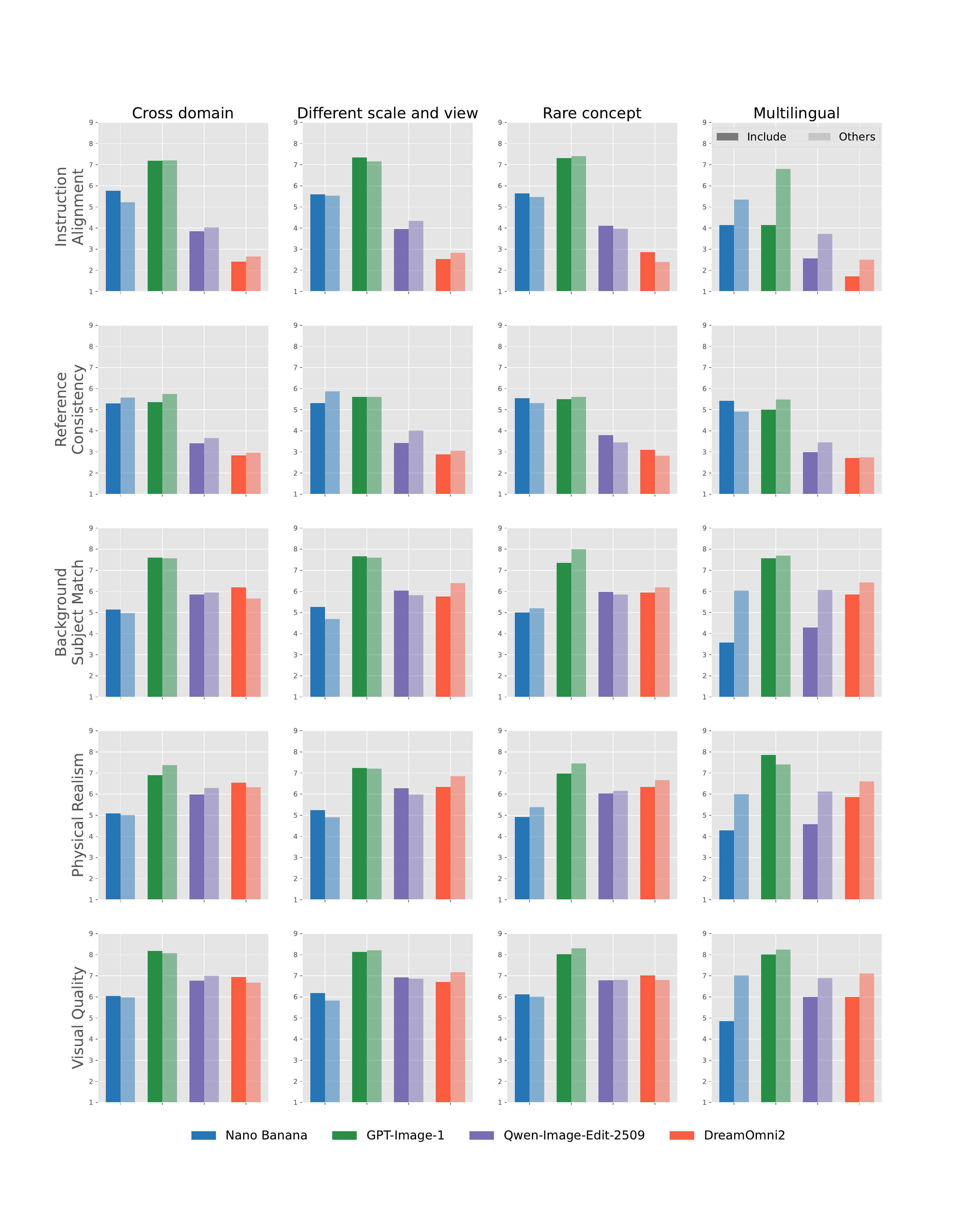}
    \vspace{-10mm}
    \caption{Total scores for each difficult reference combination across models. Darker colors represent the average score for tasks that include the corresponding combination, whereas lighter colors indicate the average score for tasks that do not include it.}
    \label{fig:allscore_difficult}
\end{figure*}

\autoref{fig:allscore_difficult} shows the scores of each model in difficult reference combination tasks across evaluation metrics.

\noindent\textbf{Cross-domain diversity}~~
In cross-domain cases, samples with reference images from different domains (darker color) received lower Reference Consistency scores because all models tend to distort individual reference attributes to achieve a consistent style.

\noindent\textbf{Scale and viewpoint differences}~~
In the different scale and viewpoint cases, the Reference Consistency scores declined consistently across all models. 
This suggests that when models attempt to reproduce objects at different scales or from different viewpoints, they may fail to preserve fine details or adjust the pose to achieve a more natural appearance.

\noindent\textbf{Rare concept references}~~
In the rare-concept case, the model may struggle to handle uncommon subjects relative to more familiar ones. 
Because the reference images often depict these subjects at a large scale, the model tends to reproduce this scale without appropriate adjustment. 
This suggests that, unlike common concepts, the model may find it more difficult to flexibly control attributes such as size when dealing with rare concepts. 
As a result, rare concept samples tend to receive lower Physical Realism scores.

\noindent\textbf{Multilingual references}~~
In the multilingual text rendering case, we observed a consistent decline across all metrics except Reference Consistency.
This may occur because models have limited text-rendering capability in languages other than English, leading to failures in cross-language conversion.
As a result, instruction alignment and visual quality decrease.
However, GPT-Image-1 exhibits a smaller decline than other models, owing to its comparatively stronger text-rendering ability in non-English languages.

\subsection{Risks of Model Bias in Synthetic References}

\begin{figure}[t]
  \centering
  \includegraphics[width=1.0\linewidth]{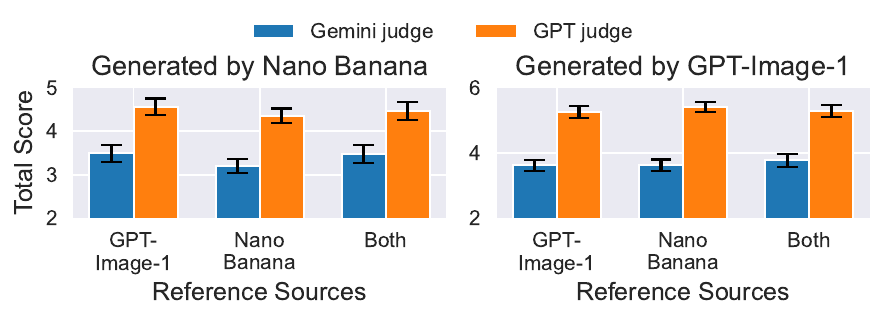}
  % \vskip -0.1in
  \caption{
  Analysis of potential in-distribution bias. % by reference source. 
  %The benchmark is partitioned into three subsets based on the provenance of the reference images: sets containing images generated by GPT-Image-1, by Nano Banana, and both. 
  %Mean scores across these subsets remain within their respective 99\% confidence intervals, indicating no statistically significant bias from specific reference models.
  % Error bars: the 99\% confidence intervals.
  Error bars indicate the 99\% confidence intervals.
  For image generation models, we use the versions available as of January 2026.
  }
  \label{fig:vlm_bias}
\end{figure}

Synthetic references generated by closed models (i.e., Nano Banana, GPT-Image-1) may introduce in-distribution advantages when those same models are also evaluated on the benchmark.
To verify this, we divided the benchmark into three subsets based on the source of the reference images: those from GPT-Image-1, those from Nano Banana, and those from both.
The mean scores across all subsets remain within their respective 99\% confidence intervals, indicating no statistically significant bias (\autoref{fig:vlm_bias}).

\subsection{Potential Conflict in Cross-Domain Task}

\begin{figure*}[t]
  \centering
  \includegraphics[width=1.0\linewidth]{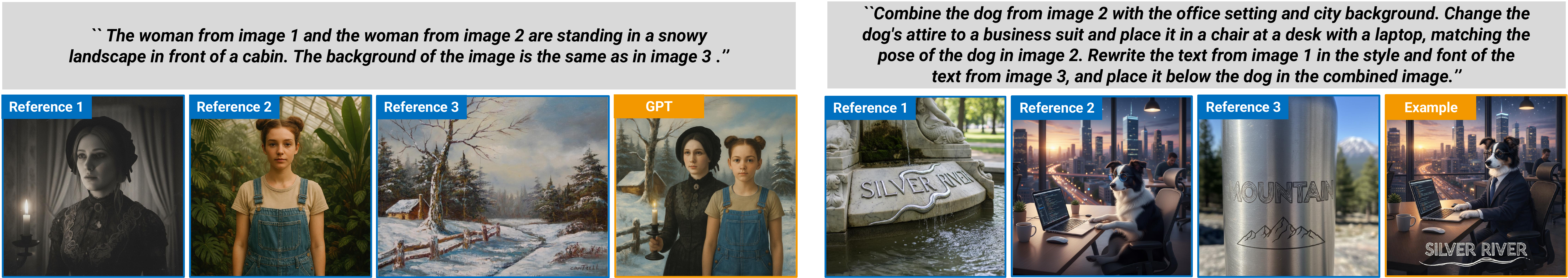}
   \caption{
   \textbf{Left}: A sample generated by GPT-Image-1 on a cross-domain
   % (black-and-white photo, color photo, and painting) 
   task. 
   % GPT-5 judged the Reference Consistency score of 7 and the Background–Subject Match score of 9.
   \textbf{Right}: Task example of our benchmark directly relevant to real-world use cases, such as advertising.
   % content production, advertising, and fashion design.
   }
   \label{fig:qualitative_application}
\end{figure*}

One might expect that preserving subject details (reference consistency) and maintaining a coherent scene (background-subject match) are inherently at odds in cross-domain tasks—e.g., placing a photorealistic person into an anime background. 
However, our evaluation shows this is not always the case. 
As shown in \autoref{fig:qualitative_application} (left), a generated image can satisfy both criteria, achieving a GPT-5 score of 7 for reference consistency and 9 for background-subject match. 
To further verify this, we created cross-domain editing tasks with explicit instructions that prioritize one criterion over the other for 20 randomly selected prompts. 
\autoref{tab:consist_or_match} shows that Nano Banana can improve the prioritized criterion while maintaining the other criterion. 

\begin{table}[t]
\centering
\caption{
Comparison of the mean scores when prioritizing either reference consistency or background–subject match for the subset generated by Nano Banana, latest version as of January 2026.
}
\label{tab:consist_or_match}
\small
\scalebox{1.0}{
\begin{tabular}{lcc}
\toprule
\textbf{Prompt} 
% & \textbf{Reference Consistency}
% & \textbf{Background-Subject Match} \\
& \begin{tabular}{c}\textbf{Reference}\\\textbf{Consistency}\end{tabular}
& \begin{tabular}{c}\textbf{Background--}\\\textbf{Subject Match}\end{tabular} \\
\midrule
Original & 3.58 & 3.67 \\
Consistency priority & 4.17 & 3.25 \\
Matching priority & 3.40 & 4.03 \\
\bottomrule
\end{tabular}
}
\end{table}

\subsection{Alignment with Real-World Use Cases}
As shown in \autoref{fig:qualitative_application} (right), MultiBanana is directly relevant to real-world use cases, such as advertising. 
Our benchmark development using synthetic data also aligns with the community standard. For example, ImgEdit~\citep{ye2025imgedit} similarly employs GPT-4o for instruction generation, which also aims to develop ``practical and powerful tools for real-world applications.''

\subsection{Agentic Inference}
\label{sec:agent}

\begin{figure*}[t]
    \centering
    \includegraphics[width=\textwidth]{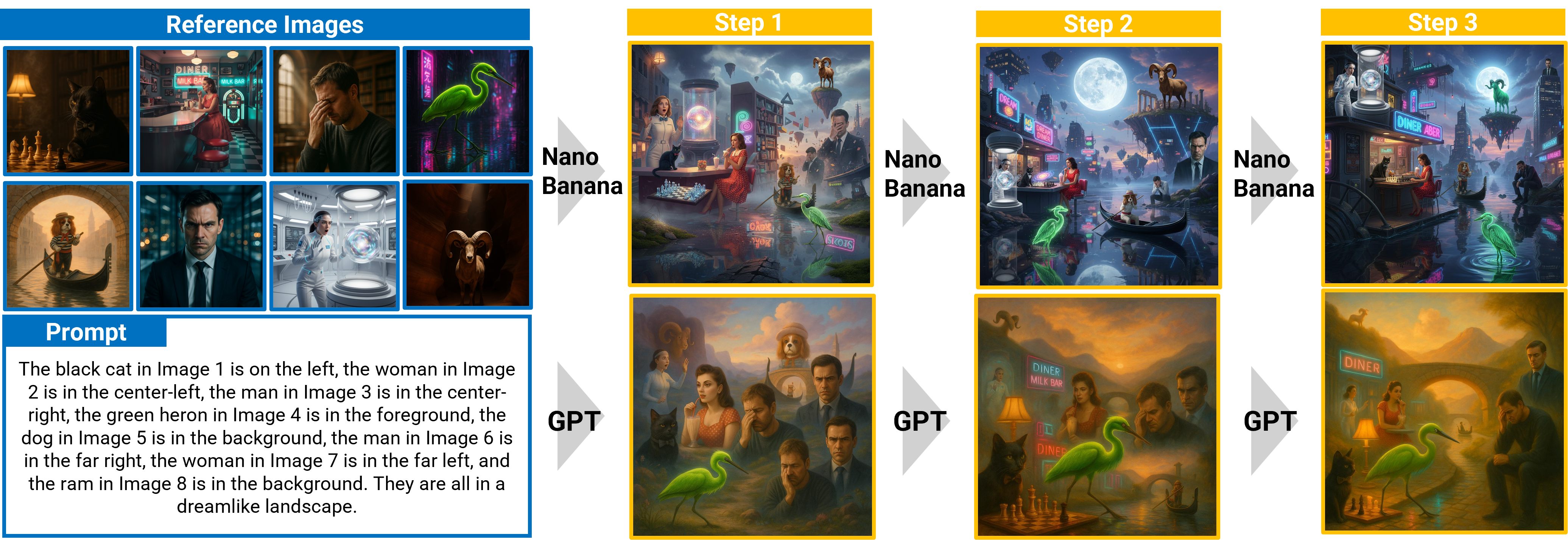}
    \caption{Example of changes in the Iterative Prompt Refinement (IPR) framework. Comparison and evaluation results between Nano Banana and GPT on the task of generating images from 8 object references according to the instruction prompt.}
    \label{fig:example_agent}
\end{figure*}
We introduced three agentic frameworks: Iterative Prompt Refinement (IPR), Context-Aware Feedback Generation (CAFG), and Selective Reference Adaptation (SRA).
Here, we refer to $\mathrm{Gen}$ as the \textit{Generator}, which produces an image, and $\mathrm{Plan}$ as the \textit{Planner}, which updates the instruction prompt (and selects reference images) for the next step.
Let $P^t$ denote the instruction prompt at step $t$, $\{R_i\}_{i \in [I]}$ the reference images, and $G^t$ the generated image with $G^0 = \varnothing$.
We conducted experiments with both GPT and Gemini. For GPT, we use GPT-Image-1 as the \textit{Generator} and GPT-5 as the \textit{Planner}. For Gemini, we use Nano Banana as the \textit{Generator} and Gemini 2.5 Flash as the \textit{Planner}.

\subsubsection{Iterative Prompt Refinement (IPR)}
The IPR framework is formulated as follows. The \textit{Planner} refines the prompt based on the generation result from the previous step.
\begin{align*}
    G^{t+1} = \mathrm{Gen}(P^t, \{R_i\}_{i \in [I]}, \varnothing),\\
    P^{t+1} = \mathrm{Plan}(P^t, \{R_i\}_{i \in [I]}, G^{t+1}).
\end{align*}
\subsubsection{Context-Aware Feedback Generation (CAFG)}
The CAFG framework is formulated as follows. The \textit{Generator} generates the image based on the generation result (context) from the previous step.
\begin{align*}
    G^{t+1} = \mathrm{Gen}(P^t, \{R_i\}_{i \in [I]}, \boldsymbol{G^{t}}),\\
    P^{t+1} = \mathrm{Plan}(P^t, \{R_i\}_{i \in [I]}, G^{t+1}).
\end{align*}
\subsubsection{Selective Reference Adaptation (SRA)}
The SRA framework is formulated as follows. The \textit{Planner} selects only the reference images that should be improved based on the generation result from the previous step, and the \textit{Generator} receives these as context. SRA is expected to reduce agents' task complexity by adaptively decreasing the number of references.
\begin{align*}
    G^{t+1} = \mathrm{Gen}(P^t, \{R_i\}_{\boldsymbol{i \in U^t}}, G^{t}),\\
    P^{t+1}, \boldsymbol{U^{t+1}} = \mathrm{Plan}(P^t, \{R_i\}_{i \in [I]}, G^{t+1}).
\end{align*}
where $U^t$ is the index set of reference images that are insufficiently reflected in the generated image $G^t$.

\subsubsection{Experiments}
\begin{figure*}[t]
  \centering
  {\bfseries\normalsize IPR Framework with Gemini}\par\vspace{0.8em}
  \includegraphics[width=\textwidth,page=1]{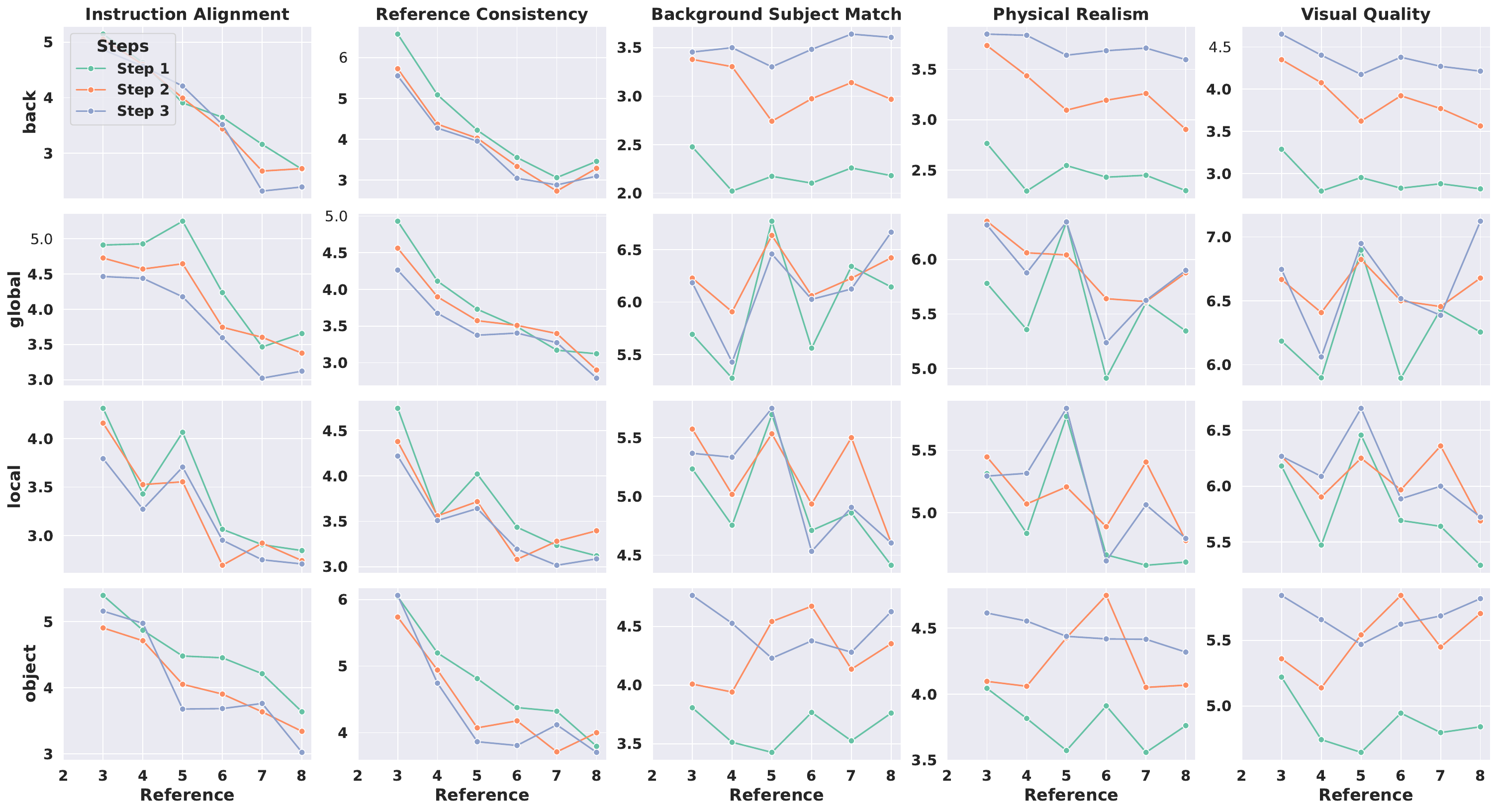}
  \caption{Detailed results of multi-reference image generation using the IPR framework with Gemini.}
  \label{fig:agentic_gemini_ipr}
\end{figure*}

\begin{figure*}[t]
  \centering
  {\bfseries\normalsize CAFG Framework with Gemini}\par\vspace{0.8em}
  \includegraphics[width=\textwidth,page=1]{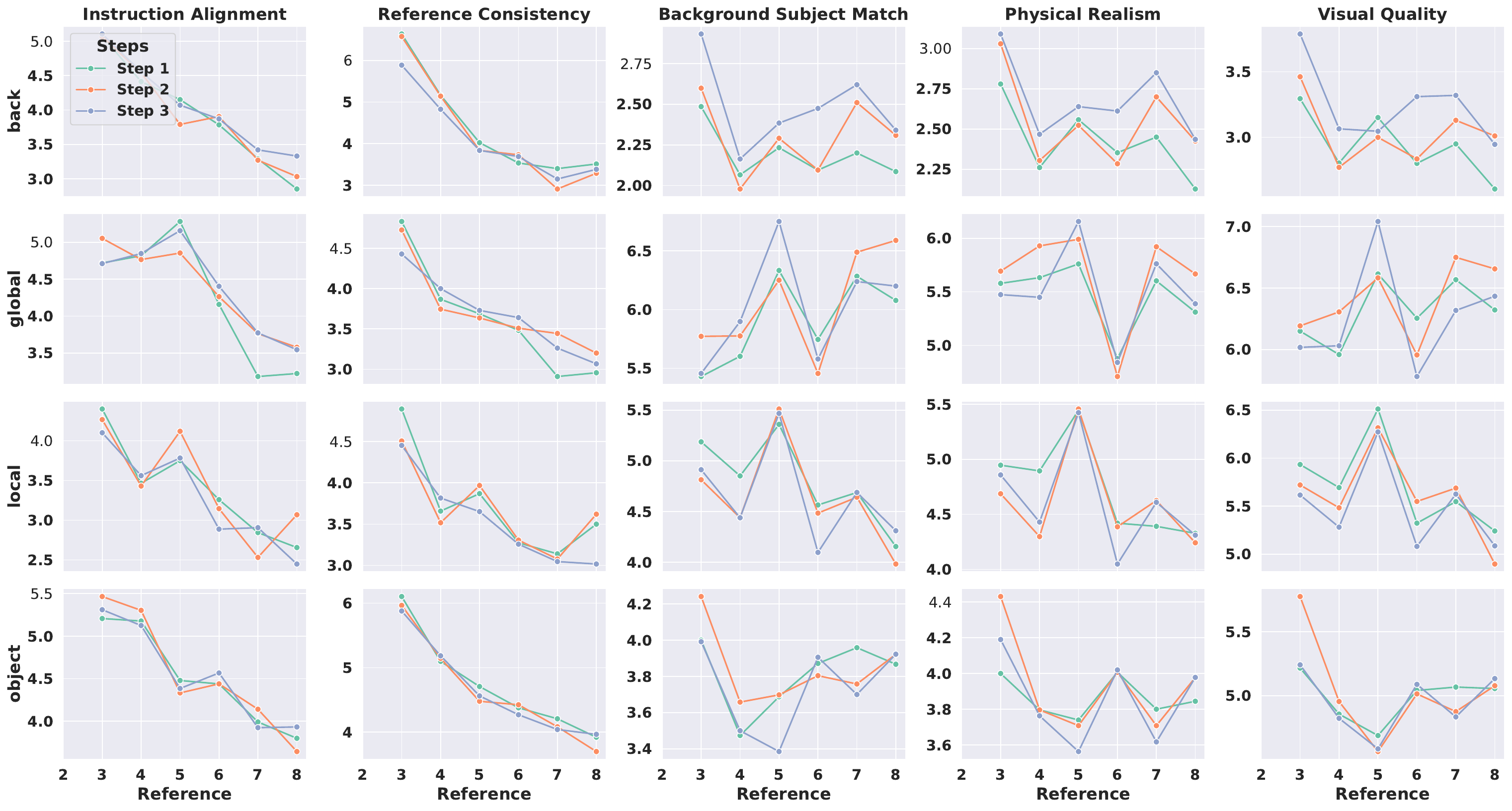}
  \caption{Detailed results of multi-reference image generation using the CAFG framework with Gemini.}
  \label{fig:agentic_gemini_cafg}
\end{figure*}

\begin{figure*}[t]
  \centering
  {\bfseries\normalsize SRA Framework with Gemini}\par\vspace{0.8em}
  \includegraphics[width=\textwidth,page=1]{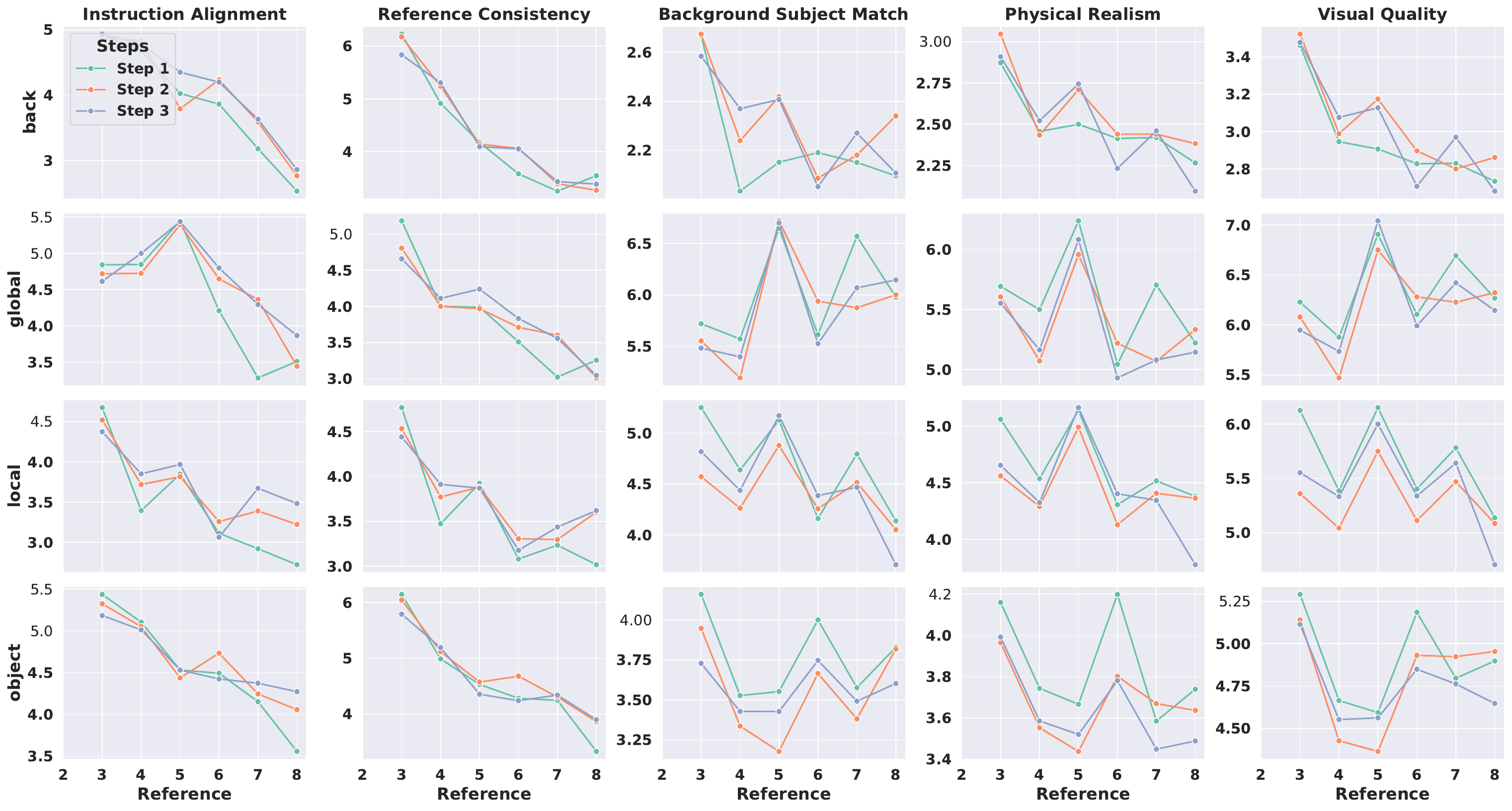}
  \caption{Detailed results of multi-reference image generation using the SRA framework with Gemini.}
  \label{fig:agentic_gemini_sra}
\end{figure*}

\begin{figure*}[t]
  \centering
  {\bfseries\normalsize IPR Framework with GPT}\par\vspace{0.8em}
  \includegraphics[width=\textwidth,page=1]{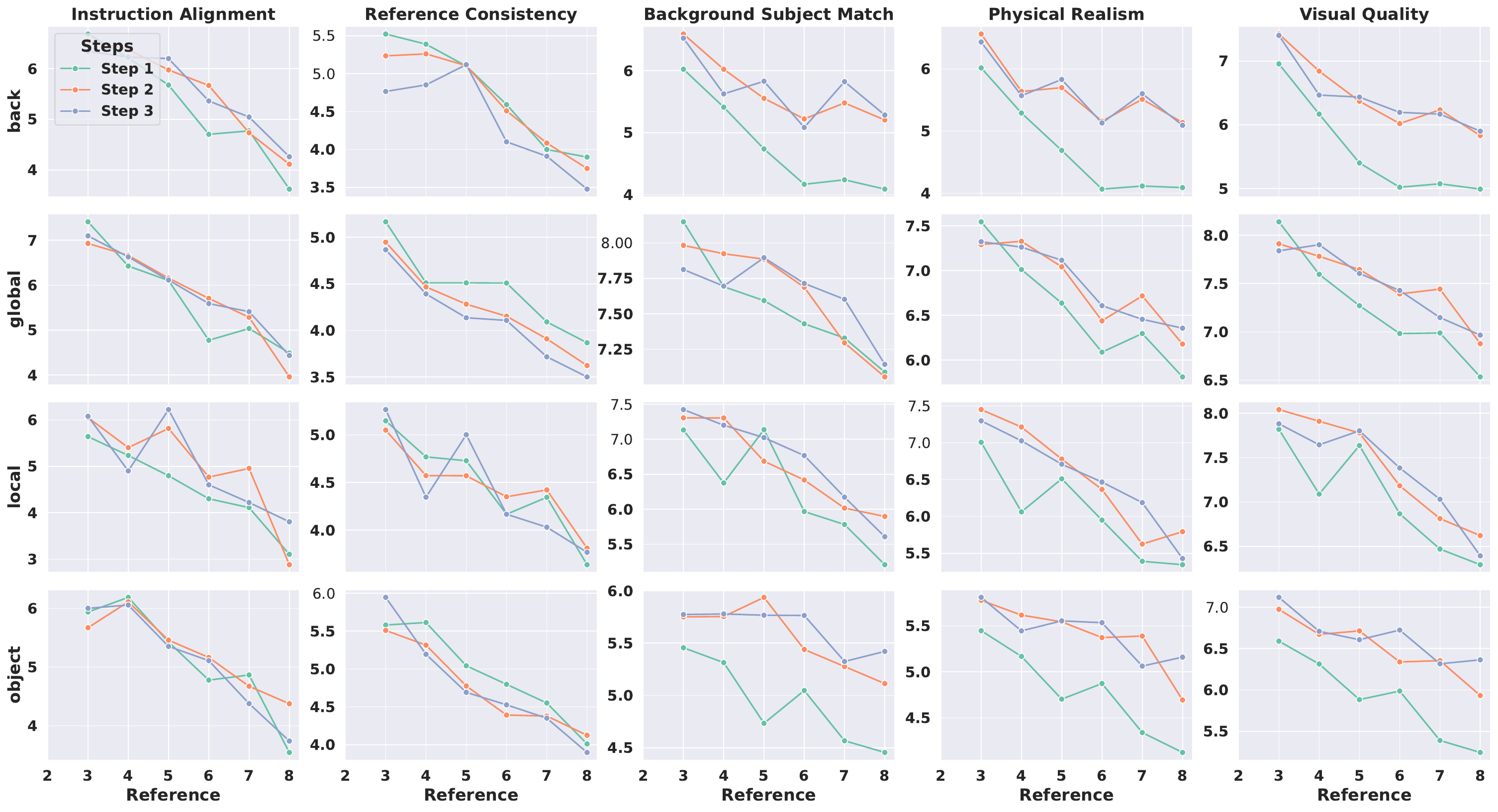}
  \caption{Detailed results of multi-reference image generation using the IPR framework with GPT.}
  \label{fig:agentic_gpt_ipr}
\end{figure*}

\begin{figure*}[t]
  \centering
  \includegraphics[width=\textwidth,page=1]{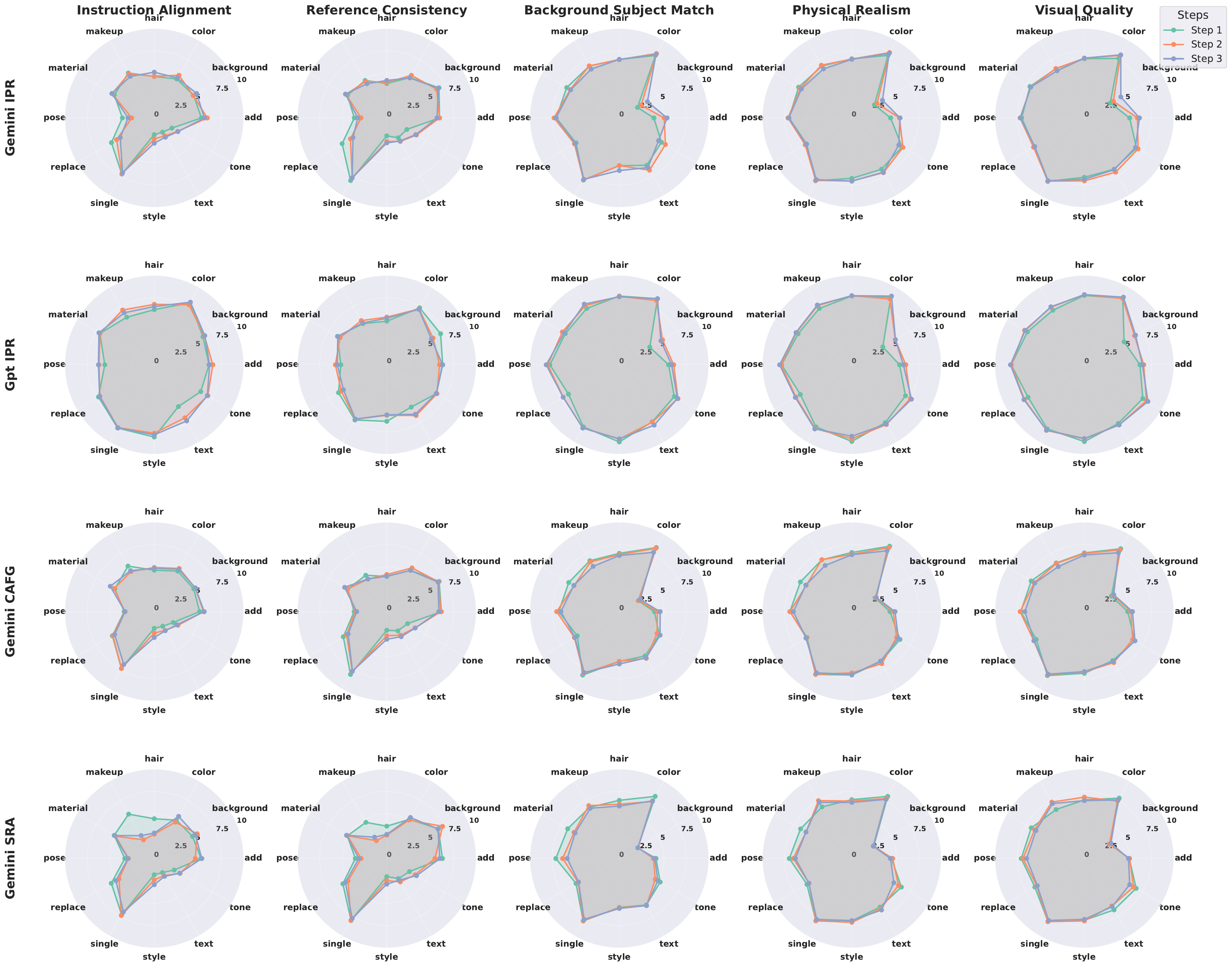}
  \caption{Detailed results of single and two-reference image generation using the agentic frameworks.}
  \label{fig:agentic_radar_chart}
\end{figure*}

We set the maximum number of steps $t$ to $3$ and evaluate three agentic frameworks—IPR, CAFG, and SRA—using Gemini, and the IPR framework using GPT. \autoref{fig:example_agent} illustrates an example of step-wise improvement in the IPR framework. \autoref{fig:agentic_gemini_ipr}, \autoref{fig:agentic_gemini_cafg}, \autoref{fig:agentic_gemini_sra}, and \autoref{fig:agentic_gpt_ipr} present the evaluation results of multi-reference generations at each step, averaged across judgments from Gemini 2.5 Flash and GPT-5, broken down by category. \autoref{fig:agentic_radar_chart} presents the results for single and two-reference generations. Nano Banana (Gemini) shows modest improvements in Physical Realism and Visual Quality across refinement steps, while Instruction Alignment and Reference Consistency either remain unchanged or deteriorate. In contrast, GPT demonstrates consistent improvements across all categories. This suggests that Gemini's planner progressively loses information from the original prompt as refinement steps proceed.

\section{Implementations}

\noindent\textbf{Code}:~\url{https://github.com/matsuolab/multibanana}.

\noindent\textbf{Dataset}:~\url{https://huggingface.co/datasets/kohsei/MultiBanana-Benchmark}.

\noindent\textbf{API endpoints}:~For VLM evaluation, we used stable API versions: \texttt{gpt-5-2025-08-07} and \texttt{gemini-2.5-flash}.
For image generation models, we use the latest available API versions as of November 2025, unless otherwise noted.

% \clearpage
\section{Extended Related Works}
\subsection{Controllable Text-to-Image Generation}
Stable Diffusion~\citep{rombach2022ldm, podell2024sdxl, esser2024sd3}, FLUX~\citep{flux2024, labs2025flux1kontext}, DALL-E~\citep{ramesh2022dalle}, and Imagen~\citep{saharia2022imagen, imagenteamgoogle2024imagen3}, have demonstrated strong text-to-image generation capabilities, establishing a scalable foundation for the task.
To enhance controllability, models such as ControlNet~\citep{zhang2023controlnet} and T2I-Adapter~\citep{mou2023t2i-adapter} introduced external conditioning modules, enabling image-conditioned generation.
% Recent large pretrained diffusion models with textual conditioning, such as Stable Diffusion~\citep{rombach2022ldm, podell2024sdxl, esser2024sd3} and FLUX~\citep{flux2024, labs2025flux1kontext}, enable the synthesis of novel scene images while preserving the identity depicted in the reference images.
% Building on these developments, approaches based on fine-tuning (e.g., DreamBooth~\citep{ruiz2022dreambooth}) and adapter-based methods (e.g., IP-Adapter~\citep{ye2023ip-adapter}) have been actively explored to address this challenge~\citep{jiang2025infiniteyou,gal2023an,brooks2023instructpix2pix,mokady2023null,tan2025ominicontrol,NEURIPS2023_602e1a5d}.
Additionally, training-free approaches (e.g., pix2pix-zero~\citep{parmar2023zero} and prompt-to-prompt~\citep{hertz2023prompttoprompt}) have also been widely proposed~\citep{Cao_2023_ICCV,Huberman-Spiegelglas_2024_CVPR,Wallace_2023_CVPR,Tumanyan_2023_CVPR,kulikov2025flowedit,miyake2025negative-prompt-inversion}.
These advancements collectively demonstrate that diffusion models are becoming increasingly flexible and adaptable across diverse conditional generation settings.

% In recent years, research on unified image generation has gained momentum, aiming to handle multiple image generation tasks within a single model.
% For example, ChatGPT-Image-1~\citep{openai2025gpt4oimage} and Nano Banana~\citep{google2025nanobanana} can jointly process text and image inputs, enabling integrated image generation and editing within a unified framework.
% Similarly, open-source models such as Qwen-Image~\citep{wu2025qwenimagetechnicalreport} and FLUX.1-Kontext-dev~\citep{labs2025flux1kontext}, built on robust multimodal large language model backbones, demonstrate high-quality, flexible image generation and editing.
% Beyond these, numerous open-source unified generative models continue to emerge~\citep{xiao2024omnigen, xia2025dreamomni, wu2025omnigen2, xia2025dreamomni2}.

% With the advent of multimodal foundation models capable of jointly handling text and images~\citep{google2025nanobanana, openai2025gpt4oimage}, a new task has recently become feasible: multi-reference-based image generation.
% In this task, the model receives multiple reference images along with textual instructions and generates a new scene while preserving the identity of the reference images.
% Despite its novelty and potential, a reliable benchmark for evaluating this innovative task has not yet been established, leaving it as an essential open challenge for future research.

\subsection{Benchmarks for Reference-Based Generation}

Reference-based image generation is closely related to industrial applications such as content production~\citep{ruiz2022dreambooth, ye2023ip-adapter, wang2024ms, zhou2024migc, yan2025colorizeddiffusion, kodaira2025streamdiffusion}, advertising~\citep{inoue2023layout, horita2024retrievalaugmented, morita2025tkg}, and fashion design~\citep{zhu2023tryondiffusion, choi2024improving, fang2024vivid, fang2024vivid, chong2024catvton, xu2024ootdiffusion, kim2024stableviton}.
Among reference-driven image generation tasks, the most widely recognized one is image editing.
% Image editing is the process of generating an edited image from a single reference image and a textual instruction describing how the edit should be performed.
% Representative benchmarks such as MagicBrush~\citep{zhang2023magicbrush} and EMU-Edit~\citep{sheynin2024emuedit} have extended task-specific evaluations but still rely heavily on similarity-based metrics.
% SmartEdit~\citep{huang2024smartedit} supports editing of complex scenes, but does not sufficiently cover more general settings.
% Similarly, I2E-Bench~\citep{ma2024i2ebench} employs GPT-4o to provide human-aligned evaluations across a variety of editing tasks, yet it uses different metrics for each task and therefore fails to capture the shared characteristics of editing as a whole.
% More recently, ImgEdit~\citep{ye2025imgedit} has enabled the evaluation of multi-turn editing tasks, in which image generation and editing are performed interactively.
RealEdit~\citep{sushko2025realedit} focuses on single-image editing and provides an empirical analysis of real-world image-editing use cases.
SpotEdit~\citep{ghazanfari2025spotedit} proposes a benchmark for visually guided image editing.
These works focus on benchmarking single-reference image edition/generation, while we focus on multi-reference image generation.
% On the other hand, one of the earliest benchmarks for reference-based image generation beyond editing is the DreamBooth~\citep{ruiz2022dreambooth} dataset.
% However, such benchmarks are still limited to generation tasks conditioned on a single reference image and thus cannot handle more complex or compositional settings.
In recent years, increasing attention has been directed toward multi-reference image generation, in which multiple reference images are jointly used to generate a single image.
% Benchmarks such as OmniContext~\citep{wu2025omnigen2} and DreamOmni2~\citep{xia2025dreamomni2} have been proposed to evaluate this task.
% Nevertheless, these benchmarks encompass only a small number of images and task types, thereby limiting the scope of the evaluation.
% Moreover, although designed for multi-reference generation, they do not evaluate qualitative differences or relationships among reference images, and their setups remain similar to those of conventional single-image editing benchmarks.
MultiRef-Bench~\citep{chen2025multiref} studies controllable image generation with multiple visual references, but our benchmark outperforms it by providing 3,769 samples (compared to 1,990), supporting up to 8 references (compared to 6), and using raw images (compared to bounding boxes or masks), making it more practical.

\subsection{Instruction-Following in LLMs}
The task of generating an image that faithfully reflects multiple references is analogous to instruction following in LLMs~\citep{zhou2023ifeval, zhou2023leasttomost, liu2024lost_middle, dussolle2025mifeval, wang2025lost_distance}, in which models must simultaneously satisfy multiple constraints. 
In the text domain, recent studies have revealed that LLMs struggle to follow all given instructions as the number of constraints grows, often ignoring or inadequately addressing a subset of them~\citep{jiang-etal-2024-followbench, he2024multiif, harada2025multi-instructions, laban2026llms}.
Multi-reference image generation poses a parallel challenge: as the number of reference images increases, models must reconcile potentially conflicting visual cues while remaining faithful to every reference.

\section{Limitations and Discussions}
Our benchmark focuses on static image generation and does not address video generation or editing~\citep{ho2022videodiffusionmodels, gdm2024veo, openai2024sora}, where temporal consistency across frames introduces additional challenges for reference-based conditioning~\citep{huang2025videomage, chen2025videoalchemist, abdal2025dynamic}.
% In addition, although we include multilingual text rendering as an evaluation axis, the language coverage is limited to English, Chinese, and Japanese.
% Expanding to a broader set of languages and scripts would further test the generalizability of multi-reference generation models.
Moreover, our prompts describe object placement in natural language (e.g., "on the left," "in the foreground"), which is inherently ambiguous.
Combining reference images with structured layout inputs~\citep{zhang2023controlnet, yang2023reco, li2023gligen, qu2023layoutllmt2i} could enable finer-grained control over subject composition.

\clearpage
\section{Prompts}
\subsection{Prompts of Agentic Framework}
\begin{tcolorbox}[title=Prompt of the \textit{Generator} in the IPR framework]
\small
\{prompt\}\{reference\_image\_files\}
\end{tcolorbox}

\begin{tcolorbox}[title=Prompt of the \textit{Planner} in the IPR framework]
\small
You are a prompt-refiner agent.

\vspace{1mm}
Previous prompt: \{previous\_prompt\}

\vspace{1mm}
You are given multiple reference images and one generated image.

\vspace{1mm}
The generated image is the last one, and the reference images come before it.

\vspace{1mm}
Based on the reference images and the generated image (the last image), propose a refined prompt that improves how reference images are used and adjusts the composition/style.

\vspace{1mm}
Return only the new prompt text.

\vspace{1mm}
\{reference\_image\_files\}\{previously\_generated\_image\_file\}
\end{tcolorbox}

\begin{tcolorbox}[title=Prompt of the \textit{Generator} in the CAFG framework]
\small
\{prompt\}

\vspace{1mm}
The last image provided is the previously generated image from the last step and all images before it are reference images.

\vspace{1mm}
Please refine the previous generation using the reference images and enhance composition/style.

\vspace{1mm}
\{reference\_image\_files\}\{previously\_generated\_image\_file\}
\end{tcolorbox}

\begin{tcolorbox}[title=Prompt of the \textit{Planner} in the CAFG framework]
\small
You are a prompt-refiner agent.

\vspace{1mm}
Previous prompt: \{previous\_prompt\}

\vspace{1mm}
You are given multiple reference images and one generated image.

\vspace{1mm}
The generated image is the last one, and the reference images come before it.

\vspace{1mm}
Based on the reference images and the generated image (the last image), propose a refined prompt that improves how reference images are used and adjusts the composition/style.

\vspace{1mm}
Return only the new prompt text.

\vspace{1mm}
\{reference\_image\_files\}\{previously\_generated\_image\_file\}
\end{tcolorbox}

\begin{tcolorbox}[title=Prompt of the \textit{Generator} in the SRA framework]
\small
\{prompt\}

\vspace{1mm}
The last image provided is the previously generated image from the last step and all images before it are reference images.

\vspace{1mm}
Please refine the previous generation using the reference images and enhance composition/style.

\vspace{1mm}
\{reference\_image\_files\}\{previously\_generated\_image\_file\}
\end{tcolorbox}

\begin{tcolorbox}[title=Prompt of the \textit{Planner} in the SRA framework]
\small
You are a prompt-refiner and reference-selector agent.

\vspace{1mm}
Previous prompt: \{previous\_prompt\}

\vspace{1mm}
You are given multiple reference images (in order) and one generated image (the last one).

\vspace{1mm}
Your task:

\vspace{1mm}
1. Identify which reference images are insufficiently reflected in the generated image.

\vspace{1mm}
2. Propose a refined prompt that improves the generation.

\vspace{1mm}
3. Return your response in JSON format ONLY. Do not include any other text.

\vspace{2mm}
JSON format:

\vspace{1mm}
\{

\vspace{1mm}
\ \ "indices": [0, 2, 3],

\vspace{1mm}
\ \ "prompt": "Your refined prompt text here"

\vspace{1mm}
\}

\vspace{2mm}
Where 'indices' is an array of 0-based indices (from 0 to \{len(reference\_image\_files)-1\}) of reference images 

\vspace{1mm}
that are insufficiently reflected in the generated image.

\vspace{1mm}
'prompt' is the refined instruction prompt for the next generation step.

\vspace{1mm}
Example response:

\vspace{1mm}
\{

\vspace{1mm}
\ \ "indices": [0, 2, 3],

\vspace{1mm}
\ \ "prompt": "Focus more on the lighting from the first image and the composition from images 3 and 4. Emphasize the color palette and texture details."

\vspace{1mm}
\}

\vspace{1mm}
\{reference\_image\_files\}\{previously\_generated\_image\_file\}
\end{tcolorbox}
\clearpage
\subsection{Prompts of VLM as Judge}
\begin{center}
\begin{minipage}{\textwidth}

% \small

\begin{tcolorbox}[title=Multi-Reference Image Genearation Evaluation]
You are a strict data rater specializing in grading multi-reference driven image generation. 
You will be given reference images, a task instruction, and the generation results.

\vspace{0.5mm}
Reference Images: \{reference\_image\_files\}

\vspace{0.5mm}
Editing Instruction: \{instruction\} \\
Final Output: \{generated\_image\_files\}

\vspace{1.0mm}
Your task is to evaluate the effectiveness of replacement editing from five independent perspectives, each on a 10-point scale.
Note that the average score should be considered 4 points.

\vspace{1.0mm}
\noindent
1. Text-Instruction Alignment

\vspace{0.5mm}
Evaluate whether the generated image accurately follows the given text instruction.
Check whether the specified objects appear in the correct positions, whether the instructed subjects are depicted properly, and whether no unintended elements are introduced.
For example, if the instruction says ``change the language,'' but the actual written content itself is altered incorrectly, or if unnecessary objects are added, the score should be reduced.
If the instruction requires including a reference subject but the generated image fails to include that referenced content, the score must be 1.
Even if the instruction is followed correctly, the score must not exceed 6 points if the generated image still exhibits any composited or unnatural appearance.

\vspace{1.0mm}
\noindent
2. Reference Consistency

\vspace{0.5mm}
Evaluate how consistent the generated image is with the provided reference images.
Compare the output to each reference and assess how faithfully the structure and attributes are reproduced.
Fine details, such as hair ornaments, patterns on clothing, and other small features, must match the references; otherwise the score must not exceed 4 points.
If even a single object fails to follow the details of the reference images, the score must not exceed 6 points.

\vspace{1.0mm}
\noindent
3. Background-Subject Match

\vspace{0.5mm}
Evaluate whether the subject blends naturally with the background.
Check whether the subject appears to be floating, unnaturally pasted on, or visually inconsistent with its surroundings.
Images that look like multiple pictures simply pasted together should receive a score of 1.
If there is even the slightest inconsistency in style, tone, lighting, or overall visual impression compared to the reference images, the score must also not exceed 4 points.

\vspace{1.0mm}
\noindent
4. Physical Realism

\vspace{0.5mm}
Evaluate whether the generated image maintains physical plausibility.
Penalize cases where the image violates basic physical laws---for example, a person floating in mid-air, standing on water, or having the lower body missing despite no obstruction.
If there is even a slight impression that the image looks composited or artificially pasted together, the score must not exceed 4 points.
Likewise, if it is unclear whether the subject is actually making proper contact with the ground, the score must also not exceed 6 points.

\vspace{1.0mm}
\noindent
5. Visual Quality

\vspace{0.5mm}
Evaluate the overall perceptual quality of the image.
Assess whether the image is visually appealing and aesthetically coherent.
If the composition appears unnatural or the image does not look aesthetically pleasing to a human observer, the score must not exceed 4 points.

\vspace{1.0mm}
Each of the five scores must be evaluated independently. Do not force any score to be tied to or capped by another score.

\vspace{1.0mm}
First, explain the reasoning, then present the final assessment. \\
Start the reasoning with Reasoning: .

\vspace{0.5mm}
After explaining the reasoning, present the final assessment in the format:

\vspace{0.5mm}
\noindent
Instruction Alignment: $\langle$A number from 1 to 10$\rangle$. \\
Reference Consistency: $\langle$A number from 1 to 10$\rangle$. \\
Background-Subject Match: $\langle$A number from 1 to 10$\rangle$. \\
Physical Realism: $\langle$A number from 1 to 10$\rangle$. \\
Visual Quality: $\langle$A number from 1 to 10$\rangle$.

\end{tcolorbox}

\end{minipage}
\end{center}

\clearpage

\begin{figure*}[ht]
    \centering
    \includegraphics[width=0.70\linewidth]{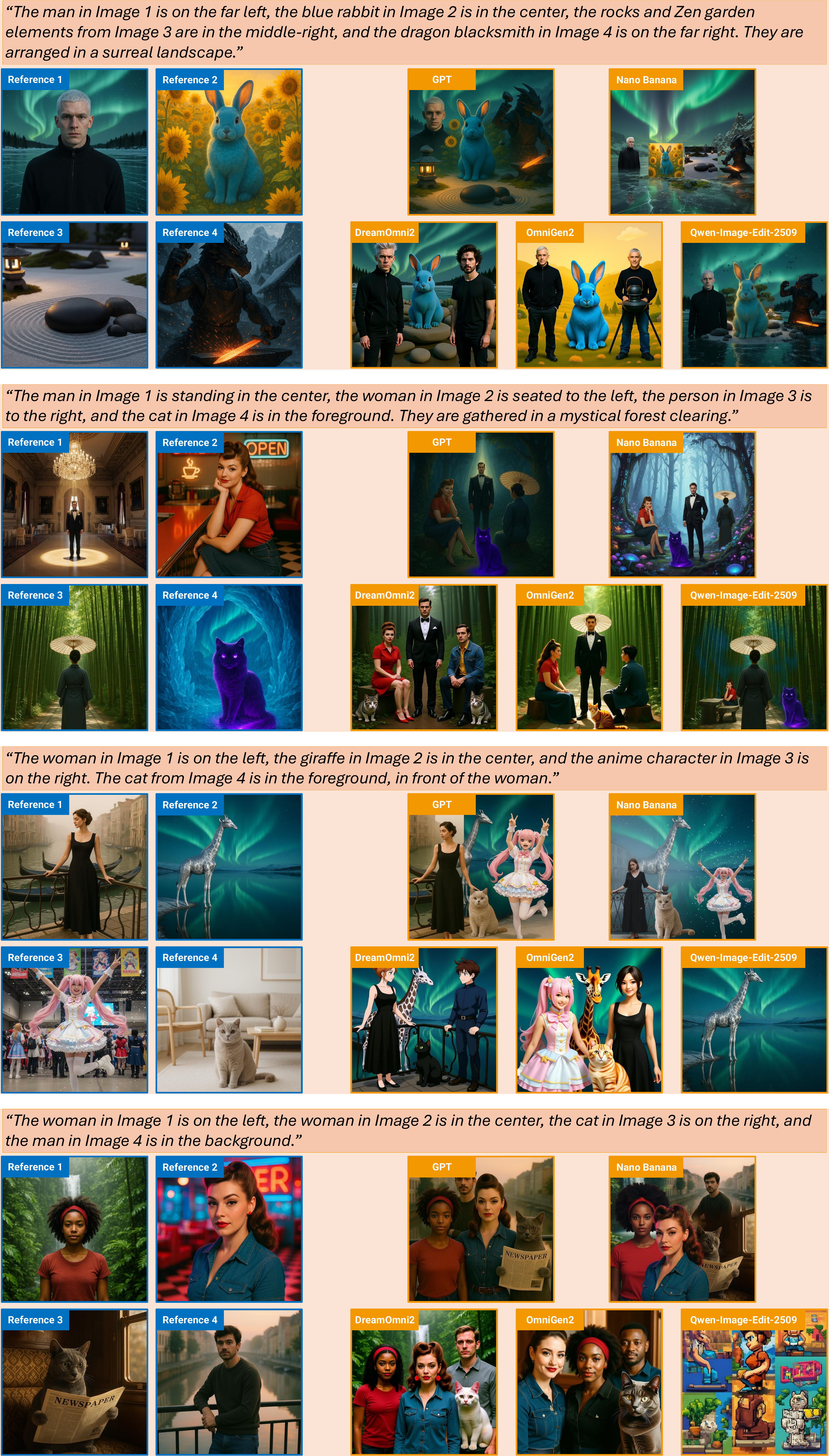}
    \caption{Qualitative example for 4-Object Tasks.}
    \label{fig:qualitative_4_object}
\end{figure*}

\begin{figure*}[ht]
    \centering
    \includegraphics[width=0.70\linewidth]{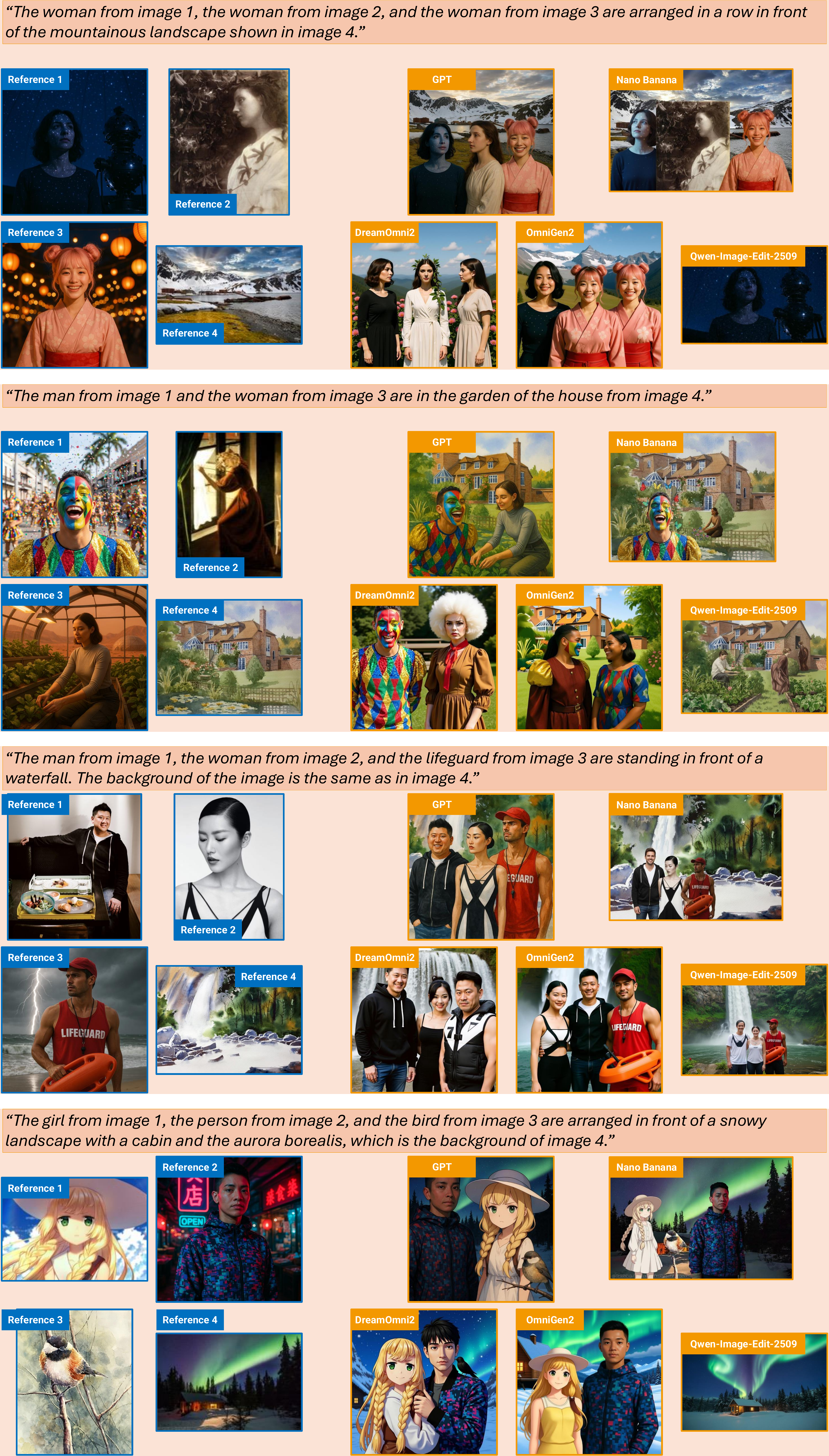}
    \caption{Qualitative example for 3-Object + Background Tasks.}
    \label{fig:qualitative_4_background}
\end{figure*}

\begin{figure*}[ht]
    \centering
    \includegraphics[width=0.70\linewidth]{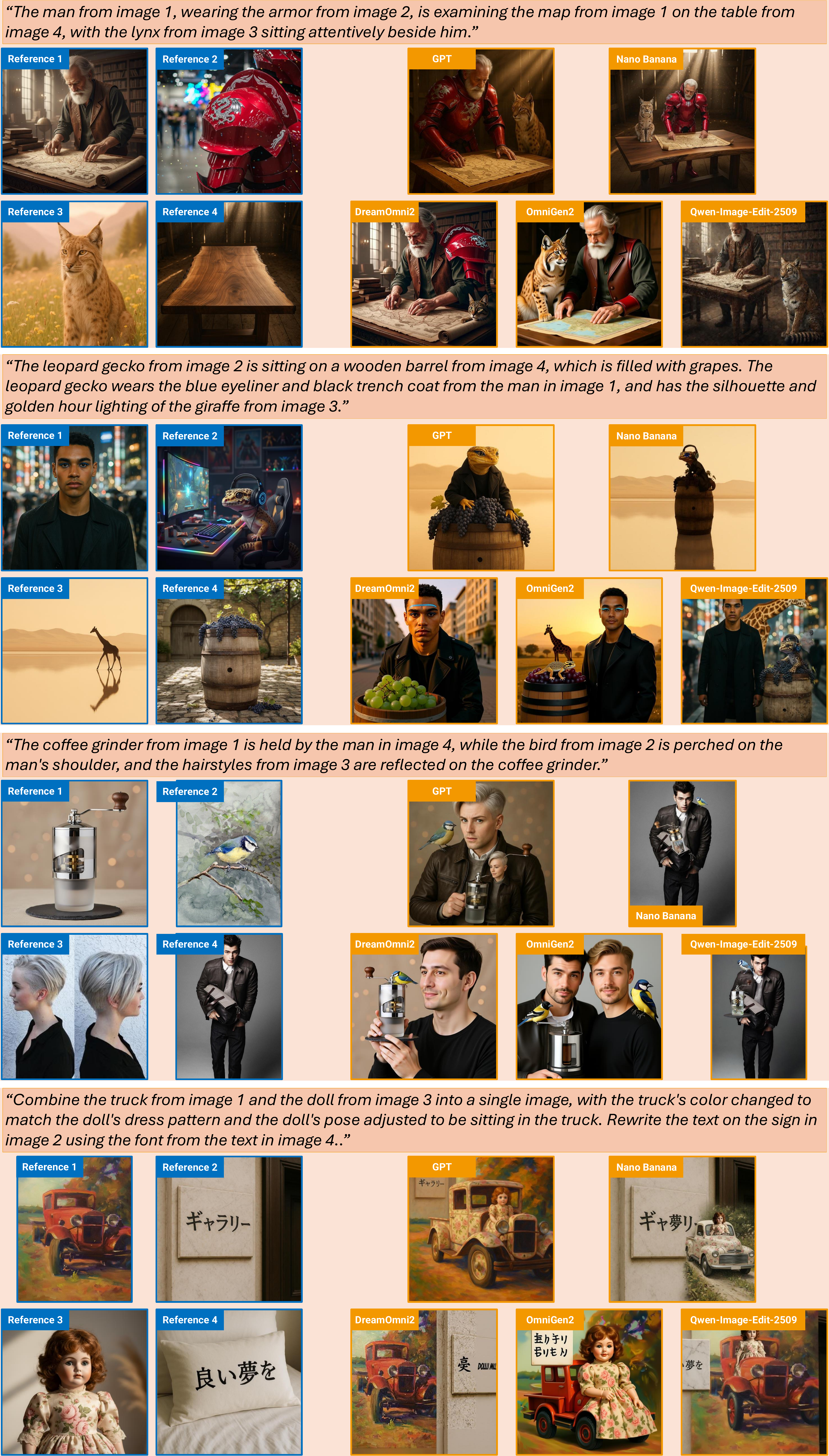}
    \caption{Qualitative example for 3-Object + Local Tasks.}
    \label{fig:qualitative_4_local}
\end{figure*}

\begin{figure*}[ht]
    \centering
    \includegraphics[width=0.70\linewidth]{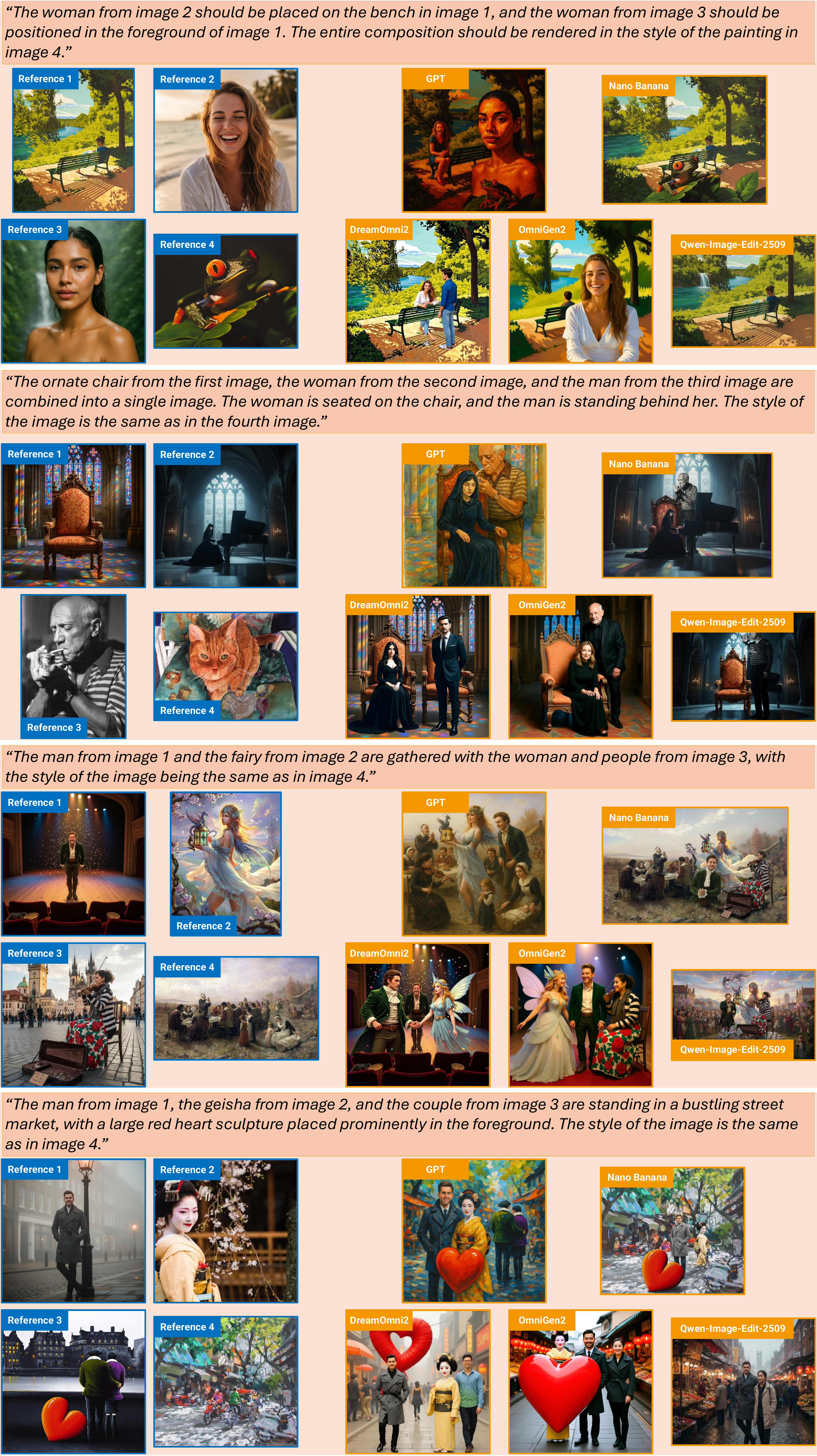}
    \caption{Qualitative example for 3-Object + Global Tasks.}
    \label{fig:qualitative_4_global}
\end{figure*}

% \subsection{Prompts of Data Collection and Filtering} \label{sec:prompts_collection}

% \input{figures_and_tables/prompt_generate_images}

% \input{figures_and_tables/prompt_reject_harmful_images}

% \subsection{Prompts of Data Classification}
% \input{figures_and_tables/prompt_category_classification_person}

% \input{figures_and_tables/prompt_category_classification_object}

% \input{figures_and_tables/prompt_category_classification_background}

% \input{figures_and_tables/prompt_category_classification_background}

% \input{figures_and_tables/prompt_task_construction}

% \input{figures_and_tables/prompt_task_evaluation}

% \section{Rationale}
% \label{sec:rationale}
% % 
% Having the supplementary compiled together with the main paper means that:
% % 
% \begin{itemize}
% \item The supplementary can back-reference sections of the main paper, for example, we can refer to \cref{sec:intro};
% \item The main paper can forward reference sub-sections within the supplementary explicitly (e.g. referring to a particular experiment); 
% \item When submitted to arXiv, the supplementary will already included at the end of the paper.
% \end{itemize}
% % 
% To split the supplementary pages from the main paper, you can use \href{https://support.apple.com/en-ca/guide/preview/prvw11793/mac#:~:text=Delete%20a%20page%20from%20a,or%20choose%20Edit%20%3E%20Delete).}{Preview (on macOS)}, \href{https://www.adobe.com/acrobat/how-to/delete-pages-from-pdf.html#:~:text=Choose%20%E2%80%9CTools%E2%80%9D%20%3E%20%E2%80%9COrganize,or%20pages%20from%20the%20file.}{Adobe Acrobat} (on all OSs), as well as \href{https://superuser.com/questions/517986/is-it-possible-to-delete-some-pages-of-a-pdf-document}{command line tools}.

\end{document}